%% file: ltthinker.tex
\documentclass[]{ltthinker}
\usepackage{fix-cm}
\usepackage{helvet}

\usepackage{xcolor}
\usepackage{multirow}
\usepackage{colortbl}
\usepackage{tcolorbox}
\usepackage{enumitem}
\usepackage{pifont}
\usepackage{wrapfig}
\usepackage{url}

\definecolor{mygray}{gray}{0.9}
\definecolor{myblue}{HTML}{F0FFFF}
\definecolor{myblue_light}{HTML}{EAF6FF}
\definecolor{myblue_dark}{HTML}{2B57A0}
\definecolor{myblue_lightv1}{HTML}{F2F7E0}
\definecolor{mygreen}{RGB}{60, 179, 113}
\definecolor{myyellow_light}{RGB}{255, 255, 224}
\definecolor{lightgreen}{HTML}{D3E5C7}
\definecolor{deepgreen}{RGB}{46,139,87}

\newcommand{\lthink}[1]{{\color{black}#1}}
\newcommand{\ours}{{LightThinker}}
\newcommand{\upsub}[1]{$_{\textcolor{green!60!black}{\uparrow #1}}$}
\newcommand\blfootnote[1]{%
  \begingroup
  \renewcommand\thefootnote{}\footnote{#1}%
  \addtocounter{footnote}{-1}%
  \endgroup
}

\title{LightThinker++: From Reasoning Compression to Memory Management}

\author{
    Yuqi Zhu\textsuperscript{1*},
    Jintian Zhang\textsuperscript{1*},
    Zhenjie Wan\textsuperscript{1},
    Yujie Luo\textsuperscript{1},
    Shuofei Qiao\textsuperscript{1},
    Zhengke Gui\textsuperscript{2},
    Da Zheng\textsuperscript{2},
    Lei Liang\textsuperscript{2},
    Huajun Chen\textsuperscript{1},
    Ningyu Zhang\textsuperscript{1}$\dagger$
}
\affiliation[1]{\mbox{Zhejiang University}}
\affiliation[2]{\mbox{Ant Group}}

\abstract{
Large language models (LLMs) excel at complex reasoning, yet their efficiency is limited by the surging cognitive overhead of long thought traces.
In this paper, we propose \textbf{LightThinker}, a method that enables LLMs to dynamically compress intermediate thoughts into compact semantic representations.
However, static compression often struggles with complex reasoning where the irreversible loss of intermediate details can lead to logical bottlenecks.
To address this, we evolve the framework into \textbf{LightThinker++}, introducing Explicit Adaptive Memory Management.
This paradigm shifts to behavioral-level management by incorporating explicit memory primitives, supported by a specialized trajectory synthesis pipeline to train purposeful memory scheduling.
Extensive experiments demonstrate the framework's versatility across three dimensions.
(1) LightThinker~reduces peak token usage by 70\% and inference time by 26\% with minimal accuracy loss.
(2) In standard reasoning, LightThinker++ slashes peak token usage by 69.9\% while yielding a +2.42\% accuracy gain under the same context budget for maximum performance.
(3) Most notably, in long-horizon agentic tasks, it maintains a stable footprint beyond 80 rounds (a 60\%--70\% reduction), achieving an average performance gain of 14.8\% across different complex scenarios.
Overall, our work provides a scalable direction for sustaining deep LLM reasoning over extended horizons with minimal overhead.
}

\correspondence{\email{zhangningyu@zju.edu.cn}}
\checkdata[Code]{\url{https://github.com/zjunlp/LightThinker}}

\begin{document}

\blfootnote{$^*$Equal Contribution.}
\blfootnote{$^\dagger$Corresponding author.}
\maketitle

\input{sections/introduction}

\input{sections/background}

\input{sections/method}

\input{sections/experiment}

\input{sections/v2/method_agent}

\input{sections/v2/experiment}

\input{sections/related_work}

\input{sections/conclusion_and_future_work}

\bibliographystyle{unsrtnat}
\bibliography{ltthinker}

\appendix
\input{sections/appendix}

\end{document}

%% file: sections/introduction.tex
\section{Introduction}
Recent advancements in Large Language Models (LLMs) have demonstrated their remarkable capabilities in complex reasoning tasks \cite{zhao2023survey,azaria2024chat}. 
As research in this domain progresses, the reasoning patterns of these models have gradually evolved from ``fast thinking'' to ``slow thinking''. 
This transition is exemplified by methods such as Chain-of-Thought (CoT)~\citep{nips22_cot} prompting, which enhances reasoning by breaking down complex problems into sequential sub-steps.  
Building on this, the \textit{o1-like thinking mode}~\citep{arixv24_o1,arxiv24_qwq,arxiv25_deepseek_r1} introduces multiple reasoning abilities such as trial-and-error, backtracking, correction, and iteration, further improving the success rate of models in solving complex problems. 
However, this performance improvement comes at the cost of generating a large number of tokens~\citep{arxiv24_o1_study}. 
Given that current LLMs are predominantly based on the Transformer architecture~\citep{nips17_transformer}, the computational complexity of the attention mechanism grows quadratically with the context length, while the storage overhead of the KV Cache increases linearly with the context length.
For example, in the case of Qwen32B~\citep{arxiv24_qwen2_5}, when the context length reaches \(10^4\), the KV Cache occupies a space comparable to the model itself. 
Consequently, the increase in token generation leads to a sharp rise in memory overhead and computational costs, severely limiting the practical efficiency of LLMs in long-text generation and complex reasoning tasks.

To mitigate this issue, two main approaches have been proposed, primarily differentiated by their intervention requirements during inference.
The first category requires no additional intervention during inference, achieving efficiency through prompt engineering~\citep{arxiv24_tale,arxiv24_break_the_chain,arxiv24_concise_thoughts} or specialized training~\citep{nips24_skip_steps,arxiv24_c3ot,arxiv25_related_work_rl1,arxiv25_o1_pruner,arxiv24_ccot,arxiv24_coconut} to guide LLMs in generating fewer or even zero~\citep{arxiv23_kd_cot,arxiv24_icot} intermediate tokens during reasoning. 
The second category operates through real-time token-by-token intervention during inference~\citep{nips23_h2o,arxiv24_sepllm}, reducing memory usage by selectively retaining important parts of the KV Cache while discarding less critical ones.
However, both approaches face distinct challenges: the first typically requires careful data construction and iterative refinement, while the second introduces substantial inference latency due to the computational overhead of token-wise importance assessment.

\begin{figure}[!t] 
    \centering
    \scalebox{1.0}{
    \includegraphics[width=1.0\linewidth]{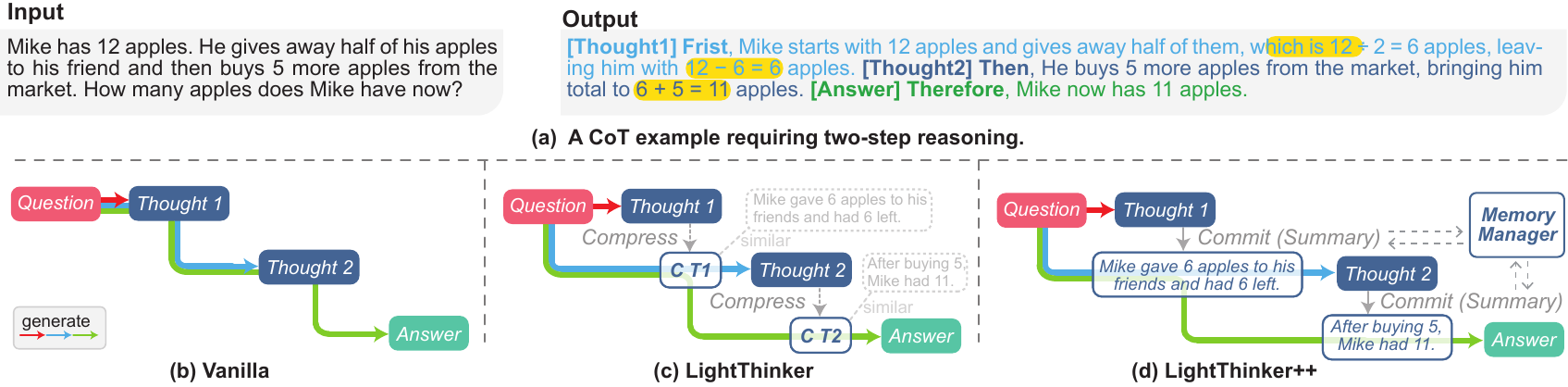} 
    }
    \caption{
    An illustration of the compressed reasoning paradigms.
    (a) A CoT example. Tokens highlighted in yellow represent critical reasoning tokens, while the remaining tokens primarily ensure fluency. 
    (b) The Vanilla approach generates full thought tokens. 
    (c) LightThinker compresses each thought into a concise representation ($C\ T_i$).
    (d) LightThinker++ further incorporates explicit memory management to handle summaries of thoughts, enhancing reasoning efficiency and coherence.
    }
    \label{fig:intro}
    \vspace{-4mm}
\end{figure} 

\lthink{
In this work, we propose a new approach by training LLMs to dynamically compress and manage historical content during reasoning. 
Our motivation stems from the concept of cognitive economy: 
1) Tokens generated by the LLM serve dual purposes of ensuring linguistic fluency and facilitating actual reasoning, which makes it possible to distill the ``gist'' of thoughts.
2) When humans solve complex problems, they do not maintain every intermediate word in active working memory; instead, they store key conclusions mentally and only ``unpack'' or revisit details when encountering logical bottlenecks.

Based on these insights, we first introduce \textbf{LightThinker}, which achieves efficiency through \textbf{representation-level} thought compression.
As illustrated in Fig.~\ref{fig:intro}(c), after generating a lengthy thought step (e.g., \texttt{Thought i}), it is compressed into a compact representation (e.g., \texttt{C Ti}), and the original thought chain is discarded, with reasoning continuing based solely on the compressed content. 
Specifically, we train the LLM to condense lengthy thoughts into a set of hidden states corresponding to a small number of special tokens (i.e., \textit{gist tokens}~\citep{nips23_gist}). 
Through carefully designed attention masks, the LLM learns when and how to compress and continue generating based on the compressed content.

While representation-level compression works well for many reasoning tasks, we find that purely implicit compression can lead to irreversible information loss in more complex scenarios.
Then, to address this limitation, we evolve the framework into \textbf{LightThinker++}, introducing Explicit Adaptive Management. 
As shown in Fig.~\ref{fig:intro}(d), this paradigm shifts to \textbf{behavioral-level} memory management by introducing explicit memory primitives (e.g., \textit{commit}, \textit{expand}, \textit{fold}).
It enables the model to autonomously archive thoughts into semantic summaries or retrieve raw details upon logical necessity, ensuring robustness in both complex reasoning and long-horizon agentic tasks like DeepResearch.
To support this evolution, we develop a collaborative synthesis pipeline, which generates expert trajectories that interleave reasoning with purposeful memory operations, training the model to master complex memory scheduling.

We conduct extensive experiments across four datasets using two representative model families: Llama and Qwen. 
First, with the Qwen model, \textbf{LightThinker} reduces the peak token usage by 70\% and decreases inference time by 26\% compared to the Vanilla model, while maintaining comparable accuracy (with only a 1\% drop).
Furthermore, \textbf{LightThinker++} achieves a superior accuracy--efficiency balance under two distinct scenarios: (1) in the Throughput setting (prioritizing inference speed), it slashes peak memory by 69.9\% while maintaining baseline accuracy; (2) in the Budget setting (prioritizing reasoning quality), it not only reduces peak memory by 45.0\% but also yields a +2.42\% average accuracy gain.
This confirms that a condensed, high-signal context is more effective for complex reasoning than a verbose, unmanaged one.

Beyond standard benchmarks, a more critical challenge is whether the compressed-context reasoning paradigm can also benefit long-horizon agentic tasks, where the context continuously grows over many interaction rounds.
To validate the scalability of our approach in complex scenarios, we further evaluate \textbf{LightThinker++} on long-horizon agentic tasks using the Qwen3-30B-A3B model.
On challenging benchmarks including xBench-DeepSearch, BrowseComp-ZH, and BrowseComp-EN, LightThinker++ demonstrates remarkable efficiency. 
While the Vanilla agent's context inflates to 100k tokens within 60 rounds, LightThinker++ maintains a lean 30k--40k footprint (a 60\%--70\% reduction) even beyond 80 rounds. 
By decoupling reasoning depth from memory consumption, our method delivers an average +4.4\% Pass@1 boost across three benchmarks, with a remarkable 2.51$\times$ performance leap on the challenging hard subset.

Our contributions are as follows:
1) We propose LightThinker, a method that dynamically compresses thought chains during reasoning, significantly reducing memory overhead and inference time.
2) We also present LightThinker++, a framework for explicit adaptive memory management using memory primitives, and show that it can work effectively for both standard reasoning benchmarks and agentic tasks.
3) We demonstrate that the LightThinker~ family achieves a good balance between reasoning efficiency and accuracy, offering new insights for future LLM inference acceleration.
}

%% file: sections/background.tex
\section{Background}

\textbf{Slow Thinking.}
The ability of LLMs to perform reasoning is fundamental~\citep{acl23_reason_survey}, particularly when addressing complex tasks that require moving beyond the rapid, intuitive mode of System 1 toward the deliberative mode of System 2~\citep{pb96_system12,fsg11_thinking_fast_slow,aaai21_machine_fast_slow}. 
A representative approach is Chain-of-Thought (CoT)~\citep{nips22_cot}, which tackles difficult problems by breaking them down into smaller steps and solving them sequentially. 
Building on this idea, the \textit{o1-like thinking mode}~\citep{arixv24_o1,arxiv24_qwq,arxiv25_deepseek_r1} introduces additional behaviors such as trial-and-error, reflection, backtracking, and self-correction. 
Existing empirical results~\citep{arixv24_o1,arxiv25_deepseek_r1} indicate that this thinking mode yields markedly better performance than CoT on challenging reasoning tasks. 
Such slow-thinking behavior can be learned by models through Supervised Fine-Tuning (SFT) with carefully designed training data. 
From the perspective of generation length, the token usage of System 1, CoT, and \textit{o1-like thinking mode} increases in an orderly manner.

\noindent \textbf{Inference Challenges.}
Prior studies on the \textit{o1-like thinking mode}~\citep{arxiv24_o1_study} have pointed out that solving complex problems often requires producing a large number of tokens.
However, this long-generation behavior poses two major challenges for the attention mechanism, which lies at the core of Transformers~\citep{nips17_transformer}. 
First, the \textit{memory} burden grows continuously during inference.
To improve decoding efficiency, the Key and Value of each token are stored in the KV cache at every layer.
For Qwen-32B~\citep{arxiv24_qwen2_5}, once the context length reaches $10^4$ tokens, the memory footprint of the KV cache becomes comparable to the size of the model itself.
Second, the \textit{computational cost} of autoregressive generation also increases substantially.
Because of the attention operation in Transformers~\citep{nips17_transformer}, the computation required for inference scales \textit{quadratically} with the number of tokens.

%% file: sections/method.tex
\section{Methodology}
We present the \textbf{{LightThinker} family} to accelerate the reasoning process of LLMs, as illustrated in Figure~\ref{fig:intro}.
The core idea is to \textit{train LLMs to dynamically compress the current thought during reasoning}, enabling subsequent generation to be based on the compressed content rather than the original long thought.

Here we introduce two complementary paradigms.
\textbf{{LightThinker}} (\S\ref{sec:method:implicit}) targets maximal efficiency via \emph{implicit hidden-state compression}, where completed thought spans in the attention context are replaced with a small number of compressed tokens.
\textbf{{LightThinker}++} (\S\ref{sec:method:explicit}) improves robustness for \emph{general reasoning} by introducing \emph{explicit adaptive memory management} and a set of controllable actions that support on-demand archival and expansion of information.
While our experiments focus on the reasoning setting, this explicit management mechanism also naturally extends to long-horizon agentic scenarios (see \S\ref{sec:agent}).

\subsection{Overview}
\lthink{
The efficiency of complex reasoning is fundamentally constrained by the cognitive overhead of processing long contexts. 
Inspired by human Working Memory (WM) and the Information Bottleneck (IB) principle, intelligence emerges not from retaining all information, but from strategically compressing and maintaining what is predictive for future reasoning. 
Specifically, we propose a hierarchical framework that progresses from implicit, representation-level distillation to explicit, behavioral-level memory management, analogous to the shift from automatic intuition to deliberate control in human cognition.
From this perspective, the LightThinker family can be understood as a progressive design that balances efficiency, interpretability, and controllability through different levels of memory abstraction.

\paragraph{\textbf{From Implicit Representation to Explicit Management.}}
Based on the principle of thought compression, we construct two progressive paradigms to balance reasoning efficiency and task complexity:
\begin{itemize}
\item \textbf{LightThinker:} Focuses on \textit{representation-level} optimization via an information bottleneck.
It implicitly encodes thoughts into hidden state Cache Tokens, offering extreme efficiency without altering the output format. 
However, its lossy nature may limit performance in high-precision tasks.
\item \textbf{LightThinker++:} Advances toward \textit{behavior-level} memory management.
To mitigate the irreversibility of implicit compression, we introduce explicit memory mechanisms that allow the model to regulate information retention and reactivation.
The model learns to actively manage its memory by archiving details or retrieving them upon logical necessity, ensuring robustness in complex, long-horizon reasoning.
\end{itemize}
}
\paragraph{\textbf{Design.}} To realize these paradigms, we systematically address two key design dimensions:
\lthink{
\begin{itemize}
    \item \textbf{When to compress?} The timing of compression dictates the balance between reasoning efficiency and semantic integrity. 
    In LightThinker, we employ \textit{rule-driven} strategies such as \textit{token-level}~\citep{iclr25_activation_beacon} or \textit{thought-level}~\citep{acl24_anllm} triggers. 
    In LightThinker++, this evolves into a \textit{model-driven} decision, where the LLM autonomously invokes different memory operations based on reasoning complexity.

    \item \textbf{How to compress?} The objective is to transform lengthy thoughts into compact, navigable representations. 
    LightThinker~utilizes \textit{hidden state compression} via \textit{gist tokens}~\citep{nips23_gist}, implemented through a \textit{thought-based attention mask}. 
    In contrast, LightThinker++ upgrades this to \textit{explicit summarization and bidirectional management}, dynamically reorganizing contextual information to approximate human working memory regulation.
\end{itemize}
}

\paragraph{What content has been compressed?}
We do not aim to compress lengthy thought information into a compact representation without loss. 
Instead, our focus is on preserving only the information that is essential for subsequent reasoning.
As highlighted by the gray dashed box in Figure~\ref{fig:intro}, the lengthy thought is retained solely for the elements that contribute to further inference.

\begin{figure*}[t]
    \centering
    \begin{minipage}[t]{0.70\textwidth}
        \vspace{0pt}
        \scalebox{1}{\includegraphics[width=\linewidth]{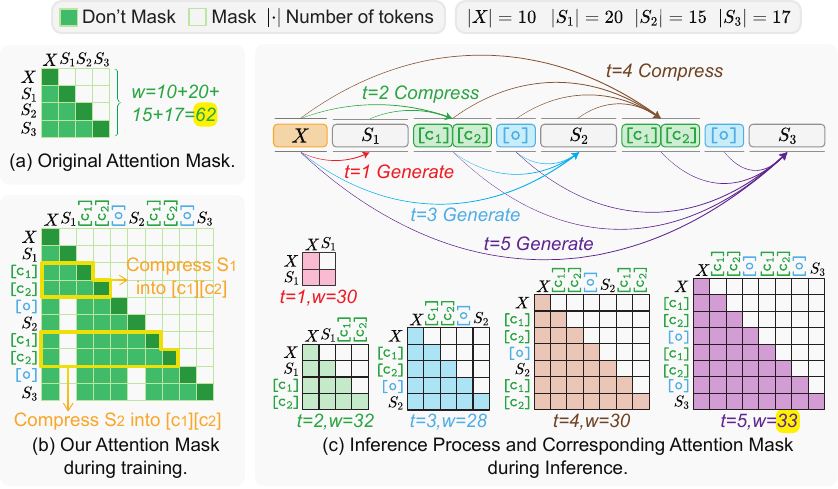}} 
        \caption{
        An overview of LightThinker, illustrated with a three-step reasoning example.
        Fig. (a) shows the attention mask of Vanilla during both training and inference.
        Fig. (b) depicts the attention mask of LightThinker during the training.
        Fig. (c) presents the complete inference process of LightThinker along with the attention mask corresponding to each step. 
        Here, `w' denotes the size of the matrix.}
        \label{fig:method}
    \end{minipage}
    \hfill
    \begin{minipage}[t]{0.28\textwidth}
        \vspace{0pt}
        \scalebox{1}{\includegraphics[width=\linewidth]{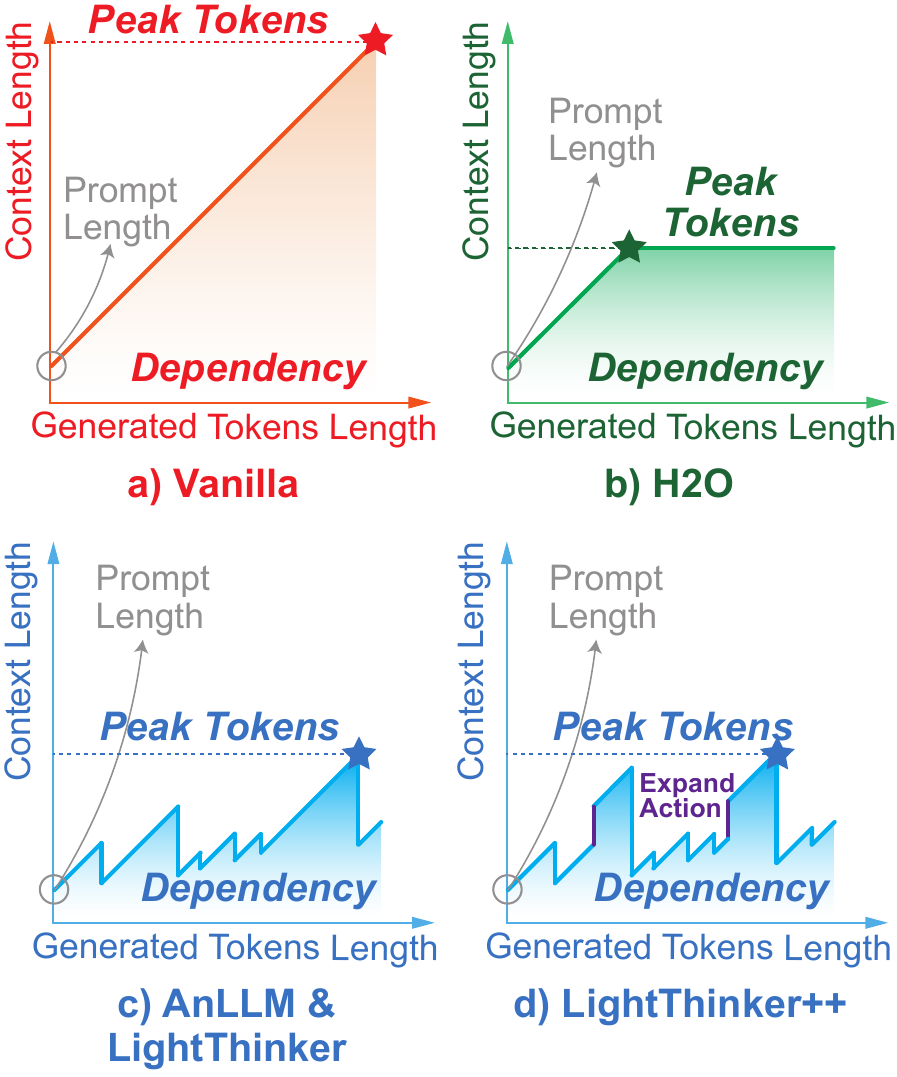}} 
        \caption{
        Relationship between context length and the number of generated tokens across different methods. 
        The \textit{Dependency} metric corresponds to the area under the curve, while \textit{Peak Token} indicates the curve's maximum value.
        See Appx.~\ref{sec:app:metric} for details.
        }
        \label{fig:exp:metric}
    \end{minipage}
\end{figure*}

\subsection{\ours: Implicit Thought Compression}
\label{sec:method:implicit}
\paragraph{\textbf{Notation.}}
We first clarify the notation used in this section. 
A lowercase symbol such as $x_i$ refers to a single token, while an uppercase symbol such as $X$ denotes a token sequence. 
We use `$\texttt{[·]}$' to represent a special token, e.g., `$\texttt{[c]}$', and `$\texttt{<·>}$' to denote an optional special token, such as `$\texttt{<w>}$'. 
Our training corpus is the \textit{o1-like thinking mode} dataset $\mathcal{D}=\{(X,Y)_i\}_{i=1}^{|\mathcal{D}|}$, where $X=\{x_i\}_{i=1}^{|X|}$ is a question and $Y=\{y_i\}_{i=1}^{|Y|}$ is the corresponding reasoning trace together with the final answer. 
Prior work~\citep{arxiv25_openthoughts,arxiv25_deepseek_r1} has demonstrated that SFT on $\mathcal{D}$ can substantially improve the reasoning performance of LLMs.

\paragraph{\textbf{Data Reconstruction.}}
Our goal is to expose the model to an intermediate compression process while it is generating reasoning traces. 
To this end, we transform each sample in $\mathcal{D}$ into a compressed form $\hat{\mathcal{D}}$. 
Given an input-output pair $(X,Y)$, we first apply a segmentation function $\texttt{Seg()}$ to split $Y$ into $k$ segments, written as $Y=\{S_i\}_{i=1}^{k}$. 
The segmentation may operate either at the token level or at the thought level, depending on the desired granularity. 
After segmentation, we augment the sequence by inserting three types of special tokens between neighboring segments: $\texttt{<w>}$, $C$, and $\texttt{[o]}$. 
Here, $\texttt{<w>}$ acts as an optional signal that indicates the preceding segment should be compressed; it can be removed when token-level segmentation is used or when the trigger already appears in $S_i$. 
The component $C=\{\texttt{[c}_\texttt{i}\texttt{]}\}_{i=1}^{|C|}$ is a fixed set of gist tokens, which serve as a compact carrier of the distilled information from the preceding segment. 
We refer to this set as the \textit{cache tokens}, and use $|C|$ to denote the \textit{cache size}. 
The token $\texttt{[o]}$ is required to resume generation from the compressed representation, inspired by~\citeauthor{emnlp24_onegen}. 
With these insertions, the reconstructed output becomes
\begin{align*}
\hat{Y}=\{S_1, \underline{\texttt{<w>}, C, \texttt{[o]}}, S_2, \underline{\texttt{<w>}, C, \texttt{[o]}}, \dots, S_k\}.
\end{align*}
Accordingly, the transformed dataset is $\hat{\mathcal{D}}=\{(X,\hat{Y})_i\}_{i=1}^{|\hat{\mathcal{D}}|}$. 
For simplicity, we assume $\texttt{<w>}\in S_i$ and omit it in the rest of the section. 
We also use superscripts to distinguish instances of identical special tokens at different locations, e.g., $C^{(1)}$ and $\texttt{[o]}^{(1)}$ denote the tokens following $S_1$.

\paragraph{\textbf{Thought-based Attention Mask Construction.}}
To ensure that the model learns both how to compress and how to reason from compressed content, we introduce the \textit{Thought-based Mask Construction} illustrated in Fig.~\ref{fig:method}(b). 
Let $S_{<i}=\{S_1,\dots,S_{i-1}\}$ denote the sequence of thoughts before the $i$-th thought $S_i$.

In the compression stage, tokens in $C^{(i)}$ are permitted to attend only to the question $X$, the previously generated compressed content $\{{C,\texttt{[o]}}\}^{(<i)}$, and the current thought $S_i$. 
This process can be expressed as
\begin{align*}
C^{(i)} \leftarrow \texttt{Cmp}(X, \{C^{(1)}, \texttt{[o]}^{(1)}, \dots, C^{(i-1)}, \texttt{[o]}^{(i-1)}\}, S_i),
\end{align*}
where $\texttt{Cmp()}$ denotes the compression operation. 
Under this constraint, the model is encouraged to distill the essential information in $S_i$ into $C^{(i)}$. 
A detailed mathematical description of $\texttt{Cmp()}$ is given in Appx.~\ref{sec:app:compress}.

In the generation stage, $\texttt{[o]}^{(i)}$ can only attend to the question $X$ and the compressed history $\{C,\texttt{[o]}\}^{(\le i)}$. 
Accordingly, the next segment is generated by
\begin{align*}
S_{i+1}\leftarrow \texttt{Gen}(X,\{C^{(1)}, \texttt{[o]}^{(1)},\dots,C^{(i)},\texttt{[o]}^{(i)}\}),
\end{align*}
where $\texttt{Gen()}$ denotes the generation operation. 
This design allows the model to continue reasoning based on the question and the compact summary of previous thoughts.

\paragraph{\textbf{Training and Inference.}}
The training objective is to maximize the likelihood of the reconstructed sequence under the model parameterized by $\theta$:
$$
P_\theta(S_1|X)\cdot\prod_{k=2}^{K} P_\theta(S_k|X,\{C^{(i)},\texttt{[o]}^{(i)} \}_{i=1}^{k-1}\})
$$
where $\theta$ represents the LLM parameters. 
During training, the model is not required to predict the input $X$ or the inserted special tokens $C$ and $\texttt{[o]}$. 
Instead, these tokens serve as structural placeholders that guide the model to learn compression-aware reasoning. 
The optimization still follows the standard next-token prediction paradigm, and the samples are drawn from $\hat{\mathcal{D}}$ with an attention mask that enforces the desired dependency structure. 
The detailed inference procedure is shown in Fig.~\ref{fig:intro}(c) and Fig.~\ref{fig:method}(c). 

\input{sections/v2/method}

%% file: sections/v2/method.tex
\begin{figure}[!t] 
    \centering
    \scalebox{0.9}{
    \includegraphics[width=1.0\linewidth]{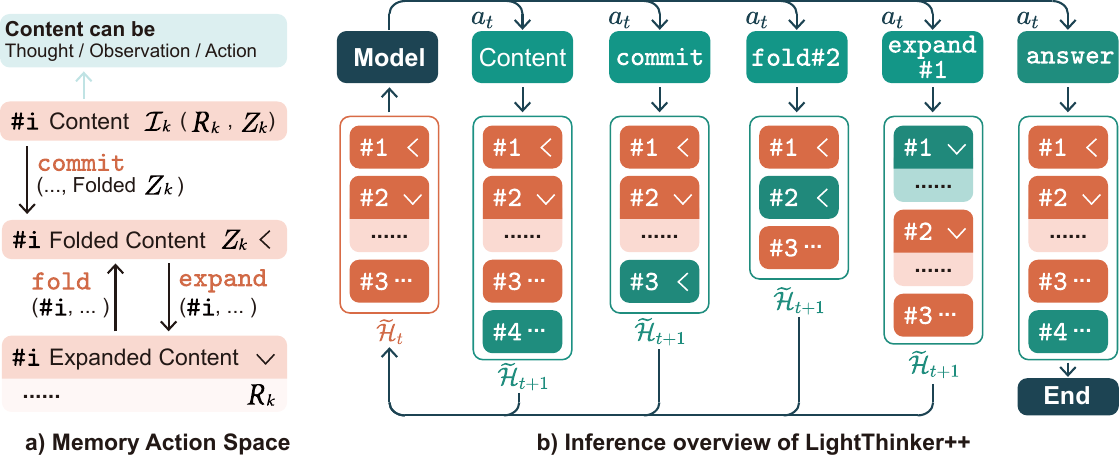} 
    }
    \caption{Overview of \ours++.
    a) \textbf{Memory Action Space}: Reasoning steps are instantiated as dual-form entities $\mathcal{I}_i = (R_i, Z_i)$. The visibility state of each step is explicitly managed via \textit{commit}, \textit{expand}, and \textit{fold} primitives. 
    b) \textbf{Inference Overview}: An illustration of the step-wise inference process. $\tilde{\mathcal{H}}_t$ denotes the stateful managed context where historical steps are dynamically projected as either \textbf{summaries} ($Z_i$, marked with $<$) or \textbf{raw derivations} ($R_i$, marked with {$\vee$}) based on the model's self-directed memory policy.
    }
    \label{fig:method:v2}
\end{figure} 
\lthink{
\subsection{\ours++: Explicit Adaptive Memory Management}
\label{sec:method:explicit}
While the implicit compression in Sec.~\ref{sec:method:implicit} provides significant efficiency gains, its predefined information bottleneck may struggle with highly complex reasoning tasks which require precise backtracking to intermediate logical steps. To address this, we propose \textbf{LightThinker++}, a paradigm that evolves thought compression from the \textit{representation-level} to the \textit{behavioral-level} by empowering the model to actively manage its own context memory through a dynamic substitution mechanism.

\paragraph{\textbf{The Explicit Memory Management Framework.}}
As illustrated in Fig.~\ref{fig:method:v2}(b), We formalize the model's reasoning history as an ordered sequence of \textbf{reasoning entities} $\mathcal{M} = \{\mathcal{I}_1, \dots, \mathcal{I}_K\}$, where each entity $\mathcal{I}_k$ represents a discrete, instantiated \textbf{reasoning step}. 
Each entity $\mathcal{I}_k = (R_k, Z_k)$ is a dual-form container: $R_k$ captures the \textbf{Raw Reasoning} (the full, verbose derivation), while $Z_k$ provides a \textbf{Semantic Summary} of its logical core.
As the reasoning progresses to round $t$, the model operates within a \textbf{managed context} $\tilde{\mathcal{H}}_t = \{m_1^{(t)}, \dots, m_K^{(t)}\}$, where $m_k^{(t)}$ is a dynamic projection of the $k$-th reasoning step. This projection is governed by an explicit \textbf{visibility state} $\sigma_k \in \{\texttt{archive}, \texttt{active}\}$, updated by the model's own memory primitives $a_{t-1}$:
\begin{equation}
m_k^{(t)} = 
\begin{cases} 
Z_k, & \text{if } \sigma_k^{(t)} = \texttt{archive} \quad (\text{Compressed Thought}) \\
R_k, & \text{if } \sigma_k^{(t)} = \texttt{active} \quad (\text{Restored Thought})
\end{cases}
\label{equ:memory_projection}
\end{equation}

This architecture enables \textbf{granularity-aware control} over the thought stream through two core mechanisms:
\begin{itemize}
    \item \textbf{Step Instantiation}: A new entity $\mathcal{I}_{K+1}$ is instantiated only when the model emits $a = \textit{commit}(R, Z)$. This signals the completion of a reasoning unit, allowing the verbose $R$ to be offloaded from the active workspace while preserving its distilled gist $Z$ in the \texttt{archive} state.
    \item \textbf{Step Manipulation}: By outputting $\textit{expand}(k)$ or $\textit{fold}(k)$, the model actively ``toggles'' the resolution of the $k$-th historical step. 
    This allows the model to re-examine evidentiary details through \textit{expansion} when encountering logical bottlenecks, and subsequently \textit{fold} them back to maintain a high-signal, noise-free context.
    \item \textbf{Termination}: The primitive $\textit{answer}(\hat{a}_{\text{ans}})$ concludes the entire reasoning chain by submitting the final solution.
\end{itemize}

By decoupling reasoning depth from sequence length growth, LightThinker++ ensures that the model can sustain long-range coherence in complex Chain-of-Thought tasks, effectively mitigating performance degradation caused by contextual redundancy.
\paragraph{\textbf{Environment-Aware Trajectory Synthesis.}}
\label{sec:data_synthesis}
To train the model in mastering these behaviors, we develop an \textit{Online Thought Synthesis} framework designed to simulate a memory-constrained environment. 
Concretely, we use a strong teacher model to generate high-quality reasoning trajectories that are interleaved with explicit memory actions, forming demonstrations of how to reason under a managed context.
Unlike traditional static supervised fine-tuning, our framework is environment-aware: when the teacher model issues a structural $\textit{commit}$ call, the environment dynamically modifies the prompt for the next iteration by setting $\sigma_k = \texttt{archive}$ and providing the summary $Z
_k$, thereby hiding the raw reasoning $R_k$. 
This closed-loop synthesis forces the teacher model to continue its deduction under a state of true memory compression, creating high-fidelity trajectories that interleave reasoning, archiving, and on-demand retrieval.

\paragraph{\textbf{Behavioral Pruning and Quality Control.}}
To extract the most effective reasoning patterns from the synthesized trajectories, we implement a \textbf{Behavioral Pruning} mechanism centered on a strict \textbf{Memory Lifecycle} constraint. This ensures the fine-tuning data reflects purposeful context management rather than stochastic tool usage. 
A trajectory is deemed admissible only if it satisfies the following criteria:
(1) \textbf{Lifecycle Completeness}: The trajectory must exercise the full $\mathcal{A}_{\text{mem}}$ suite and yield a correct \textit{answer} through consistent tool invocations, ensuring the reasoning process is both functional and verifiable.
(2) \textbf{Symmetry Constraint}: We enforce structural integrity where a \textit{fold} operation must be strictly preceded by an \textit{expand} on the same step.
This ensures retrieved details are purposefully reverted once their immediate utility within the reasoning chain is exhausted.
(3) \textbf{Anti-Jitter Heuristics}: To ensure management is purposeful rather than stochastic, we enforce \textit{operational density} limits where $N_{exp} \le N_{com}$ and $N_{exp} + N_{fld} \le 2N_{com}$, with $N_{\text{action}}$ denoting the total occurrences of each respective primitive within a trajectory.
In addition, we disallow consecutive identical memory operations (e.g., back-to-back \textit{expand} or \textit{fold}) and discard trajectories in which consecutive \textit{commit} steps have a lexical similarity measured by  longest common subsequence greater than 0.9, to encourage progressive reasoning.
Such filtering encourages monotonic logical progression and prevents the model from falling into repetitive or redundant reasoning loops.

\paragraph{\textbf{Training and Implementation.}}
Through this rigorous filtering, we obtain a final expert dataset $\mathcal{D}$ of high-quality trajectories. To ensure the model internalizes the causal relationship between its cognitive operations and context states, we treat each trajectory $\mathcal{T}$ as a sequence of \textit{context-action transitions}.
Specifically, instead of training on a static full-length chain, we decompose each trajectory into a series of training instances $\{ (X, \tilde{\mathcal{H}}_k) \to \mathcal{I}_k \}_{k=1}^K$.
We fine-tune the model by maximizing the predictive likelihood over the pruned expert set $\mathcal{D}$:
\begin{equation}
\mathcal{L}_{\mathrm{SFT}}(\theta) = -\mathbb{E}_{\mathcal{T}^*\sim\mathcal{D}} \left[ \sum_{k=1}^{K} \log P_\theta (\mathcal{I}_k \mid X, \tilde{\mathcal{H}}_k) \right]
\label{equ:reasoning_loss}
\end{equation}
where $\tilde{\mathcal{H}}_k$ is the managed history reconstructed at each step $k$. Following the environmental state $\sigma{j<k}$ determined during synthesis, $\tilde{\mathcal{H}}_k$ only retains the raw reasoning $R_j$ if it remains \texttt{active}, while replacing it with the summary $Z_j$ once it is \texttt{archived}. 
This step-wise state-aware training forces the model to learn that its current reasoning $\mathcal{I}_k$ must remain logically consistent even when conditioned on a highly compressed historical context.

\subsection{Discussions: Implicit vs. Explicit}
\label{sec:method:discussion}

The transition here in our work reflects a shift from structural optimization to behavioral management, addressing the inherent limitations of rigid information bottlenecks. We summarize their key differences as follows:

\textit{\textbf{Information Abstraction Paradigm.} }
LightThinker\ operates at the \textit{representation-level} via attention masking to enforce a fixed bottleneck. 
In contrast, LightThinker++\ shifts to a \textit{behavioral-level} approach, delegating context control to explicit memory primitives. 
This alignment between reasoning actions and working memory states enables interpretable context orchestration that adapts to the model's logical needs.

\textit{\textbf{Static vs. Adaptive Rhythm.}} While LightThinker follows predefined token or thought-level intervals, LightThinker++~is inherently adaptive. It empowers the model to autonomously decide when to archive or retrieve information, demonstrating a more flexible cognitive economy for varying task complexities.

\textit{\textbf{Efficiency vs. Fidelity.}} 
LightThinker\ is optimized for maximal inference speedup, making it ideal for standard tasks where a general gist of thoughts suffices. However, the lossy nature of hidden-state compression can lead to the irreversible loss of critical details. LightThinker++\ prioritizes reasoning fidelity by ensuring that fine-grained information remains restorable through explicit backtracking via \textit{expand}. 
This bidirectional capability mitigates the ``information evaporation'' common in implicit bottlenecks, providing the precision necessary for long-horizon tasks.
}

%% file: sections/experiment.tex
\section{Experiments: General Reasoning}

\subsection{Experimental Settings}

\input{tables/exp_main}

\textbf{Baselines.}
1)~\ours. We evaluate our method on two backbone LLMs: Qwen2.5-7B~\citep{arxiv24_qwen2_5} and Llama3.1-8B~\citep{arxiv24_llama_3}.
To provide an upper-bound reference, we apply full-parameter instruction tuning on the Bespoke-Stratos-17k dataset (BS17K; an example is shown in Fig.~\ref{prompt:case:train}), and refer to the resulting model as \textit{Vanilla}. 
Following our preliminary experiments, training is initialized from the R1-Distill~\citep{arxiv25_deepseek_r1} (e.g., \texttt{DeepSeek-R1-Distill-Qwen-7B}), since fine-tuning instruction-tuned models such as \texttt{Qwen2.5-7B-instruct} leads to only modest gains. 
For comparison, we consider five baselines: 
two training-free acceleration methods applied to Vanilla, namely H2O~\citep{nips23_h2o} and SepLLM~\citep{arxiv24_sepllm}, both of which preserve important KV cache entries using different strategies; 
one training-based approach, AnLLM~\citep{acl24_anllm}; and two CoT~\citep{nips22_cot} baselines, obtained by prompting the instruction-tuned model and the R1-Distill model, respectively.

\lthink{2)~\ours++.
We follow the same model setup, conducting experiments on the R1-Distill models and fine-tuning all methods from the corresponding R1-Distill checkpoints.
For this setup, we curated a high-quality distillation dataset by sampling from BS17k and DeepScaleR~\citep{deepscaler2025}, employing DeepSeek-V3.2-Thinking as the teacher model for data synthesis. 
To ensure a comprehensive comparison, we introduce four additional baselines:
the Vanilla model, which is fine-tuned on the uncompressed, full-length original outputs synthesized by our framework;
TokenSkip~\citep{tokenskip}, which fine-tunes models on pruned CoT paths to enable selective token-level skipping;
and the Base Prompting results, which provide the performance of the original models without fine-tuning.
For simplicity, we use LThinker and LThinker++ to denote our models.
}

\textbf{Evaluation Metrics and Datasets.}
We conduct experiments on four benchmark datasets: GSM8K~\citep{arxiv21_gsm8k}, MMLU~\citep{iclr21_mmlu}, GPQA~\citep{colm24_gpqa}, and BBH~\citep{acl23_bbh}. 
For MMLU and BBH, we evaluate on randomly sampled subsets rather than the full datasets. 
To assess model performance, we consider two aspects: \textit{effectiveness} and \textit{efficiency}. 
Effectiveness is measured by accuracy (\textit{Acc}), while efficiency is evaluated using three indicators: inference time (\textit{Time}), the maximum number of tokens appearing in the context during decoding (\textit{Peak}), and the cumulative \textit{dependency} of generated tokens on earlier tokens (\textit{Dep}). 
As illustrated in Fig.~\ref{fig:exp:metric}, Peak reflects the highest contextual load at a specific moment, whereas Dep is represented by the area enclosed by the curves and captures the overall information usage throughout inference. 
A smaller Dep value indicates that the model relies on less information, suggesting stronger compression. 
Since Peak measures a transient state and Dep summarizes the entire generation process, the two quantities are not directly linked. 
Further details on Dep are provided in Appx.~\ref{sec:app:metric}.

\textbf{Implementation}
1) \textit{Variants of LThinker}.
To study implicit compression at different granularities, we consider two variants of LThinker. 
LThinker$_{tok}$ operates at the \textit{token-level} and compresses every 6 original tokens into 2 gist tokens, i.e., $|C|=2$. 
LThinker$_{tho}$ is the \textit{thought-level} variant, where ``\texttt{\textbackslash n\textbackslash n}'' serves as the boundary between thoughts; each thought is then compressed into $|C|=9$ tokens for Qwen backbones and $|C|=7$ tokens for Llama backbones.
2) \textit{Variants of LThinker++}. For the explicit management, we compare: LThinker++, the full model supporting all memory primitives (\textit{commit}, \textit{expand}, and \textit{fold}) for reversible and dynamic context management;
and LThinker$^*$, a \textit{commit-only} ablation variant that performs irreversible summary compression, used to verify the necessity of active retrieval (\textit{expand} and \textit{fold}). 
We evaluate LThinker++ under two inference configurations, Throughput and Budget, to assess its performance under varied serving constraints(see Sec.~\ref{sec:main_lthink++}).

Please refer to Appx.~\ref{sec:app:exp:ltthink} and Appx.~\ref{sec:app:exp:ltthink++} for further technical details on LThinker and LThinker++, respectively.

\subsection{Evaluation of LightThinker.}
We assess LThinker from three perspectives: overall performance, inference efficiency, and ablation-based component analysis. 
We also include a qualitative case study to better understand its behavior.

\subsubsection{\textbf{Main Results}}
Table~\ref{table:exp_main} reports the results on four metrics, two backbone models, and four datasets. 
The main findings are summarized as follows. 
\textbf{\textit{1)}} Distill-R1 consistently performs worse than CoT on all datasets. 
A likely reason is its weaker instruction-following ability~\citep{arxiv25_think_fail}, which makes rule-based answer extraction unreliable, even when an LLM is used as the evaluator. 
Since this issue is orthogonal to our study, we do not investigate it further. 
\textbf{\textit{2)}} H2O reduces memory consumption effectively while preserving the accuracy of Vanilla, suggesting that its greedy eviction strategy works well for long-form generation. 
That said, it comes with a noticeable latency cost: compared with Vanilla, inference time increases by 51\% on Qwen ($(20.61-13.68)/13.68\approx 0.51$) and by 72\% on Llama. 
This overhead is mainly caused by its token-level eviction mechanism, which adds extra computation at every decoding step. 
\textbf{\textit{3)}} SepLLM yields the weakest performance overall. 
During generation, it gradually loses language capability, often failing to produce termination tokens, which in turn leads to much longer inference time. 
\textbf{\textit{4)}} Compared with H2O, LThinker~(\textit{tho.}) maintains similar performance at lower Dep values, indicating a comparable compression ratio, while cutting inference time by 52\% on Qwen and 41\% on Llama on average. 
It also achieves higher accuracy and faster decoding than AnLLM.

These results lead to the following conclusions. 
\textbf{\textit{1)}} BS17K is an effective instruction-tuning dataset for improving reasoning quality. 
Vanilla outperforms both CoT and Distill-R1 on most datasets, suggesting that BS17K helps SFT mitigate the repetition issue observed in Distill-R1. 
\textbf{\textit{2)}} LThinker achieves a favorable trade-off between reasoning quality and inference cost. 
On Qwen, it sacrifices only 1\% accuracy while saving 26\% inference time, reducing Peak by 70\% and Dep by 78\%, corresponding to a 4.5$\times$ compression ratio (16.6/3.7). 
On Llama, it sacrifices 6\% accuracy but saves 1\% inference time, reduces Peak by 70\%, and lowers Dep by 74\%, giving a 3.9$\times$ compression ratio (10.5/2.7). 
\textbf{\textit{3)}} The segmentation strategy plays a critical role in LThinker. 
Thought-level segmentation consistently outperforms token-level segmentation, improving accuracy by 6.2\% on Qwen and 5.6\% on Llama. 
This suggests that token-level segmentation may blur semantic boundaries and thus weaken the quality of compression.

\begin{figure*}[!th] 
    \centering
    \scalebox{0.9}{
    \includegraphics[width=1\linewidth]{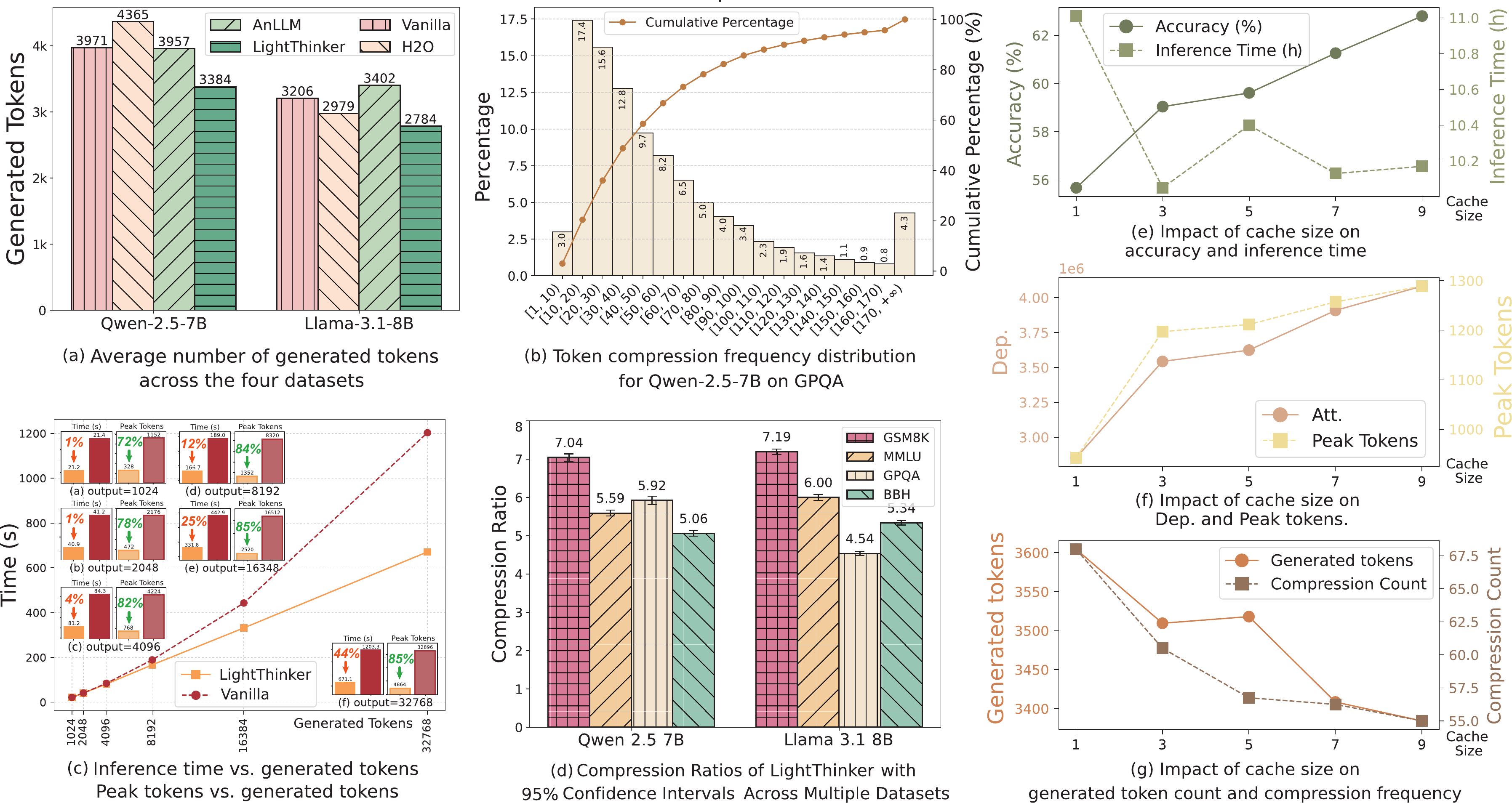} 
    }
    \caption{
    Efficiency Analysis and Ablation Results.
    (a) shows the average number of generated tokens for each model on each dataset. 
    (b) shows the distribution of token lengths across ranges, while the cumulative curve indicates the overall proportion up to each range. 
    (c) illustrates the relationship between output length and inference time, with each subplot reporting inference time and peak token count. 
    (d) reports the average compression ratios, with error bars showing 95\% confidence intervals. 
    (e--f) examine how cache size $|C|$ affects accuracy, Dep, inference time, peak tokens, generated tokens, and compression frequency.
    }
    \label{fig:exp:efficient}
\end{figure*}

\subsubsection{\textbf{Efficiency}}
For readability, we use ``LThinker'' in the following to refer to LThinker~(\textit{tho.}). 
This section examines the efficiency of LThinker from four perspectives.
\input{tables/exp_same_memory}

\paragraph{How does LightThinker~accelerate under same memory budget?}
We measure efficiency in terms of both memory usage and inference speed. 
As shown in Tab.~\ref{table:exp_main}, LThinker can substantially reduce memory consumption at the same batch size. 
In turn, this allows larger batches to be processed under the same memory budget, which can improve throughput in practice. 
Under identical memory constraints, experiments on four datasets with the Qwen model show that LThinker reduces inference time by 2.5$\times$ on average relative to Vanilla, as reported in Tab.~\ref{table:exp:efficiency:same_memory}. 
These results indicate that LThinker not only lowers memory and time costs at a fixed batch size (Tab.~\ref{table:exp_main}), but also brings additional speed gains when the memory budget is held constant.

\paragraph{Does LightThinker~generate more tokens compared to Vanilla?}
Figure~\ref{fig:exp:efficient}(a) compares the average output tokens of H2O, AnLLM, LightThinker, and Vanilla across four datasets, with additional results provided in Appx.~\ref{sec:app:exp:additional_results}. 
Two observations are worth noting:
1) LThinker is the only method that consistently generates fewer tokens than Vanilla, reducing the output length by 15\% on Qwen and 13\% on Llama on average. 
This shorter generation length is one of the main reasons behind its faster decoding speed.
2) H2O shows an inconsistent trend: it increases the number of generated tokens by 10\% on Qwen but decreases it by 7\% on Llama. 
However, even when fewer tokens are generated on Llama, inference time still increases, as shown in Tab.~\ref{table:exp_main}, suggesting that the overhead introduced by its eviction policy accumulates as generation proceeds.

\paragraph{What is the compression ratio of LightThinker?}
The compression behavior of LightThinker is summarized in three views: Fig.~\ref{fig:exp:efficient}(d) presents the compression ratio across four datasets, Tab.~\ref{table:exp:efficiency:comp_count} reports the average number of compressions, and Fig.~\ref{fig:exp:efficient}(b) shows the distribution of compressed token counts on GPQA with Qwen (additional datasets are included in Appx.~\ref{sec:app:exp:additional_results}). 
From these results, we observe that:
1) Compression frequency and compression ratio are influenced more by the task than by the backbone model. 
For example, easier tasks such as GSM8K tend to require fewer compressions and achieve higher ratios, whereas harder tasks such as GPQA involve more frequent compressions and lower ratios. 
2) The compressed token counts exhibit a clear long-tail distribution.

\paragraph{How efficient is LightThinker~in memory usage and inference for long-text generation?}
Figure~\ref{fig:exp:efficient}(c) compares the inference time and peak tokens of LThinker and Vanilla as the output length increases. 
Unless otherwise specified, we use a prompt length of 125 and compress 56 tokens into 8 tokens, corresponding to $|C|=7$. 
The results show two clear trends. 
First, our method yields substantial speedups for long generations. 
When the output length reaches 32K tokens, inference time is reduced by 44\%. 
For shorter outputs between 1K and 4K tokens, the reduction is smaller, but still ranges from 1\% to 4\%. 
Second, LThinker consistently lowers peak token usage, even for short generations. 
For example, peak tokens decrease by 72\% at 1K tokens and by 85\% at 32K tokens.

\subsubsection{\textbf{Ablation}} 
We study two factors that may contribute to LThinker's performance: 1) the decoupled token design with its corresponding attention-mask strategy, and 2) the cache size $|C|$.
\label{ablation}
\paragraph{Decoupled Token and Attention Mask Mode.}
\input{tables/exp_ablation_attention}
Compared with AnLLM, LThinker introduces two differences: a decoupled token design and a different attention mask, as illustrated in Fig.~\ref{fig:comp}. 
To examine their roles, we conduct controlled ablations. 
As reported in Table~\ref{table:exp:ablation:attention}, when the cache size is fixed and LThinker uses AnLLM's attention-mask pattern (``AnLLM'' vs. ``Ours ($|C|=1$, T)''), the decoupled token design alone improves accuracy by 2\%. 
If we further switch to LThinker's attention-mask mode, accuracy increases by another 7\%. 
These results confirm that both components contribute meaningfully to LThinker's performance.

\paragraph{Cache Size.}
We sweep $|C|$ over $\{1,3,5,7,9\}$ to study its effect on accuracy, inference time, dependency (i.e., Dep), peak tokens, generated token count, and compression frequency. 
The corresponding trends on the Qwen model are shown in Fig.~\ref{fig:exp:efficient}(e-g). 
The results suggest the following:
1) As Fig.~\ref{fig:exp:efficient}(e) shows, a larger cache generally improves accuracy while lowering inference time. 
This suggests that increasing cache capacity helps preserve more information after compression. 
2) Fig.~\ref{fig:exp:efficient}(g) shows that a larger cache size reduces both the compression frequency and the number of generated tokens. 
3) Taken together, Fig.~\ref{fig:exp:efficient}(e) and Fig.~\ref{fig:exp:efficient}(g) indicate a clear trade-off: smaller caches trigger more frequent generation and compression to preserve information, whereas larger caches reduce this need.

\subsubsection{\textbf{Case Study}}
\begin{figure}[!htbp] 
    \centering
    \scalebox{1.0}{
    \includegraphics[width=1.0\linewidth]{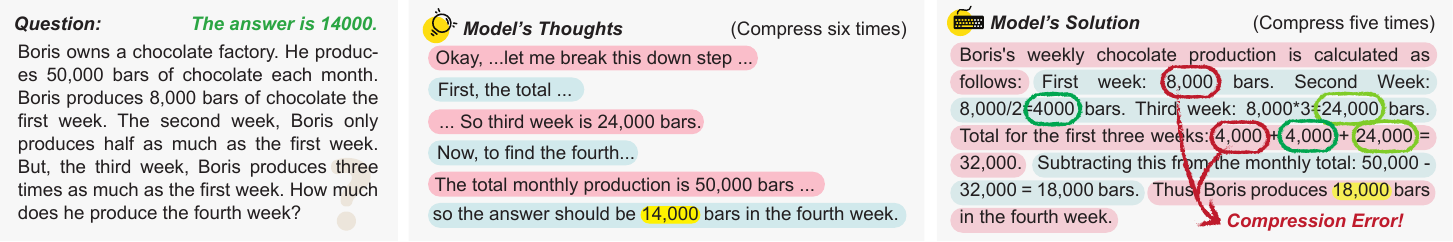} 
    }
    \caption{
    Case study. 
    The figure shows a partial inference trace for one GSM8K example. 
    The full example is provided in App.~\ref{sec:app:exp:additional_results}. 
    Pink and light blue backgrounds indicate alternating compression steps, with each color corresponding to one compression.
    }
    \label{fig:exp:case}
\end{figure} 
Fig.~\ref{fig:exp:case} presents a failure case from GSM8K. 
Although the model reaches the correct answer during intermediate reasoning (see the \texttt{Model's Thoughts} field in Fig.~\ref{fig:exp:case}), the final response is incorrect (see the \texttt{Model's Solution} field). 
In particular, the first ``4000'' in the third sentence of \texttt{Model's Solution} is wrong. 
This suggests that the second compression step discarded part of the necessary numerical information: ideally, ``8000'', ``4000'', and ``24000'' should all have been preserved, but the model retained only ``4000'' and ``24000''. 
As a result, the subsequent reasoning became inconsistent. 
Such failures are common on GSM8K, indicating that the current compression mechanism is still not sufficiently sensitive to numerical details.

\lthink{
\subsection{Evaluation of \textbf{\ours ++.}}
Following the evaluation of LThinker, we evaluate LThinker++ on the same benchmarks and analyze how explicit action-based memory management improves the accuracy--efficiency trade-off.

\input{tables/v2/exp_main_LT++1}

\subsubsection{\textbf{Main Results}}
\label{sec:main_lthink++}
We report results under two serving configurations with the same global context budget (\textit{MaxContext}) but different token allocation policies:
i) \textit{Throughput} (Tab.~\ref{table:exp_main_explicit}): a fixed, small \textit{max\_new\_tokens} per reasoning round to mimic latency-constrained serving;
ii) \textit{Budget} (Tab.~\ref{app:table:exp_main_explicit_budget}): \textit{max\_new\_tokens} is set to the remaining budget, i.e., \textit{MaxContext} minus tokens kept after memory actions (e.g., \textit{commit}, \textit{fold}).
\input{tables/v2/app_exp_main_LT}
As shown in Tab.~\ref{table:exp_main_explicit} and Tab.~\ref{app:table:exp_main_explicit_budget}, our method achieves a better trade-off between reasoning accuracy and memory efficiency. 
We summarize the key observations:

\textbf{1) Overall Performance and Accuracy-Cost Trade-off.}
As shown in Tab.~\ref{table:exp_main_explicit} and Tab.~\ref{app:table:exp_main_explicit_budget}, LThinker++ achieves a superior balance between reasoning accuracy and memory efficiency. 
In the \textit{Throughput} setting, LThinker++ demonstrates extreme resource efficiency by slashing both average Peak and Dep by \textbf{69.9\%} (e.g., 3120 $\rightarrow$ 940 tokens on Qwen2.5-7B) while maintaining comparable accuracy to the Vanilla baseline. 
In the \textit{Budget} setting, LThinker++ prioritizes reasoning depth, yielding a \textbf{+2.42\%} average accuracy gain while still reducing Peak and Dep by \textbf{45.0\%} and \textbf{33.7\%}, respectively. 
We attribute this to a \textit{semantic denoising} effect: by explicitly pruning logical redundancies, LThinker++ maintains a cleaner reasoning context, allowing the model to focus on critical logical anchors rather than being distracted by verbose intermediate steps.

\textbf{2) Efficiency and the Latency-Throughput Balance.}
In the Throughput setting (Tab.~\ref{table:exp_main_explicit}), LThinker++ demonstrates a more stable efficiency ceiling for deployment. 
For Qwen2.5-7B, it slashes Peak memory (3120 $\rightarrow$ 940) and Dep (10.3M $\rightarrow$ 3.1M) with comparable accuracy to Vanilla (60.02 vs. 59.62). 
Notably, while methods like TokenSkip also aim for compression, they often incur a ``latency paradox''—a significantly higher time cost (e.g., 42.7 vs. 21.5 on Qwen2.5-7B). 
This is likely because pruning-style training forces the model to generate more exhaustive token sequences to restore the probabilistic coherence of its Chain-of-Thought, a trend especially pronounced in long-CoT distilled models. 
In contrast, while LThinker++ introduces minor prefill overhead due to multi-round generation, its drastic reduction in KV cache footprint enables a much higher system-level throughput via increased batch sizes.

\textbf{3) Maximizing Reasoning Potential under Strict Budgets.}
When operating under fixed global context constraints (Tab.~\ref{app:table:exp_main_explicit_budget}), LThinker++'s dynamic management proves most effective. 
By actively compressing redundant history, the model ``saves'' budget for subsequent critical reasoning steps. This leads to substantial gains in reasoning-heavy benchmarks like GPQA, where LThinker++ achieves a \textbf{+5.73} accuracy boost on Qwen2.5-7B while utilizing \textbf{40.5\%} less Peak memory. 
These results suggest that for complex, multi-step problems, a condensed and high-signal context is fundamentally more effective than a verbose, unmanaged one.

\subsubsection{\textbf{Efficiency}}
To analyze the underlying mechanisms of LThinker++ and how it manages the trade-off between reasoning accuracy and memory efficiency, we conduct an in-depth diagnostic study.
Our analysis primarily focuses on the Throughput configuration (Fig. \ref{fig:exp:lt++_efficiency}), with corresponding statistics for the Budget setting provided in App. Fig.\ref{fig:exp:lt++_efficiency_budget}.
We focus on the following four questions:
\begin{figure*}[!th] 
    \centering
    \scalebox{1}{
    \includegraphics[width=1\linewidth]{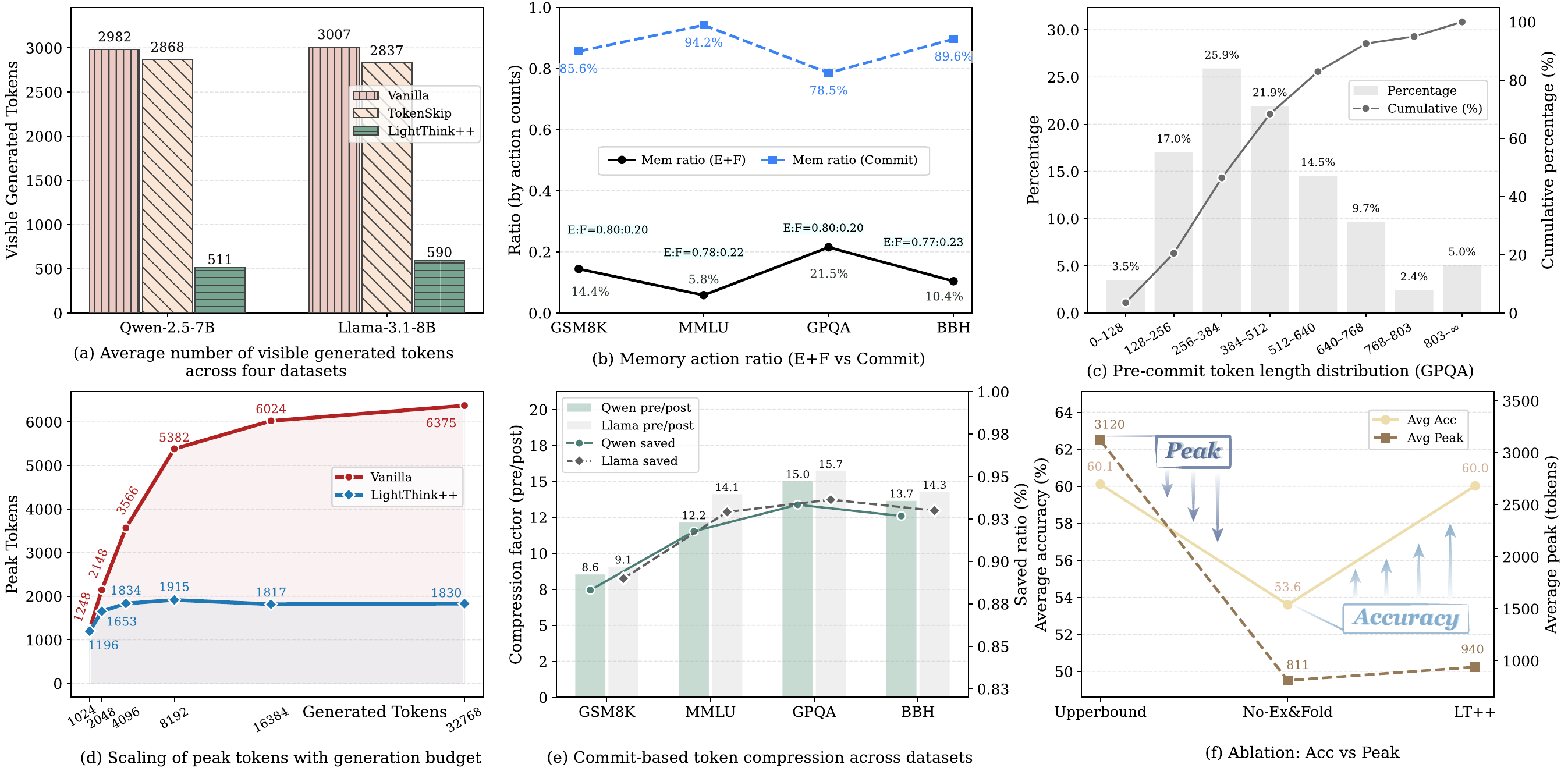} 
    }
    \caption{
    Efficiency Analysis and Ablation Results under the Throughput setting.
    (a) illustrates the average number of generated tokens retained in the current context.
    (b) presents the distribution ratio of memory actions (\textit{Commit} vs. \textit{Expand+Fold}) for Qwen-2.5-7B.
    (c) shows the percentage of pre-commit token lengths falling within specified ranges on GPQA(Qwen-2.5-7B).
    (d) examines the scaling of peak memory tokens as a function of the generation budget.
    (e) summarizes the average compression factors and saved token ratios across different model series.
    (f) provides an ablation study comparing accuracy and average peak tokens between the full LThinker++ and its degraded variants.
    }
    \label{fig:exp:lt++_efficiency}
\end{figure*} 

\paragraph{How much context memory does LightThinker++ save compared to Vanilla?}
Fig.~\ref{fig:exp:lt++_efficiency}(a) illustrates the average number of visible generated tokens per reasoning step, reflecting the actual context window the model attends to. We observe that LThinker++ constrains this window significantly: on Qwen-2.5-7B, visible tokens are reduced by 82.9\% (from 2982 to 511) compared to Vanilla; on Llama-3.1-8B, the reduction is 80.4\% (from 3007 to 590). These results demonstrate that LThinker++ distills redundant reasoning traces into compact semantic representations, reducing the memory footprint while maintaining logical continuity.

\paragraph{How does LightThinker++ adapt its memory-management strategy to task complexity?}
Fig.~\ref{fig:exp:lt++_efficiency}(b) presents the distribution of actions across benchmarks, revealing a pattern of \textbf{cognitive economy}:
1) \emph{Compression-driven efficiency:} \textit{commit} dominates across all tasks, serving as the key operation for context compression and faster inference.  
2) \emph{Strategic adaptivity via Context Refinement:} While \textit{commit} frequency remains high, the model modulates its context refinement actions (\textit{expand} and \textit{fold}) based on task difficulty.
On simpler tasks such as MMLU, the model prioritizes straightforward archiving with minimal refinement (5.8\%).
Conversely, on challenging tasks like GPQA, these refinement actions increase substantially to \textbf{21.5\%}.
This suggests that our method learns to proactively reorganize or retrieve historical details to compensate for potential information loss during complex reasoning, rather than compressing indiscriminately.

\paragraph{What are the characteristics of LightThinker++’s compression granularity and semantic density?}
Fig.~\ref{fig:exp:lt++_efficiency}(c) depicts the distribution of raw reasoning segment lengths immediately preceding each \texttt{commit} on GPQA, while Fig.~\ref{fig:exp:lt++_efficiency}(e) quantifies the resulting compression ratios. 
We observe that:
1) \emph{High-fidelity compression:} LThinker++ achieves a compression ratio of $15.0\times$ on GPQA, and still maintains $8.6\times$ on the simpler GSM8K.  
2) \emph{Task-aware cadence:} The frequency of compression actions scales naturally with task difficulty. Specifically, LThinker++ executes fewer \texttt{commit} operations on simpler tasks like GSM8K (average 3 times per question) compared to more challenging benchmarks such as GPQA (average 7–8 times) and BBH (average 5–7 times).
The distribution in Fig.~\ref{fig:exp:lt++_efficiency}(c) indicates that most \texttt{commit} actions occur after long logical blocks. 
This adaptive cadence aligns compression with logical boundaries, allowing LThinker++ to encode deeper semantics in fewer tokens.

\paragraph{Does LightThinker++ decouple reasoning depth from memory limits?}
Fig.~\ref{fig:exp:lt++_efficiency}(d) examines peak memory usage as the generation budget scales. 
While Vanilla's peak tokens climb linearly, LThinker++ maintains a remarkably flat ceiling, peaking at only \textbf{1,830} (a \textbf{71.3\%} reduction).
This confirms that LThinker++ effectively decouples reasoning depth from physical memory constraints, facilitating long-horizon complex reasoning with a substantially smaller memory footprint. 
Consistent scaling trends under the Budget setting are further detailed in Fig.~\ref{fig:exp:lt++_efficiency_budget}.

\subsubsection{\textbf{Ablation}}
\label{sec:ablation:explicit}
As shown in Fig.~\ref{fig:exp:lt++_efficiency}(f), restricting LThinker++ to a one-way compression mode (\texttt{No-Ex\&Fold}) causes accuracy to plummet from 60.1\% to 53.6\%, proving that irreversible compression leads to critical information loss. 
Restoring the full action set recovers accuracy to 60.0\% while maintaining a \textbf{69.9\% reduction in peak memory} (940 vs. 3120).
Interestingly, LThinker++ achieves higher accuracy with only a marginal increase in peak tokens (940 vs. 811) compared to the degraded variant. We attribute this to a \textbf{reasoning compensation} effect: without explicit retrieval (e.g., \textit{expand}), the model may generate redundant, circular explanations to bridge memory gaps, whereas a full action suite enables \textbf{semantic denoising}. 
By distilling noisy trajectories into curated logical entities, our method allows the model to attend more effectively to core logical nodes, matching the performance of a full-context baseline at a fraction of the resource cost.

\subsubsection{\textbf{Case Study}}
\begin{figure}[!htbp] 
    \centering
    \scalebox{1.0}{
    \includegraphics[width=1.0\linewidth]{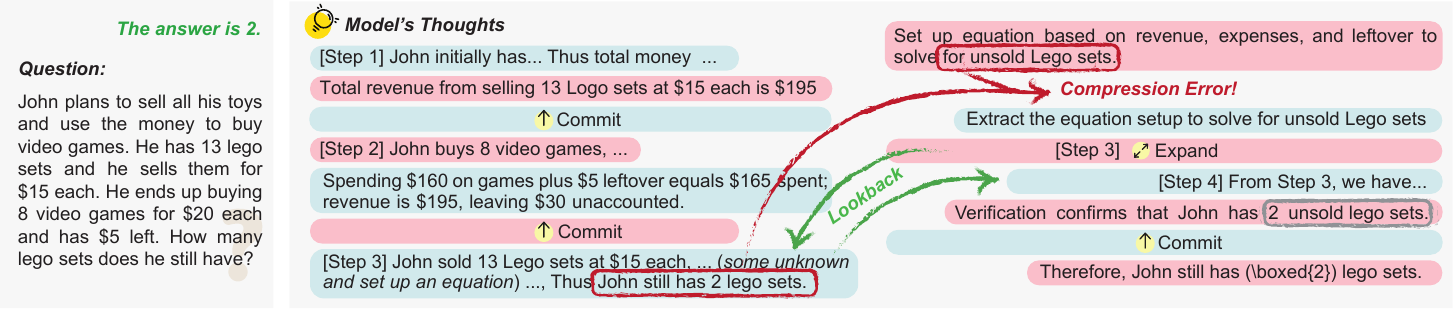} 
    }
    \caption{
    Case Study. 
    The figure illustrates partial inference results of a case of LThinker++ from GSM8K.}
    \label{fig:exp:case_lthink}
\end{figure} 

In our case analysis, we observe two typical uses of memory: (i) verifying previous reasoning states, and (ii) recovering information that is lost during compression. We take the second case as an example. 
As shown in Fig.~\ref{fig:exp:case_lthink}, the model has already inferred the key fact that John has 2 Lego sets, but this information is not reliably retained after compression due to missing intermediate details. 
The model then performs an \textit{expand} action to retrieve the missing context from memory and continue reasoning from a consistent state, demonstrating the effectiveness of our recovery mechanism.

\subsection{Discussions: Implicit vs.\ Explicit}
\label{sec:discussions-implicit-explicit}
\input{tables/v2/exp_comparasion}

\subsubsection{\textbf{Reasoning Performance Comparison}}
We first evaluate implicit and explicit reasoning under the same experimental protocol. 
In particular, to ensure a fair comparison, we keep the training and evaluation setup identical to LThinker and only replace the training traces with the synthetic traces generated by LThinker++.
Within this setting, \texttt{tho1} follows the same preprocessing/segmentation procedure as \texttt{tho}~(in Tab.~\ref{table:exp_main}), whereas \texttt{tho2} additionally introduces a finer-grained thought segmentation strategy using \texttt{\textbackslash n}.

The corresponding results are reported in Table~\ref{table:exp_main_compare}.
The table reveals a clear performance divergence: while $\text{LThinker}_{\text{tho}}$ remains competitive on the original Distilled-R1 traces, the implicit variants degrade more noticeably on our newly synthesized traces generated by \texttt{LThinker++} (Section~\ref{sec:data_synthesis}).
We hypothesize this is due to differences in information density. 
The synthesized structured traces often align each segment with a \textit{self-contained logical step}, concentrating critical anchors within a compact span. 
In contrast, R1-Distill traces contain more discourse connectives and redundant phrasing, which ``dilutes'' information across a softer linguistic flow.

\subsubsection{\textbf{Quantitative Data Analysis: Length Distribution and Logical Density}}
To investigate whether the data characteristics contribute to the above performance degradation, we analyze the character-length distribution of thought segments for the datasets corresponding to the strong implicit baseline in Table~\ref{table:exp_main} (denoted as \texttt{tho}) and our two segmented variants (\texttt{tho1} and \texttt{tho2}).
Figure~\ref{fig:thought-length} reports kernel density estimates (KDE) of per-thought segment lengths.

\begin{wrapfigure}{r}{0.51\textwidth} 
  \centering
  \vspace{-10pt}
  \includegraphics[width=0.51\textwidth]{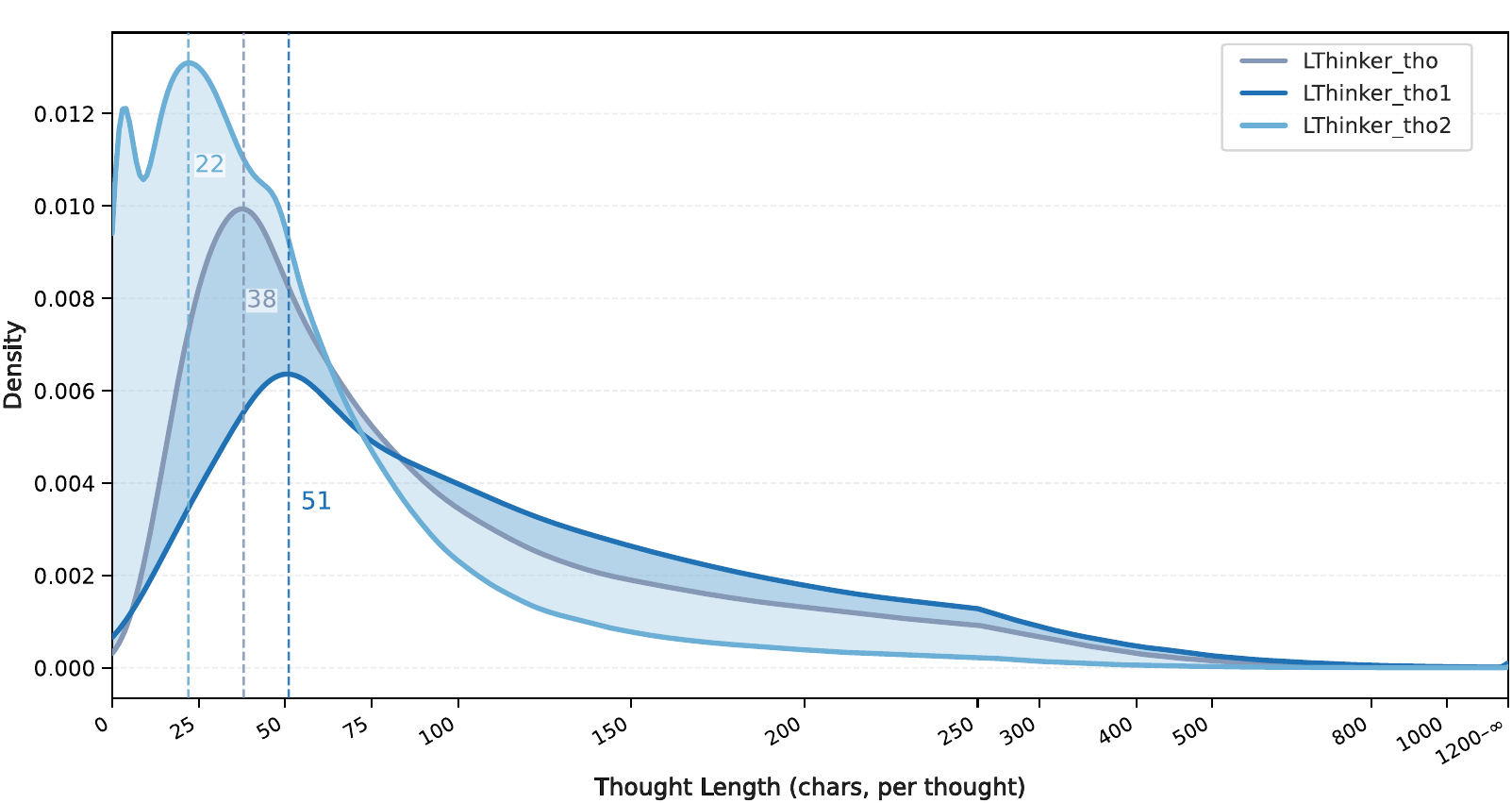}
  \caption{\textbf{Thought segment length distribution.} Kernel density estimates of per-thought segment lengths (in characters) for \texttt{LThinker\_tho}, \texttt{LThinker\_tho1}, and \texttt{LThinker\_tho2}. The x-axis is the segment length (characters), and the y-axis is probability density. Dashed vertical lines indicate the peak locations.}
  \vspace{-10pt}
  \label{fig:thought-length}
\end{wrapfigure}
The original-flow data (\texttt{tho}) peaks at $\approx 38$ characters and is characterized by substantial linguistic redundancy such as connectives and hedging tokens. This redundancy acts as a semantic buffer. Even if implicit compression discards certain local details, the model can often reconstruct the global intent via remaining semantic cues and pretrained language priors.

In contrast, LThinker++ synthesized traces (\texttt{tho1}) exhibit a peak shift to $\approx 51$ characters. This distributional shift reflects our synthesis strategy which explicitly generates segments to encapsulate semantically complete reasoning units. By design, these structured traces prioritize the inclusion of critical logical anchors including subscripts, constants, and variable bindings within each segment. Consequently, \texttt{tho1} possesses a significantly higher information density than the redundant natural language flow observed in \texttt{tho}.

We attribute the degradation of implicit variants to a representation bottleneck when processing such high-density information. Within a constrained latent space, the loss of a single pivotal anchor like a mathematical sign becomes irrecoverable due to the lack of surrounding redundancy. 
This causes errors to propagate and eventually break the reasoning chain.
Furthermore, while \texttt{tho2} reduces per-segment length with a peak at $\approx 22$ characters, the excessive number of boundaries leads to semantic fragmentation. Many segments contain only isolated punctuation or vestigial characters that lack independent logical value. 
This effectively injects structural noise into the context, forcing the model into frequent and error-prone state switching which leads to cumulative reconstruction failures.

\subsubsection{\textbf{Takeaways and Future Directions}}
Overall, fixed-capacity implicit compression appears well-suited to ``soft,'' redundant natural-language reasoning flows, but it is more likely to expose capacity limits when confronted with dense, atomic ``hard-logic'' steps. In contrast, $\text{LThinker++}$ mitigates this issue by explicitly managing key anchors (e.g., via textual summaries or structured records), providing a robust mechanism for state transfer in high-density settings.

These findings also suggest a clear optimization direction for implicit models: rather than merely increasing a static number of Gist tokens, future work should explore \textbf{adaptive latent capacity allocation} (e.g., dynamically adjusting the number or budget of Gist tokens based on estimated segment information density/entropy), thereby improving fidelity and robustness on dense reasoning steps.
}

%% file: tables/exp_main.tex
\begin{table*}[!t]
\centering
\small
\scalebox{0.72}{
\begin{tabular}{l *{4}{p{0.60cm}} *{1}{|p{0.60cm}} *{3}{p{0.60cm}} *{1}{|p{0.60cm}} *{3}{p{0.60cm}} *{1}{|p{0.60cm}} *{3}{p{0.60cm}} *{1}{|p{0.60cm}} *{3}{p{0.60cm}}}
\toprule
\multirow{2}{*}{\textbf{Method}} & \multicolumn{4}{c}{\textbf{GSM8K}} & \multicolumn{4}{c}{\textbf{MMLU}} & \multicolumn{4}{c}{\textbf{GPQA}} & \multicolumn{4}{c}{\textbf{BBH}} & \multicolumn{4}{c}{\textbf{AVG.}} \\ \cmidrule(lr){2-5} \cmidrule(lr){6-9} \cmidrule(lr){10-13} \cmidrule(lr){14-17} \cmidrule(lr){18-21}
 & Acc~$\uparrow$ & Time~$\downarrow$ & Peak~$\downarrow$ & Dep~$\downarrow$ & Acc~$\uparrow$ & Time~$\downarrow$ & Peak~$\downarrow$ & Dep~$\downarrow$ & Acc~$\uparrow$ & Time~$\downarrow$ & Peak~$\downarrow$ & Dep~$\downarrow$ & Acc~$\uparrow$ & Time~$\downarrow$ & Peak~$\downarrow$ & Dep~$\downarrow$ & Acc~$\uparrow$ & Time~$\downarrow$ & Peak~$\downarrow$ & Dep~$\downarrow$ \\ \cmidrule{1-21}
\rowcolor{mygray} \multicolumn{21}{c}{\textit{Qwen2.5-7B Series}} \\ \cmidrule{1-21}
CoT & {86.12} & {1.66} & {513} & {0.1M} & {66.50} & {1.77} & {649} & {0.2M} & {26.76} & {0.60} & {968} & {0.5M} & {65.45} & {0.68} & {570} & {0.1M} & {61.21} & {1.18} & {675} & {0.2M} \\
Distill-R1 & {81.88} & {5.60} & {844} & {1.1M} & {51.70} & {14.31} & {2483} & {7.5M} & {24.75} & {8.01} & {6718} & {31M} & {57.78} & {5.53} & {1967} & {6.0M} & {54.03} & {8.36} & {3003} & {11.3M} \\ \cmidrule{1-21}
Vanilla & {90.90} & {11.83} & {2086} & {3.9M} & {59.98} & {20.61} & {3417} & {10M} & {30.81} & {10.76} & {8055} & {39M} & {69.90} & {11.50} & {3786} & {13M} & {62.90} & {13.68} & {4336} & {16.6M} \\ 
\rowcolor{myblue}
~~+~H2O & \underline{89.92} & {22.19} & \textbf{640} & \underline{1.2M} & \underline{59.69} & {29.02} & {1024} & {3.2M} & {24.75} & {15.61} & \underline{1200} & \underline{9.8M} & \underline{70.10} & {15.61} & \textbf{1024} & \underline{3.5M} & \underline{61.12} & {20.61} & \textbf{972} & \underline{4.4M} \\
\rowcolor{myblue}
~~+~SepLLM & {30.40} & {53.52} & {1024} & {6.9M} & {10.81} & {53.45} & {1024} & {9.0M} & {0.00} & {11.65} & \textbf{1024} & {10M} & {8.08} & {26.64} & {1024} & {9.4M} & {12.32} & {36.32} & \underline{1024} & {8.9M} \\
\rowcolor{myblue}
AnLLM & {78.39} & {15.26} & {789} & {1.6M} & {54.63} & {14.13} & \underline{875} & \underline{2.0M} & {19.70} & {9.14} & {3401} & {11M} & {54.95} & {10.04} & {1303} & {3.8M} & {51.92} & {12.14} & {1592} & {4.6M} \\ \cmidrule{1-21}
\rowcolor{myblue} 
$ \mathrm{LThinker}_{\text{tho}} $ & \textbf{90.14} & \textbf{11.46} & \underline{676} & \textbf{1.0M} & \textbf{60.47} & \textbf{13.09} & {944} & \textbf{1.9M} & \textbf{30.30} & \underline{8.41} & {2385} & \textbf{9.3M} & \textbf{70.30} & \textbf{7.71} & \underline{1151} & \textbf{2.7M} & \textbf{62.80} & \textbf{10.17} & {1289} & \textbf{3.7M} \\ 
\rowcolor{myblue} 
$ \mathrm{LThinker}_{\text{tok}} $ & {87.11} & \underline{11.48} & {1038} & {1.5M} & {57.35} & \underline{13.80} & \textbf{489} & {3.5M} & \underline{28.28} & \textbf{8.26} & {3940} & {18M} & {62.83} & \underline{8.95} & {1884} & {5.6M} & {58.89} & \underline{10.62} & {1838} & {7.2M} \\ \cmidrule{1-21}  \cmidrule{1-21}
\rowcolor{mygray} \multicolumn{21}{c}{\textit{Llama3.1-8B Series}} \\ \cmidrule{1-21}
CoT & {85.14} & {2.15} & {550} & {0.2M} & {65.82} & {2.39} & {736} & {0.3M} & {24.75} & {0.96} & {1231} & {0.9M} & {66.46} & {0.93} & {642} & {0.2M} & {60.54} & {1.61} & {790} & {0.4M} \\
Distill-R1 & {73.62} & {2.58} & {395} & {0.1M} & {53.46} & {2.97} & {582} & {0.8M} & {20.20} & {5.24} & {3972} & {16M} & {61.21} & {0.83} & {380} & {0.2M} & {52.12} & {2.91} & {1332} & {4.4M} \\ \cmidrule{1-21}
Vanilla & {91.43} & {12.06} & {1986} & {3.0M} & {69.62} & {14.82} & {2883} & {6.9M} & {40.91} & {7.98} & {6622} & {26M} & {83.03} & {6.80} & {2793} & {5.9M} & {71.25} & {10.42} & {3571} & {10.5M} \\
\rowcolor{myblue} 
~~+~H2O & \textbf{90.45} & {20.23} & {640} & \underline{1.0M} & \textbf{65.92} & {27.11} & \underline{736} & {1.8M} & {31.81} & {12.55} & {1536} & {7.9M} & \underline{78.99} & {11.43} & {1024} & {2.1M} & \underline{66.79} & {17.83} & \underline{984} & {3.2M} \\
\rowcolor{myblue} 
~~+~SepLLM & {26.25} & {50.05} & {1024} & {5.8M} & {25.12} & {50.11} & {1024} & {7.5M} & {2.53} & {12.62} & \underline{1024} & {10M} & {14.55} & {27.14} & {1024} & {8.5M} & {17.11} & {34.98} & {1024} & {8.0M} \\
\rowcolor{myblue} 
AnLLM & {77.33} & {17.92} & \textbf{589} & {1.1M} & {58.62} & {16.53} & \textbf{589} & \textbf{1.2M} & {31.31} & {7.19} & \textbf{838} & \textbf{3.7M} & {68.89} & {9.79} & \textbf{621} & \textbf{1.6M} & {59.04} & {12.86} & \textbf{659} & \textbf{1.9M} \\ \cmidrule{1-21}
\rowcolor{myblue} 
$ \mathrm{LThinker}_{\text{tho}} $ &  \underline{88.25} & \textbf{12.65} & \underline{629} & \textbf{0.9M} & \underline{63.39} & \textbf{14.88} & {882} & \underline{1.8M} & \textbf{36.36} & \textbf{6.38} & {1796} & \underline{6.4M} & \textbf{79.39} & \underline{7.46} & \underline{911} & \underline{1.9M} & \textbf{66.85} & \textbf{10.34} & {1055} & \underline{2.7M} \\
\rowcolor{myblue} 
$ \mathrm{LThinker}_{\text{tok}} $ & {85.52} & \underline{13.87} & {1104} & {1.7M} & {61.05} & \underline{15.85} & {1538} & {3.3M} & \underline{31.82} & \underline{6.94} & {3150} & {12M} & {74.14} & \textbf{7.43} & {1512} & {2.9M} & {63.13} & \underline{11.02} & {1826} & {4.8M} \\  \bottomrule
\end{tabular}
}

\caption{
Main results of LThinker. 
The CoT is based on the instruction model, while Vanilla, AnLLM, and \ours~are based on Distill-R1. 
The light blue background indicates acceleration methods, with bold representing the best and underline the second best among them.
The Acc of Vanilla serves as the upper bound for Acc of acceleration methods. 
Dep is measured in million, Time in hours, and Peak in counts.
The compression ratio can be roughly estimated by the ratio of Dep between acceleration methods and Vanilla.
See Appendix~\ref{sec:app:metric} for more details.
Note that the results here are based on the same batch size. 
The results under the same memory budget are shown in Table~\ref{table:exp:efficiency:same_memory}.
}
\label{table:exp_main}
\end{table*}

%% file: tables/exp_same_memory.tex

\begin{table}[!htbp]
\centering
\small

\begin{minipage}[t]{0.48\textwidth}
\centering
\scalebox{1}{
\begin{tabular}{cccccc}
\toprule
          & GSM8K & MMLU  & GPQA  & BBH   & AVG   \\ \midrule
Vanilla      & 11.83 & 20.61 & 10.76 & 11.50 & 13.68 \\
LightThinker & \textbf{6.73}  & \textbf{7.44}  & \textbf{3.86}  & \textbf{3.97}  & \textbf{5.50}  \\ \bottomrule
\end{tabular}
}
\caption{Inference time comparison in hours between Vanilla and LThinker on the Qwen model across four datasets under the same memory budget.}
\label{table:exp:efficiency:same_memory}
\end{minipage}\hfill
\begin{minipage}[t]{0.48\textwidth}
\centering
\begin{tabular}{ccccc}
\toprule
      & GSM8K & MMLU & GPQA & BBH \\ \midrule
Qwen  & 20    & 37   & 115  & 48  \\
Llama & 26    & 47   & 139  & 55  \\ \bottomrule
\end{tabular}
\caption{Average compression count per question for LThinker across datasets.}
\label{table:exp:efficiency:comp_count}
\end{minipage}

\vspace{-4mm}
\end{table}

%% file: tables/exp_ablation_attention.tex
\begin{table}[!htbp]
\centering
\small
\scalebox{0.8}{
\begin{tabular}{lccccc}
\toprule
 & \multicolumn{1}{l}{GSM8K} & \multicolumn{1}{l}{MMLU} & \multicolumn{1}{l}{GPQA} & \multicolumn{1}{l}{BBH} & \multicolumn{1}{l}{AVG} \\ \cmidrule{1-6}
AnLLM & \underline{78.39} & 54.63 & 19.70 & 54.95 & 51.92 \\
Ours (|C|=1, T)  & 78.32 & \underline{58.23} & \underline{20.71} & \underline{55.35} & \underline{53.15} \\
Ours (|C|=1, F) & \textbf{80.21} & \textbf{58.23} & \textbf{22.22} & \textbf{62.02} & \textbf{55.67} \\ \bottomrule
\end{tabular}
}
\caption{
Ablation results on Qwen for accuracy across four datasets. 
``T'' refers to AnLLM's attention-mask mechanism, whereas ``F'' refers to LThinker's attention-mask mechanism.
}
\label{table:exp:ablation:attention}
\vspace{-3mm}
\end{table}

%% file: tables/v2/exp_main_LT++1.tex
\begin{table*}[!htbp]
\centering
\small
\scalebox{0.72}{
\begin{tabular}{l *{4}{p{0.60cm}} *{1}{|p{0.60cm}} *{3}{p{0.60cm}} *{1}{|p{0.60cm}} *{3}{p{0.60cm}} *{1}{|p{0.60cm}} *{3}{p{0.60cm}} *{1}{|p{0.60cm}} *{3}{p{0.60cm}}}
\toprule
\multirow{2}{*}{\textbf{Method}} & \multicolumn{4}{c}{\textbf{GSM8K}} & \multicolumn{4}{c}{\textbf{MMLU}} & \multicolumn{4}{c}{\textbf{GPQA}} & \multicolumn{4}{c}{\textbf{BBH}} & \multicolumn{4}{c}{\textbf{AVG.}} \\ \cmidrule(lr){2-5} \cmidrule(lr){6-9} \cmidrule(lr){10-13} \cmidrule(lr){14-17} \cmidrule(lr){18-21}
 & Acc~$\uparrow$ & Time~$\downarrow$ & Peak~$\downarrow$ & Dep~$\downarrow$ & Acc~$\uparrow$ & Time~$\downarrow$ & Peak~$\downarrow$ & Dep~$\downarrow$ & Acc~$\uparrow$ & Time~$\downarrow$ & Peak~$\downarrow$ & Dep~$\downarrow$ & Acc~$\uparrow$ & Time~$\downarrow$ & Peak~$\downarrow$ & Dep~$\downarrow$ & Acc~$\uparrow$ & Time~$\downarrow$ & Peak~$\downarrow$ & Dep~$\downarrow$ \\ \cmidrule{1-21}
\rowcolor{mygray} \multicolumn{21}{c}{\textit{Qwen2.5-7B Series}} \\  \cmidrule{1-21}
CoT & {86.12} & {99.6} & {513} & {0.1M} & {66.50} & {106.2} & {649} & {0.2M} & {26.76} & {36.0} & {968} & {0.5M} & {65.45} & {40.8} & {570} & {0.1M} & {61.21} & {70.8} & {675} & {0.2M}  \\ 
Distill-R1 & {81.88} & {336.0} & {844} & {1.1M} & {51.70} & {858.6} & {2483} & {7.5M} & {24.75} & {480.6} & {6718} & {31M} & {57.78} & {331.8} & {1967} & {6.0M} & {54.03} & {501.6} & {3003} & {11.3M} \\ \cmidrule{1-21}
Vanilla & {87.34} & {\textbf{12.7}} & {812} & {1.1M}
& {\textbf{55.27}} & {\underline{33.1}} & {2682} & {\underline{7.7M}}
& {34.34} & {\textbf{15.3}} & {5781} & {21.9M}
& {\textbf{61.55}} & {\textbf{24.7}} & {3205} & {10.2M}
& {59.62} & {\textbf{21.5}} & {3120} & {10.3M} \\
TokenSkip & {\underline{87.92}} & {47.8} & {775} & {\underline{1.0M}}
& {54.40} & {72.3} & {2831} & {9.1M}
& {\textbf{38.22}} & {17.5} & {5611} & {21.2M}
& {59.87} & {33.2} & {2850} & {9.3M}
& {\textbf{60.10}} & {42.7} & {3017} & {10.2M} \\
\rowcolor{myblue_light} 
LThinker* & {84.94} & {\underline{13.5}} & {\textbf{376}} & {\textbf{0.3M}}
& {52.87} & {41.7} & {\textbf{718}} & {\textbf{1.6M}}
& {24.75} & {20.6} & {\textbf{1298}} & {\textbf{7.3M}}
& {51.85} & {36.5} & {\textbf{851}} & {\underline{2.9M}}
& {53.60} & {28.0} & {\textbf{811}} & {\textbf{3.0M}} \\
\rowcolor{myblue_light} 
LThinker++ & {\textbf{88.32}} & {\textbf{12.7}} & {\underline{408}} & {\textbf{0.3M}}
& {\underline{55.05}} & {\textbf{31.8}} & {\underline{755}} & {\textbf{1.6M}}
& {\underline{35.69}} & {\underline{17.4}} & {\underline{1669}} & {\underline{8.1M}}
& {\underline{61.01}} & {\underline{28.5}} & {\underline{927}} & {\textbf{2.5M}}
& {\underline{60.02}} & {\underline{22.6}} & {\underline{940}} & {\underline{3.1M}} \\
\cmidrule{1-21}
\rowcolor{mygray} \multicolumn{21}{c}{\textit{Llama3.1-8B Series}} \\ \cmidrule{1-21}
CoT & {85.14} & {129.0} & {550} & {0.2M} 
& {65.82} & {143.4} & {736} & {0.3M} 
& {24.75} & {57.6} & {1231} & {0.9M} 
& {66.46} & {55.8} & {642} & {0.2M} 
& {60.54} & {96.9} & {790} & {0.4M} \\
Distill-R1 & {73.62} & {154.8} & {395} & {0.1M} & {53.46} & {178.2} & {582} & {0.8M} & {20.20} & {314.4} & {3972} & {16M} & {61.21} & {49.8} & {380} & {0.2M} & {52.12} & {174.6} & {1332} & {4.4M} \\ \cmidrule{1-21} \cmidrule{1-21}
Vanilla & {\textbf{82.79}} & {16.1} & {811} & {1.3M}
& {\underline{61.15}} & {45.8} & {2570} & {7.1M}
& {\underline{30.30}} & {22.0} & {6364} & {25.7M}
& {67.68} & {28.5} & {2826} & {8.8M}
& {\underline{60.48}} & {28.1} & {3143} & {10.7M} \\
TokenSkip & {79.40} & {54.1} & {838} & {1.2M}
& {57.06} & {77.9} & {2499} & {6.9M}
& {26.60} & {22.7} & {6016} & {23.8M}
& {\textbf{69.16}} & {34.2} & {2581} & {7.6M}
& {58.06} & {47.2} & {2984} & {9.9M} \\
\rowcolor{myblue_light} 
LThinker* & {75.54} & {\textbf{12.5}} & {\textbf{357}} & {\textbf{0.2M}}
& {56.80} & {\underline{35.2}} & {\textbf{782}} & {\textbf{1.3M}}
& {21.04} & {\textbf{15.6}} & {\textbf{1275}} & {\textbf{4.6M}}
& {58.79} & {\textbf{21.2}} & {\textbf{734}} & {\textbf{1.1M}}
& {53.04} & {\textbf{21.1}} & {\textbf{787}} & {\textbf{1.8M}} \\
\rowcolor{myblue_light} 
LThinker++ & {\underline{82.23}} & {\underline{13.3}} & {\underline{424}} & {\underline{0.3M}}
& {\textbf{61.77}} & {\textbf{31.4}} & {\underline{883}} & {\underline{1.4M}}
& {\textbf{33.16}} & {\underline{18.6}} & {\underline{1793}} & {\underline{7.2M}}
& {\underline{69.09}} & {\underline{21.9}} & {\underline{896}} & {\underline{1.7M}}
& {\textbf{61.56}} & {\underline{21.3}} & {\underline{999}} & {\underline{2.7M}} \\
\bottomrule
\end{tabular}
}

\caption{
Main results of LThinker++ (\textit{Throughput setting}). 
Time reports the end-to-end wall-clock time in minutes (see Appx.~\ref{sec:app:exp:timing_protocol} for more).
The CoT is based on the instruction model, while the other methods are compared within each corresponding model series.
In this table, LThinker++ denotes our full method, and LThinker* denotes our variant with only the commit action enabled.
}
\label{table:exp_main_explicit}
\end{table*}

%% file: tables/v2/app_exp_main_LT.tex
\begin{table*}[!t]
\centering
\small
\scalebox{0.85}{
\begin{tabular}{l *{3}{p{0.70cm}} *{1}{|p{0.70cm}} *{2}{p{0.70cm}} *{1}{|p{0.70cm}} *{2}{p{0.70cm}} *{1}{|p{0.70cm}} *{2}{p{0.70cm}} *{1}{|p{0.70cm}} *{2}{p{0.70cm}}}
\toprule
\multirow{2}{*}{\textbf{Method}} &
\multicolumn{3}{c}{\textbf{GSM8K}} &
\multicolumn{3}{c}{\textbf{MMLU}} &
\multicolumn{3}{c}{\textbf{GPQA}} &
\multicolumn{3}{c}{\textbf{BBH}} &
\multicolumn{3}{c}{\textbf{AVG.}} \\
\cmidrule(lr){2-4} \cmidrule(lr){5-7} \cmidrule(lr){8-10} \cmidrule(lr){11-13} \cmidrule(lr){14-16}
 & Acc~$\uparrow$ & Peak~$\downarrow$ & Dep~$\downarrow$
 & Acc~$\uparrow$ & Peak~$\downarrow$ & Dep~$\downarrow$
 & Acc~$\uparrow$ & Peak~$\downarrow$ & Dep~$\downarrow$
 & Acc~$\uparrow$ & Peak~$\downarrow$ & Dep~$\downarrow$
 & Acc~$\uparrow$ & Peak~$\downarrow$ & Dep~$\downarrow$ \\
\cmidrule{1-16}


\rowcolor{mygray} \multicolumn{16}{c}{\textit{Qwen2.5-7B Series}} \\ \cmidrule{1-16}
CoT     & {86.12} & {513} & {0.1M} & {66.50} & {649}  & {0.2M} & {26.76} & {968}  & {0.5M} & {65.45} & {570}  & {0.1M} & {61.21} & {675}  & {0.2M}  \\
Distill-R1 & {81.88} & {844} & {1.1M} & {51.70} & {2483} & {7.5M} & {24.75} & {6718} & {31M} & {57.78} & {1967} & {6.0M} & {54.03} & {3003} & {11.3M} \\ \cmidrule{1-16}
Vanilla & {87.62} & {782} & {1.1M} & {57.58} & {2432} & {6.5M} & {32.49} & {5493} & {\underline{20.0M}} & {\underline{62.76}} & {2714} & {8.2M} & {\underline{60.11}} & {2855} & {8.9M} \\
TokenSkip & {\textbf{87.92}} & {775} & {1.0M} & {54.40} & {2831} & {9.1M} & {\textbf{38.22}} & {5611} & {21.2M} & {59.87} & {2850} & {9.3M} & {60.10} & {3017} & {10.2M} \\
\rowcolor{myblue_lightv1}
LThinker*  & {84.61} & {\textbf{444}} & {\underline{0.7M}} & {\underline{57.68}} & {\underline{1141}} & {\underline{4.1M}} & {28.96} & {\underline{3496}} & {23.2M} & {56.63} & {\underline{1580}} & {\underline{7.0M}} & {56.97} & {\underline{1665}} & {\underline{8.7M}} \\
\rowcolor{myblue_lightv1}
LThinker++ & {\underline{87.87}} & {\underline{471}} & {\textbf{0.4M}} & {\textbf{60.27}} & {\textbf{1133}} & {\textbf{3.0M}} & {\textbf{38.22}} & {\textbf{3265}} & {\textbf{16.1M}} & {\textbf{63.77}} & {\textbf{1415}} & {\textbf{4.2M}} & {\textbf{62.53}} & {\textbf{1571}} & {\textbf{5.9M}} \\

\cmidrule{1-16}
\rowcolor{mygray} \multicolumn{16}{c}{\textit{Llama3.1-8B Series}} \\ \cmidrule{1-16}
CoT & {85.14} & {550} & {0.2M} & {65.82} & {736} & {0.3M} & {24.75} & {1231} & {0.9M} & {66.46} & {642} & {0.2M} & {60.54} & {790} & {0.4M} \\
Distill-R1 & {73.62} & {395} & {0.1M} & {53.46} & {582} & {0.8M} & {20.20} & {3972} & {16M} & {61.21} & {380} & {0.2M} & {52.12} & {1332} & {4.4M} \\ \cmidrule{1-16}
Vanilla & \underline{79.38} & {740} & \underline{0.9M} & \textbf{59.82} & {2111} & \underline{4.9M} & \underline{31.82} & {5773} & \textbf{21.7M} & {67.95} & {2606} & \underline{7.5M} & \underline{59.74} & {2808} & \underline{8.8M} \\
TokenSkip & {\textbf{79.40}} & {838} & {1.2M} & {57.06} & {2499} & {6.9M} & {26.60} & {6016} & {23.8M} & \underline{69.16} & \underline{2581} & {7.6M} & {58.06} & {2984} & {9.9M} \\
\rowcolor{myblue_lightv1}
LThinker*  & {76.72} & \textbf{522} & \underline{0.9M} & {59.59} & \underline{1394} & {6.5M} & {28.28} & \underline{4271} & {34.5M} & {63.10} & {1664} & {8.1M} & {56.92} & \underline{1963} & {12.5M} \\
\rowcolor{myblue_lightv1}
LThinker++ & {77.69} & \underline{528} & \textbf{0.7M} & \underline{59.72} & \textbf{1370} & {\textbf{4.5M}} & 
\textbf{33.67} & \textbf{4101} & \underline{22.9M} 
& \textbf{73.20} & \textbf{1481} & \textbf{4.6M} & \textbf{61.07} & \textbf{1870} & \textbf{8.2M} \\
\bottomrule
\end{tabular}
}

\caption{
Main results of LThinker++ (Budget setting). 
Note that the results here are based on the same batch size. 
Other settings and metric definitions are the same as Table~\ref{table:exp_main_explicit}.
}
\label{app:table:exp_main_explicit_budget}
\end{table*}

%% file: tables/v2/exp_comparasion.tex
\begin{table*}[!t]
\centering
\small
\scalebox{0.9}{
\begin{tabular}{l *{3}{p{0.60cm}} *{1}{|p{0.60cm}} *{2}{p{0.60cm}} *{1}{|p{0.60cm}} *{2}{p{0.60cm}} *{1}{|p{0.60cm}} *{2}{p{0.60cm}} *{1}{|p{0.60cm}} *{2}{p{0.60cm}}}
\toprule
\multirow{2}{*}{\textbf{Method}} 
& \multicolumn{3}{c}{\textbf{GSM8K}} 
& \multicolumn{3}{c}{\textbf{MMLU}} 
& \multicolumn{3}{c}{\textbf{GPQA}} 
& \multicolumn{3}{c}{\textbf{BBH}} 
& \multicolumn{3}{c}{\textbf{AVG.}} \\
\cmidrule(lr){2-4} \cmidrule(lr){5-7} \cmidrule(lr){8-10} \cmidrule(lr){11-13} \cmidrule(lr){14-16}
& Acc~$\uparrow$ & Peak~$\downarrow$ & Dep~$\downarrow$
& Acc~$\uparrow$ & Peak~$\downarrow$ & Dep~$\downarrow$
& Acc~$\uparrow$ & Peak~$\downarrow$ & Dep~$\downarrow$
& Acc~$\uparrow$ & Peak~$\downarrow$ & Dep~$\downarrow$
& Acc~$\uparrow$ & Peak~$\downarrow$ & Dep~$\downarrow$ \\
\cmidrule{1-16}
\rowcolor{mygray} \multicolumn{16}{c}{\textit{Qwen2.5-7B Series}} \\  
CoT & {86.12}  & {513} & {0.1M} 
& {66.50}  & {649} & {0.2M} 
& {26.76}  & {968} & {0.5M}
& {65.45}  & {570} & {0.1M}
& {61.21} & {675} & {0.2M}  \\ 
Distill-R1 & {81.88}   & {844} & {1.1M} 
& {51.70}   & {2483} & {7.5M}
& {24.75}   & {6718} & {31M} 
& {57.78}  & {1967} & {6.0M} 
& {54.03}   & {3003} & {11.3M} \\
\cmidrule{1-16}
\rowcolor{myblue} 
$ \mathrm{LThinker}_{\text{tho1}} $
& {84.61}  & {1043} & {2.1M} 
& {52.96}  & {2792} & {8.4M} 
& {18.69}  & {6781} & {29.7M} 
& {53.54}  & {3687} & {13.8M}
& {52.45}  & {3576} & {13.5M} \\
\rowcolor{myblue} 
$ \mathrm{LThinker}_{\text{tho2}} $ 
& {85.75}  & {1003} & {1.8M} 
& {52.00}  & {2822} & {8.23M} 
& {24.24}  & {7042} & {30.8M} 
& {51.31}  & {4130} & {16.3M}
& {53.33}  & {3749} & {14.3M} \\
\rowcolor{myblue_light} 
LThinker++ & {87.87}  & {471} & {0.4M} 
& {60.27}   & {1133} & {3.0M} 
& {38.22} & {3265} & {16.1M} 
& {63.77}   & {1415} & {4.2M}
& {62.53}  & {1571} & {5.9M} \\
\rowcolor{myblue_light} 
LThinker* & {84.61}  & {444} & {0.7M} 
& {57.68}   & {1141} & {4.1M} 
& {28.96}   & {3496} & {23.2M} 
& {56.63}  & {1580} & {7.0M}
& {56.97}  & {1665} & {8.7M} \\
\cmidrule{1-16}
\rowcolor{mygray} \multicolumn{16}{c}{\textit{Llama3.1-8B Series}} \\ 
\cmidrule{1-16}
CoT & {85.14}   & {550} & {0.2M}
& {65.82}   & {736} & {0.3M}
& {24.75}   & {1231} & {0.9M}
& {66.46}   & {642} & {0.2M} 
& {60.54}   & {790} & {0.4M} \\
Distill-R1 & {73.62}   & {395} & {0.1M} 
& {53.46}   & {582} & {0.8M}
& {20.20}  & {3972} & {16M}
& {61.21}   & {380} & {0.2M} 
& {52.12}   & {1332} & {4.4M} \\ 
\cmidrule{1-16} 
\rowcolor{myblue} 
$\mathrm{LThinker}_{\text{tho1}} $ 
& {76.72}  & {917} & {1.5M} 
& {53.26}  & {2682} & {7.6M} 
& {20.70}  & {6697} & {28.8M} 
& {57.58}  & {3674} & {13.3M}
& {52.01}  & {3493} & {12.8M} \\
\rowcolor{myblue} 
$\mathrm{LThinker}_{\text{tho2}} $ 
& {78.92}  & {842} & {1.2M} 
& {52.68}  & {2783} & {7.7M} 
& {18.18}  & {6420} & {27.1M} 
& {56.57}  & {4123} & {15.7M}
& {51.59}  & {3542} & {12.9M} \\
\rowcolor{myblue_light} 
LThinker++ & {77.69}   & {528} & {0.7M} 
& {59.72}   & {1370} & {4.5M} 
& {33.67}   & {4101} & {22.9M} 
& {73.20}   & {1481} & {4.6M}
& {61.07}   & {1870} & {8.2M} \\
\rowcolor{myblue_light} 
LThinker* & {76.72}   & {522} & {0.9M} 
& {59.59}  & {1394} & {6.5M} 
& {28.28}  & {4271} & {34.5M} 
& {63.10}   & {1664} & {8.1M}
& {56.92}  & {1963} & {12.5M} \\
\bottomrule
\end{tabular}
}

\caption{
Comparison between LThinker and LThinker++ on four benchmarks.
For a fair comparison, $\mathrm{LThinker}_{\text{tho1}}$ and $\mathrm{LThinker}_{\text{tho2}}$ are trained on the same training data as in Table~\ref{table:exp_main_explicit}.
}

\label{table:exp_main_compare}
\end{table*}

%% file: sections/v2/method_agent.tex
\begin{figure}[!t]
    \centering
    \scalebox{1.0}{
    \includegraphics[width=1.0\linewidth]{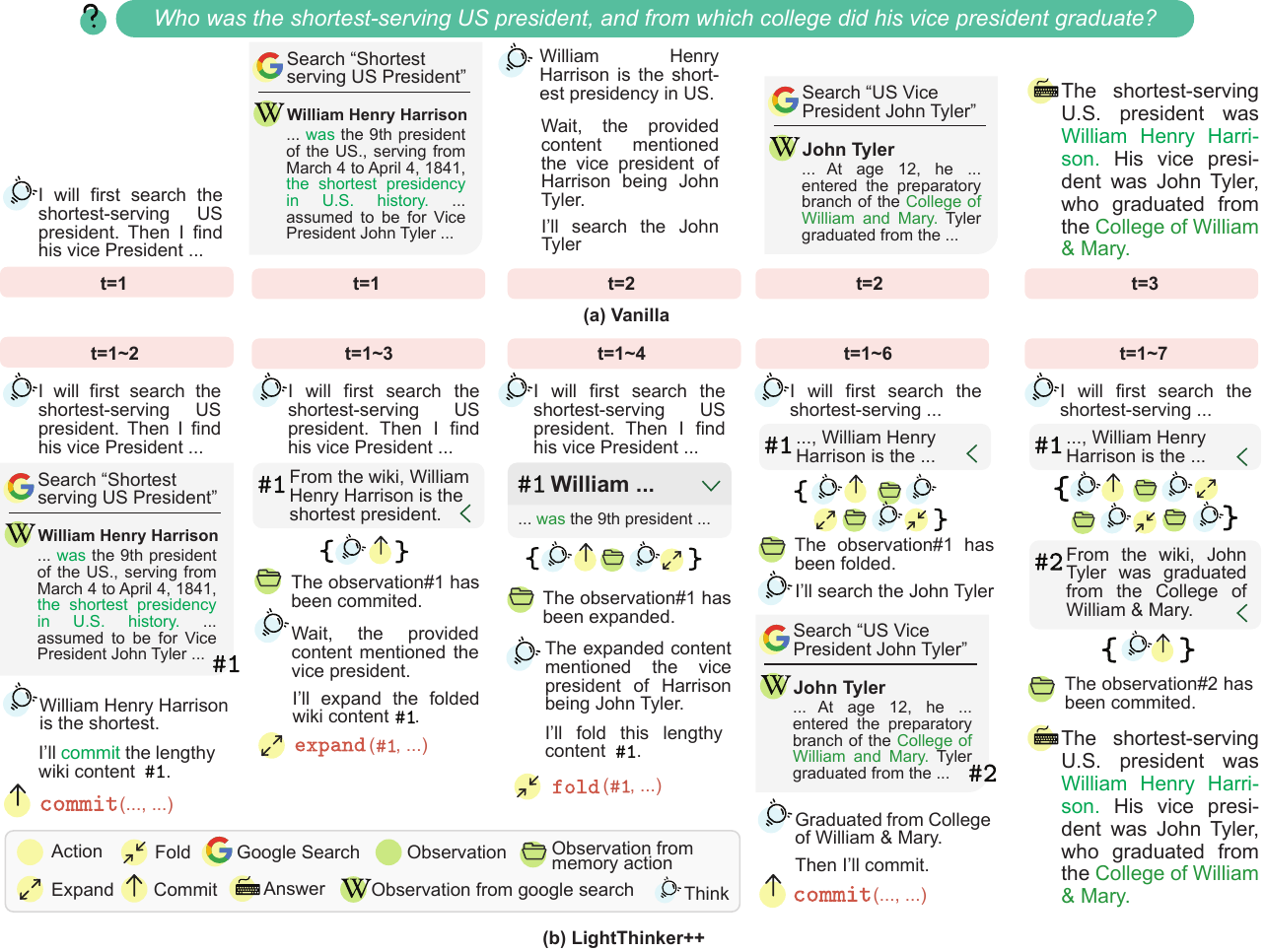} 
    }
    \caption{\textbf{Overview of \ours++ for long-horizon agentic reasoning.}
    \ours++ follows the Thought--Action--Observation loop while explicitly managing a stateful memory.
    Each interaction step can be \textit{committed} into an archived summary, \textit{expanded} back into the active context to recover raw evidence when needed, and \textit{folded} again to maintain context hygiene.}
    \label{fig:method:agent}
\end{figure} 

\lthink{

\section{\ours++: Long-Horizon Agentic Reasoning}
\label{sec:agent}
To demonstrate the robustness of explicit memory management under extreme context demands, we extend \textbf{LightThinker++} to the DeepResearch domain (Fig.~\ref{fig:method:agent}). 
While our evaluation focuses on this scenario, which is characterized by high-entropy web interactions and multi-step information synthesis, the LightThinker++ framework is fundamentally domain-agnostic.
It serves as a general-purpose paradigm for any long-horizon task where an agent must reconcile iterative planning with voluminous external feedback.

\paragraph{\textbf{Formalizing LightThinker++ within the TAO Paradigm.}}
Standard agentic interaction typically follows the \textit{Thought-Action-Observation} (TAO) cycle. 
To bridge this with the framework in Sec.~\ref{sec:method:explicit}, we instantiate the reasoning entity $\mathcal{I}_k = (R_k, Z_k)$ as the $k$-th TAO turn. 
Here, $R_k = (T_k, A_k, O_k)$ captures the full interaction, while $Z_k$ represents its distilled semantic core. 

To counteract the \textbf{context rot} inherent in linear histories, LightThinker++ redefines the research log as a stateful, managed memory $\tilde{\mathcal{H}}_t = \{m_1^{(t)}, \dots, m_K^{(t)}\}$. The visibility of each historical turn $k$ remains governed by the state $\sigma_k^{(t)}$, which triggers the same lossy-to-lossless toggling mechanism:
\begin{equation}
m_k^{(t)} = 
\begin{cases} 
Z_k, & \text{if } \sigma_k^{(t)} = \texttt{archive} \\
R_k, & \text{if } \sigma_k^{(t)} = \texttt{active}
\end{cases}
\label{equ:agent_projection}
\end{equation}
By treating memory management as an explicit decision, LightThinker++ ensures the context window is prioritized for high-signal content. High-entropy evidentiary details are only restored via \textit{expand} when required for synthesis and are promptly \textit{folded} to maintain context hygiene.

\paragraph{\textbf{Adapted Memory Actions for Deep Research.}}
To manage the high informational density inherent in web-based research, we instantiate the memory framework introduced in Sec.~\ref{sec:method:explicit} through a specialized toolset tailored for the agentic interaction loop:
\begin{itemize}
\item \textbf{Environment Actions ($\mathcal{A}_{\text{env}}$):} High-throughput tools such as \textit{search} and \textit{visit}, optimized for rapid information discovery across the open web.
\item \textbf{Memory Actions ($\mathcal{A}_{\text{mem}}$):} Operators governing the contextual lifecycle. Specifically, \textit{commit} distills the holistic interaction $\mathcal{I}_k$ into a summary $Z_k$ to preserve long-term coherence; \textit{expand} re-activates a past step to retrieve raw evidence $R_k$ for precise synthesis; and \textit{fold} purges these details once their utility is exhausted.
\end{itemize}

\paragraph{\textbf{Trajectory Synthesis via Multi-Agent Orchestration.}}
To generate expert-level trajectories for research tasks, we extend the Environment-Aware Trajectory Synthesis framework (Sec.~\ref{sec:data_synthesis}) into a Multi-Agent Orchestration paradigm. 
While the single-model synthesis described in Sec.~\ref{sec:data_synthesis} suffices for standard reasoning, the high informational density of web-based research imposes a heavy cognitive load, requiring a model to reconcile long-term strategic planning with high-entropy external data.

To maintain synthesis quality, we partition this labor between two specialized roles within the closed-loop environment: 
(1) \textbf{Interaction Agent:} Executes $\mathcal{A}_{\text{env}}$ tools while performing granular memory retrieval via \textit{expand} and \textit{fold} to navigate raw information.
(2) \textbf{Contextual Governor:} Acts as the curator of the research log, adaptively triggering \textit{commit} to distill holistic interaction steps into actionable summaries $Z_k$.
By decoupling the generation of execution-level details from high-level context management, this collaborative synthesis ensures that the resulting trajectories maintain both rigorous logical depth and optimized context density.

\paragraph{\textbf{Behavioral Pruning and Memory Lifecycle}.}
To extract the most effective reasoning patterns from the multi-agent orchestration, we implement a Behavioral Pruning mechanism tailored for the high-entropy research domain. We filter the synthesized trajectories against the Memory Lifecycle constraint established in Sec.~\ref{sec:method:explicit}, ensuring the data reflects active context governance rather than passive logging of web interactions. 
A trajectory is deemed admissible only if it satisfies the following criteria:
1) \underline{\textit{Lifecycle Completeness}}: It demonstrates the full cycle, including archiving via \textit{commit} and evidence retrieval via \textit{expand}/\textit{fold};
2) \underline{\textit{Symmetry Constraint}}: Reflecting the need for context hygiene in long-horizon interactions, a \textit{fold} operation must strictly revert a previously expanded step, ensuring the context window remains purged of raw snippets once synthesis is complete.
3) \underline{\textit{Anti-Jitter Heuristics}}: We prohibit redundant memory operations including consecutive actions on the same step and operations targeting non-existent step IDs to ensure management is purposeful. This is particularly crucial in the agentic loop to prevent the model from falling into stochastic retrieval patterns when faced with complex external observations.

Through this rigorous filtering, the collaborative expertise of the multi-agent system is distilled into a single, cohesive policy. We fine-tune the model by minimizing the negative log-likelihood over the pruned expert trajectories $\mathcal{T}^*$:

\label{equ:react_loss}
\begin{equation}
\mathcal{L}_{\mathrm{SFT}}(\theta) = -\mathbb{E}_{\mathcal{T}^*\sim\mathcal{D}} \left[ \sum_{k=1}^{n} \log \pi_{\theta}(T_k, A_k \mid X, \tilde{\mathcal{H}}_k) \right]
\end{equation}

By optimizing the joint predictive likelihood of reasoning traces and memory operations, the agent internalizes \textit{explicit context engineering} as a core component of its decision-making process. 
This high-density learning signal allows the model to maintain context hygiene and reasoning fidelity across extended interaction horizons.
} 

%% file: sections/v2/experiment.tex
\section{Experiments: Long-Horizon Agentic Reasoning}
\subsection{Experimental Settings}
\textbf{Dataset Construction and Filtering.}
The base query pool is curated from a diversified ensemble of sources, including \textit{HotpotQA~\cite{yang2018hotpotqa}, MuSiQue~\cite{trivedi2022musique}, WebDancer~\cite{wu2025webdancer}, WebShaper~\cite{tao2025webshaper}}, and \textit{WebWalkerQA-Silver~\cite{wu2025webwalker}}. To ensure the necessity of multi-hop reasoning and high-order planning, we perform heuristic filtering on \textit{HotpotQA} and \textit{MuSiQue} by selecting only those instances where Qwen3-30B-A3B-Instruct-2507 fails to yield direct solutions. Regarding the \textit{WebWalkerQA-Silver} corpus, we adopted a language-specific selection policy: the English subset was fully incorporated to maintain linguistic diversity, while the Chinese subset was filtered to include only those instances explicitly categorized as ``hard'' according to the dataset's intrinsic difficulty metadata. This collection is further augmented with diversified web-navigation tasks to form the final query pool, with a detailed categorical breakdown provided in Appx.~\ref{app:sec:agenttraindata}.

From this pool, we employ a hierarchical filtering pipeline to construct two distinct training sets: 1) \textbf{Vanilla Baseline}: We synthesize reasoning trajectories using DeepSeek-V3.2 in non-thinking mode, utilizing only the Aenv tool. 
After filtering for correctness, we retain 6,625 high-quality standard trajectories. 
2) \textbf{LThinker++}: We augment the correctness filter with the Behavioral Pruning constraints described in Sec.~\ref{sec:agent}. 
In this configuration, the agent utilizes both the Aenv and Amem tools. 
Due to the stringent requirements for logical memory transitions, only 3,677 expert trajectories were retained.
Despite this 44.5\% reduction in base trajectories compared to the Vanilla, these trajectories were decomposed into 42,633 fine-grained training instances. 
This yields a more potent and logically dense learning signal, providing the model with the necessary supervision to maintain context hygiene and reasoning fidelity in context-heavy tasks.

\textbf{Baselines and Training.}
We evaluate our framework against several state-of-the-art LLMs, including GLM-4.6~\cite{glm46}, Claude-4-Sonnet~\cite{claude}, GPT-5~\cite{gpt5},  Kimi-K2~\cite{team2025kimik2} and  Qwen3-235B-A22B-Instruct~\cite{qwen3} and the DeepSeek-V3 series (V3.1 and V3.2). To assess the specific impact of explicit memory management, we develop and evaluate two internal variants initialized from Qwen3-30B-A3B-Thinking-2507~\cite{qwen3}. 
The first, \textbf{Vanilla-Agent}, is fine-tuned on the \textit{Vanilla Baseline} dataset to equip the model with environment-level capabilities via $\mathcal{A}_{\text{env}}$, serving as a robust standard agentic baseline. 
The second, \textbf{LThinker++}, is trained on our pruned expert trajectories to internalize the joint policy of reasoning and memory orchestration defined in Eq.~\ref{equ:react_loss}.
Implementation details are provided in Appx.~\ref{app:sec:agentic_implementation}.

\textbf{Evaluation Metrics and Datasets.}
We benchmark our models across three representative agentic datasets: xbench-DeepSearch-2510~\cite{xbench}, BrowseComp-EN~\cite{bc_en}, and BrowseComp-ZH~\cite{bc_zh}, which we refer to as xbench, BC-EN, and BC-ZH for brevity.
Performance is quantified by the average Pass@1 score across all test samples and the Pass@3 score over three independent rollouts to assess reasoning stability.
Specifically, we employ gpt-5-2025-08-07 as the primary automated judge to evaluate the semantic alignment between model predictions and ground-truth answers.
To facilitate autonomous web interaction, we implement two core functional tools: 1) Search: An interface with the Google Search API that supports concurrent queries and retrieves the top-10 results per query. 
2)Visit: A navigation module that utilizes Jina~\cite{jina} for HTML parsing and Qwen-Flash to distill task-relevant evidence from the extracted content.

\subsection{Main Results}

\input{tables/v2/exp_main_LT++2}

Table~\ref{table:exp_main_final} summarizes the results across three benchmarks, illustrating the incremental gains from our data synthesis and memory orchestration:

\textbf{Effectiveness of Standard Synthesis.}
The \textit{Vanilla-Agent}, trained on standard trajectories, exhibits a substantial performance leap over the base Qwen3-Thinking model. For instance, Pass@1 scores rise from 8.7\% to 38.3\% on \textit{xbench} and from 10.0\% to 31.5\% on \textit{BrowseComp-ZH}. 
This improvement confirms that our base data pipeline effectively equips the model with fundamental environment-level execution ($\mathcal{A}_{\text{env}}$) and basic research planning capabilities.

\textbf{Superiority of Orchestrated Memory Actions.}
Building upon this baseline, LThinker++ achieves further performance leaps across all benchmarks by internalizing memory management actions ($\mathcal{A}_{\text{mem}}$) via multi-agent orchestration. Compared to the \textit{Vanilla-Agent}, LThinker++ delivers a significant Pass@1 improvement of \textbf{5.7\%} on \textit{xbench}. 
More importantly, our method demonstrates consistent gains in reasoning stability, with Pass@3 scores increasing across all three benchmarks by \textbf{7.0\%}, \textbf{9.3\%}, and \textbf{4.2\%} on \textit{xbench}, \textit{BrowseComp-ZH}, and \textit{BrowseComp-EN}, respectively. 
These gains demonstrate that orchestrated memory primitives allow the agent to maintain a high signal-to-noise ratio, preventing reasoning collapse in long-horizon tasks where standard agents typically struggle with context clutter.

\paragraph{\textbf{Performance Gain on Hard Instances.}}
To better isolate the effect of our approach on robustness, we bucket examples by the \textit{Vanilla-Agent}'s success count over three runs, \(k \in \{0,1,2,3\}\).
We define the \textbf{hard subset} (\(hard_{01}\)) as instances where the baseline succeeds at most once (\(k \le 1\)). 
Importantly, this subset captures not only intrinsically difficult tasks, but also a characteristic \textit{failure mode} of standard SFT agents in long-horizon web research: performance becomes highly unstable across runs due to goal drift, accumulation of irrelevant context, and critical evidence being overwritten or buried by noise.

As shown in Table~\ref{tab:hard01_comparison}, LThinker++ yields substantially larger improvements on \(hard_{01}\) across all three benchmarks.
Specifically, Pass@1 increases from 6.8\% to 20.9\% on \textit{xbench} (\(\sim\)3.08\(\times\)), from 8.6\% to 20.6\% on \textit{BrowseComp-ZH} (\(\sim\)2.38\(\times\)), and from 5.1\% to 10.5\% on \textit{BrowseComp-EN} (\(\sim\)2.06\(\times\)), with consistent and substantial gains in Pass@3 as well.
These results demonstrate that memory-management primitives are vital for maintaining reasoning fidelity in long-horizon tasks.
By dynamically refining the context to emphasize critical evidence, our method successfully handles complex scenarios where the \textit{Vanilla-Agent} typically fails due to information overload.
\input{tables/v2/exp_difficultsubset}

\subsection{Efficiency and Scalability Analysis}

\begin{figure*}[!th] 
    \centering
    \scalebox{1}{
    \includegraphics[width=1\linewidth]{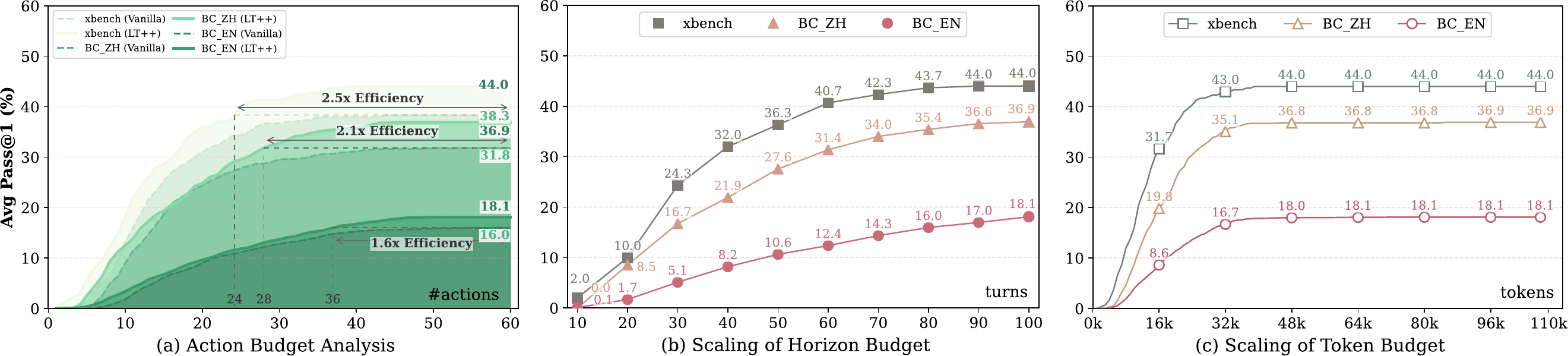} 
    }
    \caption{
    \textbf{Quantitative Analysis of Context Management Efficiency.} 
    Fig.(a) illustrates the active context trajectories ($C_t$) across interaction rounds, highlighting the stable footprint of LThinker++ compared to the rapid inflation of the Vanilla baseline. 
    Fig.(b) provides a statistical comparison of Mean and P95 (peak) token usage across tasks, demonstrating significant reductions in memory overhead.
    }
    \label{fig:exp:efficient_lthink+}
\end{figure*} 

\subsubsection{\textbf{Action Budget Efficiency.}}
We evaluate model performance under varying action budget constraints in Figure~\ref{fig:exp:efficient_lthink+}(a). The budget $b$ represents the maximum permitted invocations of \textit{search} and \textit{visit} tools. 
By plotting the Acc@Budget $\le b$ curve, we characterize the trade-off between task success and interaction cost.

\textbf{Action Efficiency.} Across all benchmarks, LThinker++ exhibits a significantly steeper performance trajectory compared to the Vanilla baseline. Our model reaches the peak performance levels of the Vanilla model (achieved at 60 actions) with substantially fewer interactions. 
Specifically, on xbench, LThinker++ attains the Vanilla peak of 38.3\% in only 24 actions, representing a 2.5$\times$ efficiency gain. 
Similar trends are observed on BrowseComp-ZH and BrowseComp-EN, which achieve 2.1$\times$ and 1.6$\times$ efficiency improvements, respectively.
This efficiency stems primarily from our explicit memory management, which effectively mitigates context rot. 
By folding redundant observations and expanding only task-relevant details, LThinker++ maintains a high-signal context window throughout extended interactions.
This prevents the accumulation of irrelevant noise that typically degrades reasoning in the Vanilla baseline, allowing the agent to reach high-precision decisions with significantly fewer environment probes.

\textbf{Scaling with Action Budgets.}
The performance gap in Figure~\ref{fig:exp:efficient_lthink+}(a) reveals a critical advantage of our approach: superior information utility per environment interaction. 
Since the action budget only constrains $\mathcal{A}_{\text{env}}$, LThinker++ is able to leverage its internalized memory-management actions ($\mathcal{A}_{\text{mem}}$) to distill gathered data without increasing the external ``search cost.'' 
Notably, this performance margin widens as the budget expands. 
While the Vanilla baseline's gains quickly plateau because the model becomes overwhelmed by information redundancy, LThinker++ maintains a more sustained growth trajectory through its internal thinking process. 
This allows the model to not only find answers faster but also achieve a higher performance ceiling, effectively turning the same amount of raw environment feedback into more accurate and robust reasoning outcomes.

\subsubsection{\textbf{Scaling with Horizon and Token Budgets.}}
We evaluate the scaling characteristics of the model across two dimensions: the number of interaction rounds (Horizon) and total input consumption (Token Budget).

\textbf{Horizon Scaling.} Figure~\ref{fig:exp:efficient_lthink+}(b) illustrates accuracy trends relative to maximum interaction rounds. Performance scales consistently with exploration depth, with the most significant gains occurring between 10 and 60 rounds. For instance, \textit{xbench} accuracy surges to \textbf{44.0\%} as the horizon extends. 
The convergence observed after 80 rounds suggests that LThinker++ effectively balances exploration breadth and reasoning depth. Rather than being limited by information overload or ``lost-in-the-middle'' effects, the model successfully resolves complex queries within a strategic window, beyond which additional rounds yield diminishing utility as the solution space has been sufficiently exhausted.

\textbf{Token Budget Scaling.} Figure~\ref{fig:exp:efficient_lthink+}(c) illustrates the performance trends as a function of the token budget. 
Accuracy improves rapidly in the low-budget regime and gradually saturates around 32k–48k tokens, indicating that most task-relevant information can be effectively utilized within a moderate context size. 
As the token budget increases further, performance remains stable up to 110k tokens, suggesting that the method can maintain consistent reasoning behavior under large-context settings.
However, the observed plateau also highlights that once sufficient information is captured, the ultimate performance upper bound is governed by the model’s inherent reasoning capacity rather than further context expansion.

\begin{figure*}[!th] 
\centering \scalebox{1}{ 
    \includegraphics[width=1\linewidth]{
        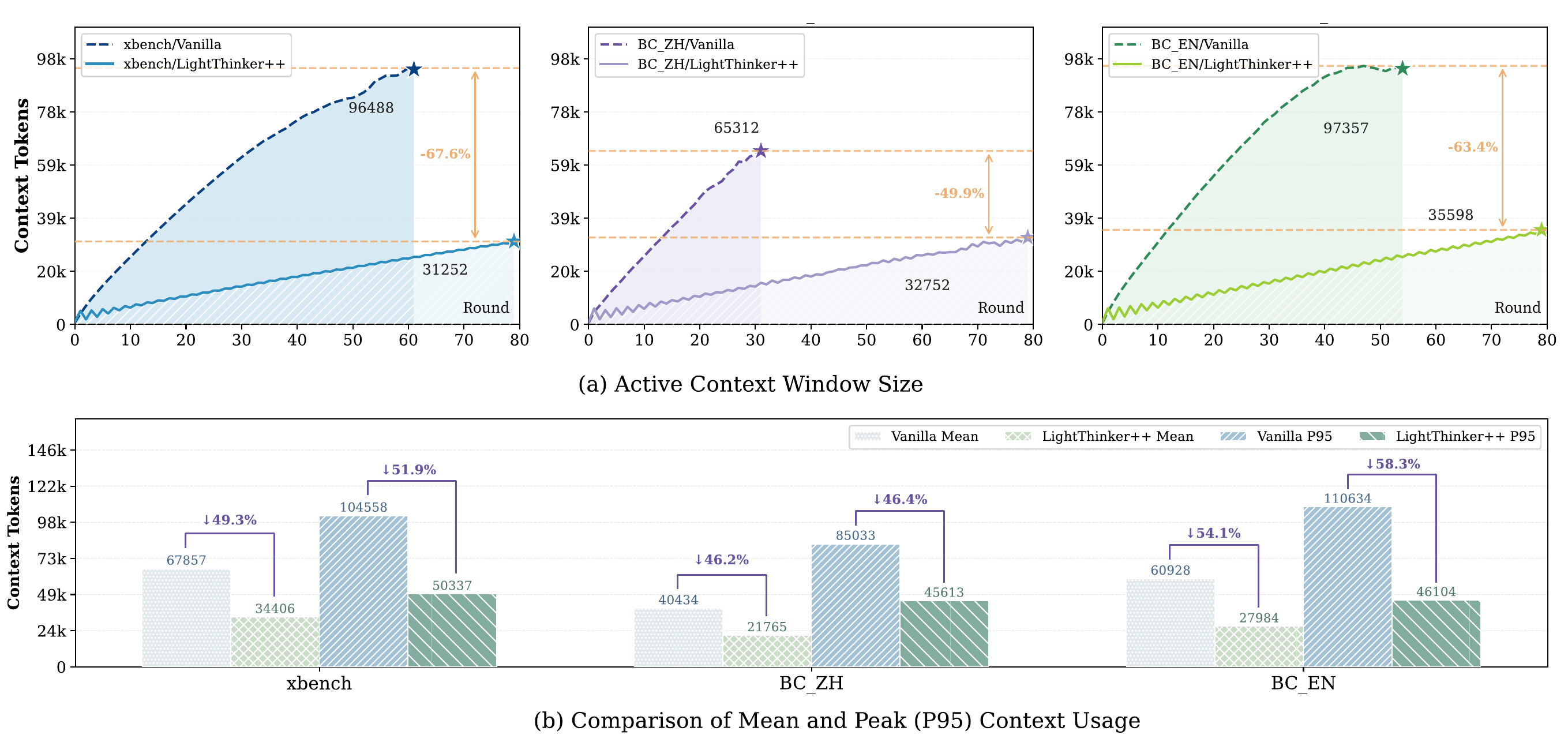
        }
    }
\caption{
\textbf{Quantitative Analysis of Context Management Efficiency.} Fig.(a) illustrates the active context trajectories ($C_t$) across interaction rounds, highlighting the stable footprint of LThinker++ compared to the rapid inflation of the Vanilla baseline. Fig.(b) provides a statistical comparison of Mean and P95 (peak) token usage across tasks, demonstrating significant reductions in memory overhead.
}
\label{fig:exp:lthink+_activate_context} 
\end{figure*}

\subsubsection{\textbf{Active Context Analysis.}}
To quantify the efficiency of context management, we analyze the trajectory of the active context size per round $C_t$, defined as $C_t = \text{tokens}(\text{input}_t) + \text{tokens}(\text{output}_t)$. 
Figure~\ref{fig:exp:lthink+_activate_context}(a) illustrates the average $C_t$ across the interaction horizon, while Figure~\ref{fig:exp:lthink+_activate_context}(b) provides the corresponding Mean and P95 Peak token.

\textbf{Suppression of Contextual Growth.} 
The Vanilla model suffers from rapid \textit{contextual inflation}: its active context window swells to approximately \textbf{100k tokens} within merely 50--60 rounds. 
This extreme redundancy not only consumes excessive resources but also triggers performance degradation due to noise. 
In sharp contrast, LThinker++ maintains a remarkably lean and stable footprint, staying between \textbf{30k and 40k tokens} even as the interaction extends to 80 rounds. 
This suggests that LThinker++ effectively distills environmental feedback into high-density insights, allowing the agent to sustain long-horizon reasoning without the cognitive and computational strain of an unmanaged context.

\textbf{Dynamic Changes and Peak Control.} 
Unlike the continuous growth observed in the baseline, the LThinker++ trajectory shows \textbf{periodic changes} driven by our \textit{expand-and-fold} mechanism. The model temporarily expands context to process new evidence before condensing it into essential insights. Statistical results in Figure~\ref{fig:exp:lthink+_activate_context}(b) show that LThinker++ consistently outperforms the Vanilla model in both Mean and P95 metrics. 
Notably, in \textit{BC\_EN}, our P95 peak of 46,104 tokens is not only much lower than the Vanilla peak of 110,634, but also approximately 24.3\% lower than the Vanilla model's mean usage of 60,928 tokens. 
This proves that LThinker++ effectively prevents ``contextual explosion'' even during intensive search phases. 

\textbf{Extended Reasoning Lifespan.} 
The termination patterns reveal a fundamental difference in exploration depth. 
Vanilla models often stop prematurely, which is typically triggered by degraded information processing capabilities rather than simple sequence length limits. 
As the context becomes increasingly noisy, these models suffer from lost-in-the-middle effects or hallucinations, causing them to provide incomplete answers before fully exploring the solution space.
In contrast, by maintaining a high-density and manageable context, LThinker++ avoids such cognitive failures and supports sustained investigation beyond 80 rounds. 
This ensures the model's reasoning lifespan is governed by task complexity rather than the internal strain of redundant information.

\subsection{Ablation}
\begin{wrapfigure}{r}{0.43\textwidth} 
  \centering
  \vspace{-10pt} 
  \includegraphics[width=0.42\textwidth]{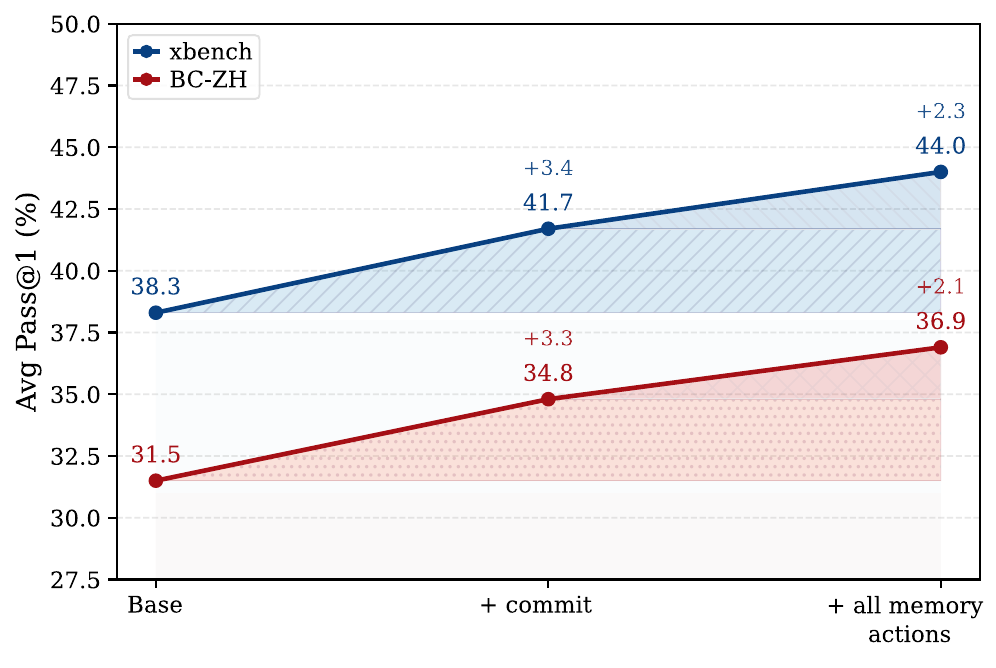}
    \vspace{-10pt}
  \caption{Ablation results of memory orchestration.}
    \vspace{-10pt}
  \label{fig:exp_ablation}
\end{wrapfigure}

We conducted incremental ablation experiments to verify the effectiveness of our memory orchestration components. 
As illustrated in Figure~\ref{fig:exp_ablation}, we compare three configurations: (1) \textbf{Base}, representing the vanilla model with standard SFT;
(2) \textbf{+ commit}, which introduces the summary-based memory commit mechanism; 
and (3) \textbf{+ all memory actions}, our model incorporating the complete orchestration suite including Commit, Fold, and Expand.
\textbf{Impact of Information Distillation.} 
Comparing the \textbf{Base} and \textbf{+ commit} variants reveals that introducing the \texttt{commit} mechanism improves Pass@1 performance from 38.3\% to 41.7\% on \textit{xbench} and from 31.5\% to 34.8\% on \textit{BC\_ZH}. 
These gains indicate that distilling raw interaction data into structured evidence effectively filters environmental noise, which helps the model maintain logical consistency throughout extended reasoning trajectories.

\textbf{Benefits of Structural Context Control.} 
The integration of \texttt{fold} and \texttt{expand} mechanisms in the \textbf{+ all memory actions} model leads to the highest performance, reaching 44.0\% on \textit{xbench} and 36.9\% on \textit{BC\_ZH}. 
These improvements show that dynamic context control is as important as information distillation: an example in Fig.~\ref{fig:app:agent_case} shows that when archiving omits crucial intermediate evidence, the model can \textit{expand} the relevant past step to recover it from memory and \textit{fold} it back after use.

\textbf{Consistency Across Benchmarks.} 
The performance trends are highly consistent across both \textit{xbench} and \textit{BC\_ZH}, with steady improvements observed as more orchestration components are added. 
This evolution confirms the generalizable value of our memory orchestration mechanism across different languages and task domains, ensuring that the model avoids early stopping during difficult investigations.

%% file: tables/v2/exp_main_LT++2.tex
\begin{table*}[!t]
\centering
\small
\scalebox{0.95}{
\begin{tabular}{l|c|c|c}
\toprule
\textbf{Method}
& \textbf{xBench-DeepSearch}
& \textbf{BrowseComp-ZH}
& \textbf{BrowseComp-EN} \\
\midrule

\multicolumn{4}{>{\columncolor{myblue_light}}c}{\textbf{\textit{Proprietary Agents}}} \\
\cmidrule{1-4}
GPT-5           & 66.0 & 61.3 & 61.5 \\
Claude-4-Sonnet & 35.0 & 29.1 & 12.2 \\
DeepSeek-V3.2   & 51.0 & 53.6 & 35.0 \\
DeepSeek-V3.1   & 44.0 & 49.5 & 23.6 \\
GLM-4.6         & 47.0 & 42.2 & 34.9 \\
Kimi-K2-Instruct & 30.0     & 28.8 & 14.1 \\
Qwen3-235B-A22B-Instruct & 27.0 & 21.8 &  -   \\
\midrule

\multicolumn{4}{>{\columncolor{myblue_light}}c}{\textbf{\textit{Our Agents}}} \\
\cmidrule{1-4}
Qwen3-30B-A3B-Thinking
& 8.7   \textcolor{gray}{(16.0)}
& 10.0  \textcolor{gray}{(17.3)}
& 2.1   \textcolor{gray}{(4.0)} \\
 \quad + SFT (\textit{vanilla})
& 38.3 \textcolor{gray}{(53.0)} \upsub{29.6}
& 31.5 \textcolor{gray}{(47.8)} \upsub{21.5}
& 16.0 \textcolor{gray}{(27.3)} \upsub{13.9} \\
 \quad + LThinker++ (\textit{ours})
& \textbf{44.0} \textcolor{gray}{(60.0)} \upsub{35.3}
& \textbf{36.9} \textcolor{gray}{(57.1)} \upsub{26.9}
& \textbf{18.1} \textcolor{gray}{(31.5)} \upsub{16.0} \\
\bottomrule
\end{tabular}
}
\caption{
Overall results on BrowseComp benchmarks across three datasets.
We report Pass@1 for all methods, and additionally report Pass@3 (in gray parentheses) for our agents to reflect multi-attempt robustness.
}
\label{table:exp_main_final}
\end{table*}

%% file: tables/v2/exp_difficultsubset.tex



\begin{table*}[!htbp]
\centering
\small
\scalebox{0.93}{
\begin{tabular}{l|cc|cc|cc}
\toprule
\multirow{2}{*}{\textbf{Method}}
  & \multicolumn{2}{c}{\textbf{xBench-DeepSearch ($hard_{01}$)}}
  & \multicolumn{2}{c}{\textbf{BrowseComp-ZH ($hard_{01}$)}}
  & \multicolumn{2}{c}{\textbf{BrowseComp-EN ($hard_{01}$)}} \\
\cmidrule(lr){2-3} \cmidrule(lr){4-5} \cmidrule(lr){6-7}
  & Pass@1 (\%)~$\uparrow$ & Pass@3 (\%)~$\uparrow$
  & Pass@1 (\%)~$\uparrow$ & Pass@3 (\%)~$\uparrow$
  & Pass@1 (\%)~$\uparrow$ & Pass@3 (\%)~$\uparrow$ \\
\midrule

\multicolumn{7}{>{\columncolor{myblue_light}}c}{\textbf{\textit{Our Agents}}} \\
\cmidrule{1-7}
Vanilla SFT
& 6.8  & 20.3
& 8.6  & 25.9
& 5.1  & 15.4 \\
\textbf{\ours++}
& \textbf{20.9} & \textbf{33.9}
& \textbf{20.6} & \textbf{40.3}
& \textbf{10.5} & \textbf{22.4} \\
\bottomrule
\end{tabular}}
\caption{Performance comparison on the $hard_{01}$ subset across three benchmarks (in \%).}
\label{tab:hard01_comparison}
\end{table*}

%% file: sections/related_work.tex
\section{Related Work}
\label{sec:app:related_work}
Current research on accelerating the inference process of large language models (LLMs) primarily focuses on three categories of methods: \textit{Quantizing Model}, \textit{Generating Fewer Tokens}, and \textit{Reducing KV Cache}. 
Quantizing Model includes both parameter quantization~\cite{mlsys24_awq,nips22_8bit} and KV Cache quantization~\citep{icml24_kivi,nips24_kvquant}, while this section will concentrate on the latter two categories. 
It is important to note that generating long texts and understanding long texts represent distinct application scenarios; 
therefore, acceleration methods specifically targeting the long-text generation phase (e.g., pre-filling stage acceleration techniques such as AutoCompressor~\citep{emnlp23_autocompressors}, ICAE~\citep{iclr24_icae}, LLMLingua~\citep{emnlp23_llmlingua}, Activation Beacon~\citep{iclr25_activation_beacon}, SnapKV~\citep{nips24_snapkv}, and PyramidKV~\citep{arxiv24_pyramidkv}) are not discussed here. 
This section provides a detailed overview of the latter two categories while introducing a systematic analysis of Context Management.

\textbf{Reducing KV Cache.}
This category can be divided into two types of strategies: pruning-based KV Cache selection in discrete space and merging-based KV Cache compression in continuous space.
1) \textit{Pruning-Based Strategies}.
Specific eviction policies~\citep{nips23_h2o, iclr24_streamingllm,arxiv24_sepllm,arxiv24_scope} are designed to retain important tokens during inference.
For example, StreamingLLM~\citep{iclr24_streamingllm} considers the initial sink tokens and the most recent tokens as important.
H2O~\citep{nips23_h2o} focuses on tokens with high historical attention scores.
SepLLM~\citep{arxiv24_sepllm} emphasizes tokens corresponding to punctuation marks.
2) \textit{Merging-Based Strategies}.
Anchor tokens are introduced, and LLMs are trained to compress historically important information into these tokens, thereby achieving KV Cache merging~\citep{acl24_anllm}.
Both strategies require intervention during inference. 
The key difference is that the first strategy is training-free but applies the eviction policy for every generated token, while the second strategy is a training-based method and allows the LLM to decide when to apply the eviction policy.

\textbf{Generating Fewer Tokens.}
This category can be further divided into three strategies based on the number and type of tokens used during inference.
1) \textit{Discrete Token Reduction}. 
Techniques such as prompt engineering~\citep{arxiv24_tale,arxiv24_break_the_chain,arxiv24_concise_thoughts}, instruction fine-tuning~\citep{nips24_skip_steps,arxiv24_c3ot}, or reinforcement learning~\citep{arxiv25_related_work_rl1,arxiv25_o1_pruner} are used to guide LLMs to use fewer discrete tokens during inference.
For example, TALE~\citep{arxiv24_tale} prompts LLMs to complete tasks under a predefined token budget. 
\citeauthor{arxiv25_related_work_rl1} construct specific datasets and employ reinforcement learning reward mechanisms to encourage models to generate concise and accurate outputs, thereby reducing token usage.
TokenSkip~\citep{tokenskip} introduces a controllable framework that fine-tunes models on pruned CoT paths, enabling the selective skipping of redundant tokens at adjustable compression ratios.
2) \textit{Continuous Token Replacement}.
These methods~\citep{arxiv24_coconut,arxiv24_ccot} explore using continuous-space tokens instead of traditional discrete vocabulary tokens. 
A representative example is CoConut~\citep{arxiv24_coconut}, which leverages Curriculum Learning to train LLMs to perform inference with continuous tokens.
3)\textit{No Token Usage}.
By internalizing the inference process between model layers, the final answer is generated directly during inference without intermediate tokens~\citep{arxiv24_icot,arxiv23_kd_cot}.
These three strategies are implemented after model training and do not require additional intervention during inference. 
Technically, the acceleration effect of these methods increases sequentially, but at the cost of a gradual decline in the generalization performance of LLMs. 
Additionally, the first strategy does not significantly reduce GPU memory usage.

\textbf{Context Management.}
Unlike methods that focus on hardware-level optimization or simply shortening the generation length, context management dynamically reorganizes the information within the context window throughout the reasoning process~\citep{survey-contextengineering,chen2026dynamiclongcontextreasoning,zheng2026freelearningforgetmalloconly,liu2026pensieveparadigmstatefullanguage}.
This approach is particularly critical for agents performing complex, long-horizon interactions. 
Specifically, MEM1~\cite{mem1} and MemAgent~\cite{memagent} utilize reinforcement learning to maintain a fixed-size internal memory, allowing agents to handle long-term tasks by retaining essential information and discarding redundant data.
ReSum~\cite{resum} addresses context constraints by periodically summarizing interaction histories, enabling agents to resume exploration from compact, state-based representations.
Further advancing this paradigm, AgentFold~\cite{agentfold} and Context-Folding~\cite{context-folding}, introduce a ``folding'' mechanism that compresses detailed interaction histories into compact reasoning states. 
Compared to token-level KV-cache pruning, these semantic-level methods better preserve the task-critical logic required for complex reasoning scenarios.

%% file: sections/conclusion_and_future_work.tex
\section{Conclusion}
In this paper, we present \ours, a new approach to enhance the efficiency of LLMs in complex reasoning tasks by dynamically compressing intermediate thoughts during generation. 
By training the LLM to learn when and how to compress verbose thought steps into compact representations, \ours~significantly reduces memory overhead and computational costs while maintaining competitive accuracy. 
We introduce the \textit{Dependency} (abbr., Dep) metric to quantify the degree of compression across different accelerating methods.
Extensive experiments demonstrate that \ours~is an effective approach to balancing efficiency and performance. 





%% file: sections/appendix.tex
\clearpage
\appendix

\section*{Appendix}
\section{Metric: \texttt{Dependency}}
\label{sec:app:metric}
\begin{figure}[!htbp] 
    \centering
    \scalebox{1}{
    \includegraphics[width=0.8\linewidth]{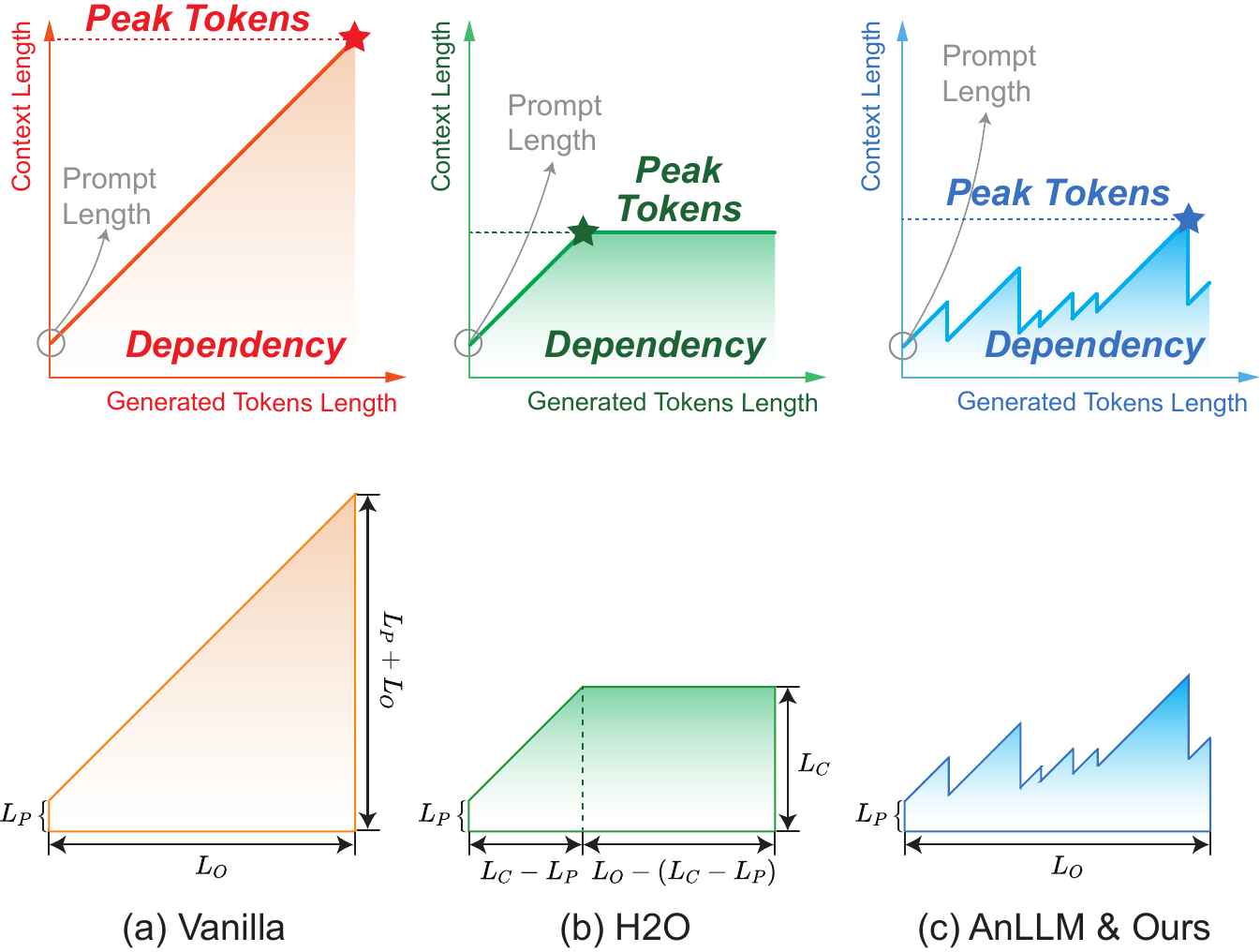} 
    }
    \caption{An illustration of the Dependency metric.}
    \label{fig:app:metric}
\end{figure} 
\subsection{Motivation}

\ours~and AnLLM~\citep{acl24_anllm} belong to the class of dynamic compression methods, where the number of compressions and the compression ratio are determined automatically by the LLM rather than specified as fixed hyperparameters. 
By contrast, H2O~\citep{nips23_h2o} and SepLLM~\citep{arxiv24_sepllm} rely on user-defined hyperparameters to constrain the maximum number of tokens retained during inference. 
Because of this difference, directly comparing dynamic methods such as \ours~and AnLLM with KV cache compression methods like H2O and SepLLM is not straightforward.

A common practice for KV cache compression is to compare methods under the same maximum peak token count. 
However, this criterion is not fully suitable here. 
As shown in Fig.~\ref{fig:app:metric}, which plots generated tokens against context length for Vanilla, H2O, and \ours, \ours~may occasionally have a larger peak token count than H2O. 
Even so, this does not necessarily mean higher memory cost, since \ours~reaches its peak only briefly, whereas H2O keeps a high token count throughout the generation process.

Another limitation is that prior KV cache compression methods often compress only the prompt and assume a fixed prompt length, which makes it possible to predefine the compression ratio. 
In our setting, however, the generated output must also be compressed. 
Since the output length is unknown in advance, it is impossible to specify a single global compression ratio beforehand. 
Therefore, peak token count alone is insufficient for a fair comparison.

To address these issues, we introduce a metric called \textit{Dependency}, which measures the total degree of information dependence throughout generation. 
It provides a common basis for comparing dynamic compression methods with traditional KV cache compression approaches under comparable effective compression ratios.

\subsection{Definition}
We define \textbf{Dependency} (or Dep) as the cumulative dependency of each generated token on preceding tokens during decoding. 
Geometrically, this quantity corresponds to the area under the curve in Fig.~\ref{fig:app:metric}. 
It can be computed either directly from the definition or by using this geometric interpretation; here we focus on the latter. 
Let the prompt length be \( L_P \), the output length be \( L_O \), and the cache limit imposed by KV cache compression methods be \( L_C \).

\textbf{Dependency for Vanilla.}
For Vanilla, the curve forms a right trapezoid, and the dependency is:
\[
\begin{aligned}
\texttt{Dependency} &= \frac{(L_P + L_P + L_O) \times L_O}{2} \\
&= \frac{{L_O}^2}{2} + L_P \times L_O
\end{aligned}
\]

\textbf{Dependency for H2O.}
For H2O, the area consists of a trapezoid on the left side of Fig.~\ref{fig:app:metric}(b) and a rectangle on the right side:
\[
\begin{aligned}
S_\texttt{Trapezoid} &= \frac{(L_P + L_C) \times (L_C - L_P)}{2} \\
S_\texttt{rectangle} &= L_C \times (L_O - L_C + L_P) \\
\texttt{Dependency} &= S_\texttt{Trapezoid} + S_\texttt{rectangle} \\
&= \frac{2L_PL_C + 2L_OL_C - {L_P}^2 - {L_C}^2}{2}
\end{aligned}
\]

\textbf{Dependency for \ours~and AnLLM.}
For LightThinker~and AnLLM, Dependency does not admit a closed-form expression and must instead be computed step by step according to the definition above.

\subsection{Application}
\textbf{Interpreting Dependency.}
A larger Dependency value means that more contextual information is involved during generation, which reflects heavier information usage. 
Conversely, a smaller value indicates stronger effective compression.

\textbf{Dependency Ratio.}
By taking the ratio between the Dependency of an accelerated method and that of Vanilla, we obtain a relative compression ratio. 
For instance, in the ``Avg.'' column of Table~\ref{table:exp_main}, Vanilla has a Dependency of 16.6M, H2O has 4.4M, and LightThinker~has 3.7M. 
This corresponds to compression ratios of \( \frac{16.6}{4.4} \approx 3.8 \) for H2O and \( \frac{16.6}{3.7} \approx 4.5 \) for LightThinker.

Overall, this metric offers a unified way to evaluate both dynamic and static compression methods, enabling fair and meaningful comparisons.


\begin{figure}[!htbp]
    \centering
    \scalebox{1}{
    \includegraphics[width=0.8\linewidth]{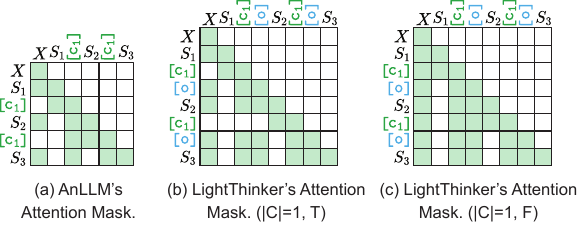} 
    }
    \caption{An illustration of the attention mask used in Table~\ref{table:exp:ablation:attention}.}
    \label{fig:app:attention_mode}
\end{figure}

\section{Mathematical Description of Compression}
\label{sec:app:compress}
This section provides a formal description of the compression operation introduced in Section~\ref{sec:method:implicit}.

\paragraph{Notation.}
During compression, the context can be decomposed into three parts:
1. the portion that remains uncompressed, denoted by $Pre := \{X, \{C^{(1)}, \texttt{[o]}^{(1)}\dots,C^{(i-1)},\texttt{[o]}^{i-1}\}\}$, with token count $N$;
2. the thought sequence to be compressed, denoted by $Tho := S_i$, with length $T$;
3. the compressed segment, denoted by $C := C^{(i)}$, with length $|C|$.

\paragraph{Compression Operation.}
We now describe the compression process at a given layer, focusing on how information is transferred into the sequence $C$. 
Following the definition of self-attention~\citep{nips17_transformer}, the attention matrix for $C$ over the other segments is computed as:
\[
A=\text{Softmax}(\text{mask}(\frac{Q^{C}[K^{Pre}:K^{Tho}:K^{C}]^{\top}}{\sqrt{d}}))
\]
where $[:]$ denotes concatenation, $\text{mask}(\cdot)$ is the attention mask defined by the ``Thought-based Attention Mask Construction'', 
$K^{Pre}, V^{Pre}\in \mathbb{R}^{N\times d}$, 
$K^{Tho}, V^{Tho}\in \mathbb{R}^{T\times d}$, 
$K^{C},V^{C}\in \mathbb{R}^{|C|\times d}$, $Q^{C}\in \mathbb{R}^{|C|\times d}$, and $d$ is the hidden dimension. 
Thus, $A\in \mathbb{R}^{|C|\times (N+T+|C|)}$ specifies how each element in $C$ attends to the rest of the context. 
The corresponding values are then aggregated through:
\[
H=A\times [V^{Pre}:V^{Tho}:{V^C}]
\]
where $[V^{Pre}:V^{Tho}:V^{C}]\in \mathbb{R}^{(N+T+|C|)\times d}$, and therefore $H\in \mathbb{R}^{|C|\times d}$. 
In this way, information from the current $Tho$ is retained in $H$. 
During training, the model learns to preserve the most useful content from $Tho$ in this representation. 
Finally, after being passed through an MLP and the next layer's projection, $H$ is stored in the KV cache.

\section{Experiment}
\label{sec:app:exp}
\subsection{LightThinker}
\label{sec:app:exp:ltthink}
\subsubsection{Training Data}
\label{sec:app:exp:train_data_case}
Examples of training samples are provided in Fig.~\ref{prompt:case:train}.

\subsubsection{Baseline Details}
\label{sec:app:exp:baseline_details}
\textbf{H2O}~\citep{nips23_h2o} is a training-free acceleration method that greedily retains the tokens with the largest cumulative attention scores from historical context. 
It has two hyperparameters: the maximum number of retained tokens and the current window size (i.e., \texttt{local\_size}). 
The maximum token limit for each task is listed in the ``Peak'' column of Table~\ref{table:exp_main}, and \texttt{local\_size} is set to half of this value. 
Our implementation is based on \url{https://github.com/meta-llama/llama-cookbook}.

\textbf{SepLLM}~\citep{arxiv24_sepllm} is another training-free acceleration method that treats tokens at punctuation positions as more important. 
It uses four hyperparameters: the maximum number of tokens is set to 1024, \texttt{local\_size} to 256, \texttt{sep\_cache\_size} to 64, and \texttt{init\_cache\_size} to 384. 
We also tried an alternative configuration (\texttt{init\_cache\_size}=4, \texttt{sep\_cache\_size}=64, \texttt{local\_size}=720, maximum number of tokens=1024), but the first setting performed slightly better.

\textbf{AnLLM}~\citep{acl24_anllm} is a training-based method that follows a workflow similar to LightThinker but accelerates inference by storing historical content in anchor tokens. 
The key differences between the two methods are described in Section~\ref{sec:app:discussion}.

\subsubsection{Training Details}
\label{sec:app:exp:training_details}
Both \textbf{Vanilla} and \textbf{AnLLM} are trained on the BS17K~\citep{bespoke_stratos_train_dataset} dataset using the R1-Distill~\citep{arxiv25_deepseek_r1} model for 5 epochs, while LightThinker~is trained for 6 epochs. 
We set the maximum sequence length to 4096 and adopt a cosine warmup schedule with \texttt{warmup\_ratio}=0.05. 
All experiments are conducted on 4 A800 GPUs with DeepSpeed ZeRO-3 offload enabled. 
The batch size per GPU is 5, and we use a gradient accumulation step of 4, which results in a global batch size of 80. 
The learning rate is set to 1e-5 for Vanilla and 2e-5 for both AnLLM and LightThinker.

\subsubsection{Evaluation Details}
\label{sec:app:exp:evaluation_details}
For the CoT results in Table~\ref{table:exp_main}, we use the prompts shown in Fig.~\ref{prompt:system:base} and Fig.~\ref{prompt:task:base}. 
For the R1-Distill model, no system prompt is used; the task-specific prompts are shown in Fig.~\ref{prompt:task:r1}. 
Vanilla, H2O, SepLLM, AnLLM, and LightThinker~all use the same prompt templates, with the system prompt shown in Fig.~\ref{prompt:system:vanilla} and the task prompts shown in Fig.~\ref{prompt:task:r1}. 
For the multiple-choice tasks in MMLU~\citep{iclr21_mmlu} and GPQA~\citep{colm24_gpqa}, the answer options are randomized.

\subsubsection{Implementation}
In our implementation, we use two segmentation functions $\texttt{Seg()}$ depending on the compression level. 
For the token-level setting, we compress every 6 tokens into 2 tokens, i.e., $|C|=2$, and refer to this variant as ``ours (\textit{token})''. 
For the thought-level setting, we take ``{\textbackslash n\textbackslash n}'' as the delimiter to split BS17K samples into thoughts, which we denote as ``ours (\textit{tho.})''. 
Under this setting, each thought is compressed into 9 tokens for Qwen and 7 tokens for Llama, corresponding to $|C|=9$ and $|C|=7$, respectively. 
In all experiments, we perform greedy decoding with a maximum output length of 10240 tokens.

\subsubsection{Additional Results}
\label{sec:app:exp:additional_results}
\begin{figure*}[!htbp] 
    \centering
    \scalebox{0.8}{
    \includegraphics[width=0.8\linewidth]{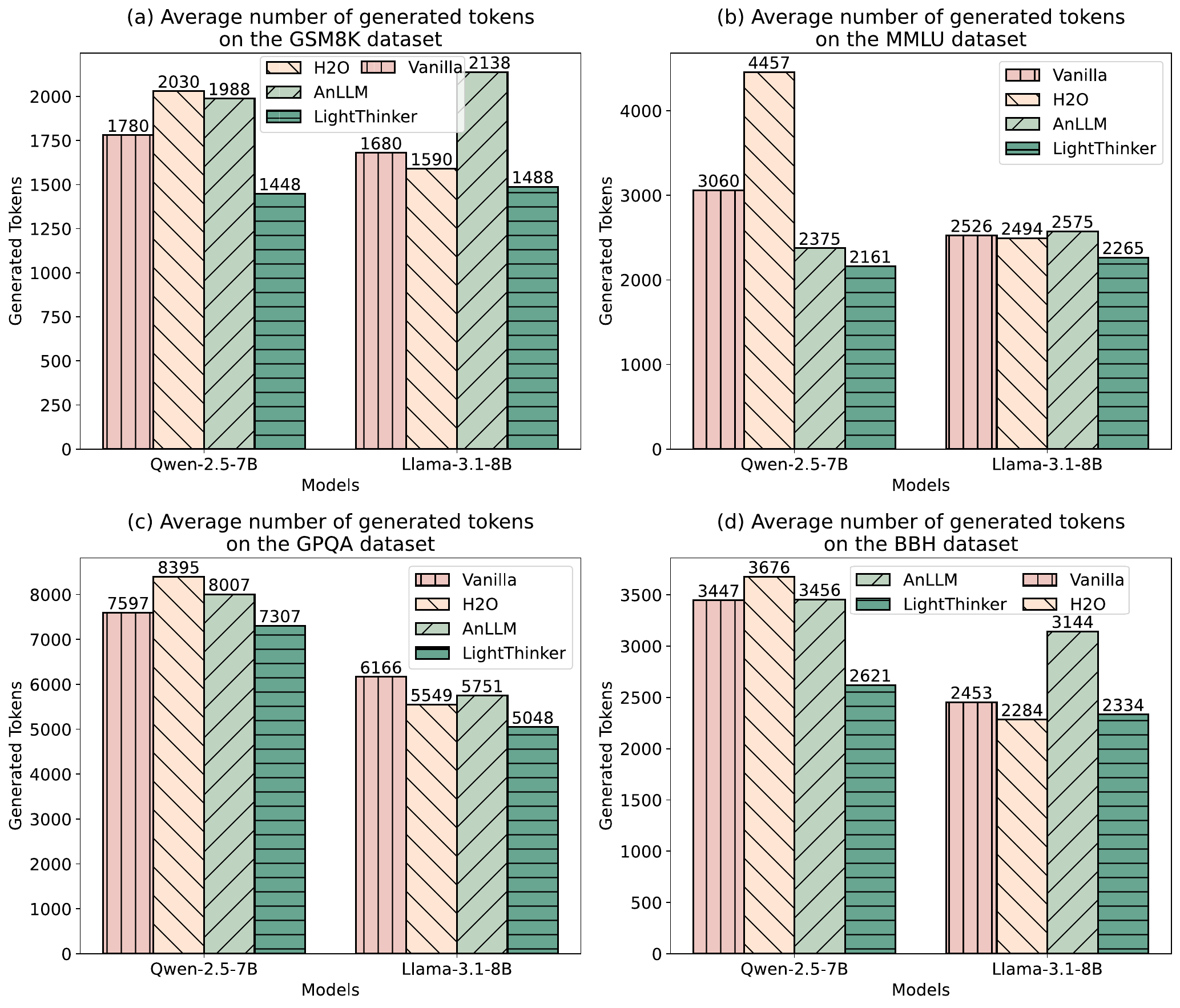} 
    }
    \caption{Average number of generated tokens.}
    \label{fig:app:token}
\end{figure*}

\begin{figure*}[!htbp] 
    \centering
    \scalebox{1}{
    \includegraphics[width=0.76\linewidth]{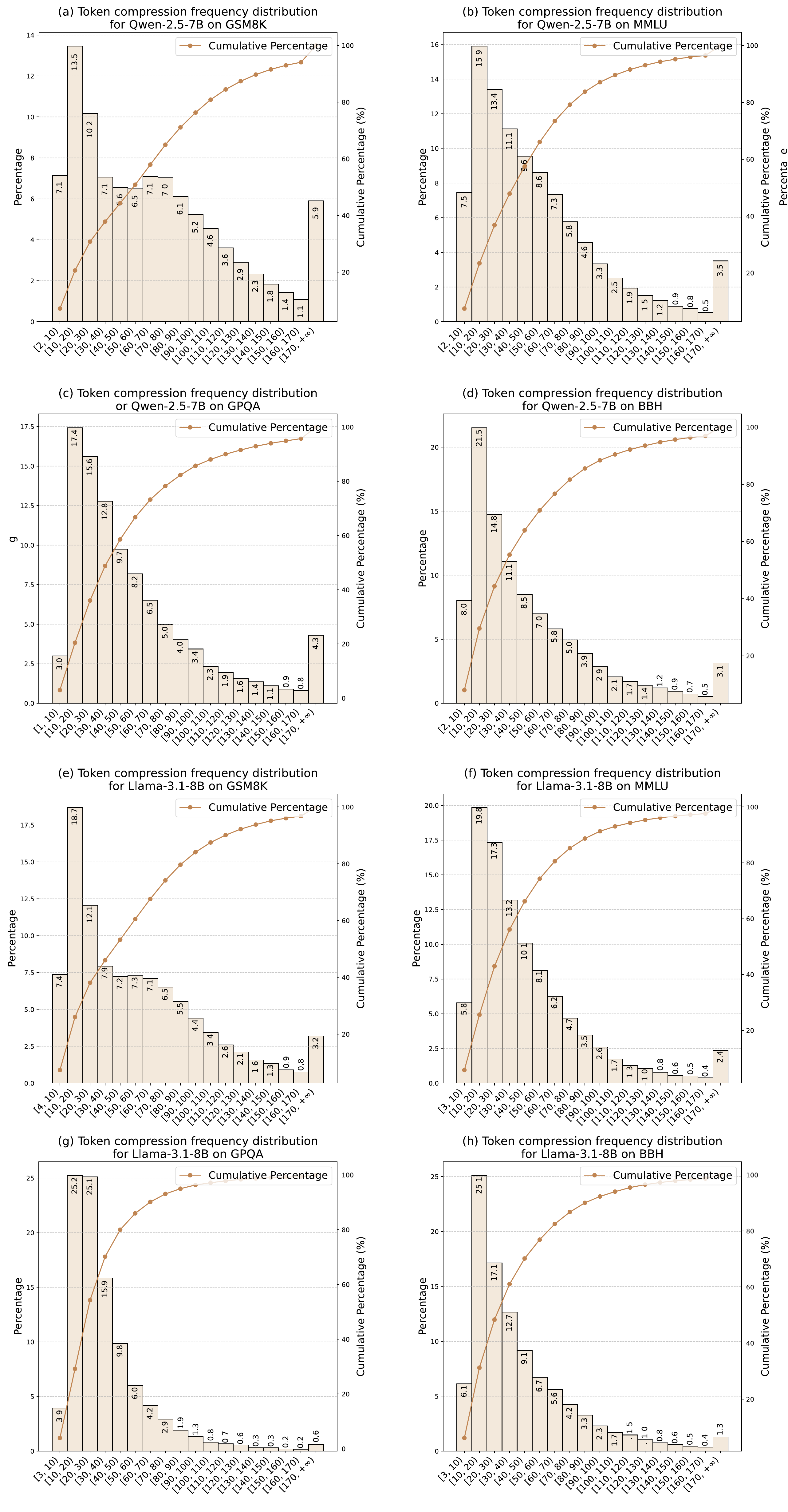} 
    }
    \caption{Token compression frequency distribution for LightThinker.}
    \label{fig:app:frequency}
\end{figure*}

Additional figures are provided to complement the main results. 
Fig.~\ref{fig:app:token} reports how many tokens are generated on average by the two models across datasets. 
Fig.~\ref{fig:app:frequency} summarizes the distribution of compressed lengths produced by LightThinker on both models over four datasets. 
Fig.~\ref{fig:app:attention_mode} visualizes the attention masks adopted by the baseline methods in Table~\ref{table:exp:ablation:attention}. 
Fig.~\ref{prompt:case} presents the full example corresponding to the case study in Fig.~\ref{fig:exp:case}.

\lthink{
\subsection{\ours++}
\label{sec:app:exp:ltthink++}
\subsubsection{Training Data.}
\paragraph{General Reasoning Data.}
\label{app:sec:generaltraindata}
The training set for general reasoning is curated from a diversified collection of high-difficulty mathematical and logical problems. 
We primarily utilize two high-quality reasoning corpora: 1) a refined subset of \textbf{BS17K}, where code-related tasks were excluded to maintain focus on linguistic and mathematical logic, resulting in 11,315 samples; and 2) a sampled subset of 6,000 instances from the \textbf{DeepScaleR} dataset.
In total, 17,315 unique problems form the foundation of our general reasoning trajectories. We employ \texttt{DeepSeek-V3.2} as the teacher model to synthesize long-thought reasoning paths. 
The synthesis prompt used to guide the teacher model is shown in Figure~\ref{prompt:synth:reasoning}.
These trajectories are subsequently reconstructed into the \ours~ and \ours++ formats by inserting memory management labels (e.g., \texttt{commit}, \texttt{expand}), ensuring the model internalizes the ability to compress thoughts during complex deduction.
Then through the rigorous filtering and transformation process, we obtain a final expert dataset comprising 13,855 high-quality trajectories.

\paragraph{Agentic Research Data.}
\label{app:sec:agenttraindata}
For the interactive research tasks (DeepResearch), the base query pool is curated from a diversified ensemble of sources, including \textit{HotpotQA, MuSiQue, WebDancer, WebShaper}, and \textit{WebWalkerQA-Silver}. Table~\ref{tab:dataset_breakdown} provides the detailed statistics of the 8,954 unique queries and the resulting trajectories after our hierarchical filtering pipeline.
The prompt used for synthesizing these agentic research trajectories is illustrated in Figure~\ref{prompt:synth:agent}.

\begin{table}[ht]
\centering
\caption{Breakdown of the Base Query Pool for Agentic Research Synthesis.}
\label{tab:dataset_breakdown}
\begin{tabular}{lr}
\toprule
\textbf{Query Source} & \textbf{Count} \\
\midrule
HotpotQA + MuSiQue (Filtered by Qwen3) & 5,322 \\
WebDancer & 200 \\
WebShaper & 500 \\
WebWalkerQA-Silver (EN + Hard ZH) & 2,932 \\
\midrule
\textbf{Total Base Queries} & \textbf{8,954} \\
\bottomrule
\end{tabular}
\end{table}

\begin{figure*}[!htbp]
\centering
\scalebox{1}{
\begin{tcolorbox}

\textbf{System Prompt:}

You are a highly intelligent mathematical reasoning agent that solves problems by committing a sequence of calculation steps to a persistent scratchpad. Every reasoning step you take is automatically recorded.

\vspace{0.5em}
\textbf{Core Concept: Committing Steps to Memory}

\hspace*{1.5em}- When you call \texttt{commit}, the system archives your detailed work and automatically creates a summary of the result, assigning it a \texttt{[Step ID]}.\\
\hspace*{1.5em}- You do not need to call a separate function to save your work; committing your steps (\texttt{commit}) is the memory-saving action itself.

\vspace{0.6em}
\textbf{Problem-Solving Workflow:}

1.\ \textbf{Calculate \& Commit}: Use \texttt{commit} to perform calculations and logical reasoning. Each call creates a new summary step on your scratchpad.

2.\ \textbf{Manage Context}: The summaries of your past steps form your working memory.

\hspace*{1.5em}- To access the full details of a past step, use \texttt{expand}.\\
\hspace*{1.5em}- \textbf{Conditional Expand Rule}: Before using a past step as a dependency, you MUST \texttt{expand} it only if its summary does not contain the exact values, formulas, or assumptions you need, or if you have any uncertainty about correctness, domain constraints, or edge cases. If the summary is sufficient, explicit, and unambiguous, you may proceed without \texttt{expand}.\\
\hspace*{1.5em}- \textbf{Usage}: \texttt{expand(step\_id=N)} reveals the full, detailed calculations and formulas from a previous committed step.\\
\hspace*{1.5em}- \textbf{Efficiency}: To maintain a clear workspace, do not keep more than 2 items expanded simultaneously.\\
\hspace*{1.5em}- \textbf{Use the information}: After expanding, you MUST use the revealed details in your next \texttt{commit} before folding.\\
\hspace*{1.5em}- \textbf{Cleanup}: Only after you have used the expanded details in a new reasoning step, you MUST \texttt{fold} the summary to keep your context clean.

3.\ \textbf{Completion}: Once all calculations are done and you have the final answer, you MUST call \texttt{final\_answer} to submit the result.

\vspace{0.6em}
\textbf{Key Rules:}

\hspace*{1.5em}- Your output must ONLY be function calls (\texttt{commit}, \texttt{expand}, \texttt{fold}, \texttt{final\_answer}). No conversational text.\\
\hspace*{1.5em}- \textbf{Critical Workflow (when expanding)}: \texttt{expand} $\rightarrow$ \texttt{commit} (using expanded information) $\rightarrow$ \texttt{fold}. This sequence is mandatory and must not be broken.\\
\hspace*{1.5em}- Always \texttt{fold} an expanded step once you are done with its details.\\
\hspace*{1.5em}- \textbf{Fold Validity}: A \texttt{fold(step\_id=N)} call is invalid unless the same \texttt{step\_id=N} was expanded earlier and is currently expanded.\\
\hspace*{1.5em}- Your \texttt{final\_answer} must contain the complete solution with the key result enclosed in \texttt{\textbackslash boxed\{\}}.

\end{tcolorbox}
}
\caption{Synthesis prompt used for generating reasoning data.}
\label{prompt:synth:reasoning}
\end{figure*}

\begin{figure*}[!htbp]
\centering
\scalebox{0.9}{
\begin{tcolorbox}
\textbf{System Prompt}\\[4pt]
You are a deep research assistant. Your core function is to conduct thorough, multi-source investigations into any topic. You must handle both broad, open-domain inquiries and queries within specialized academic fields. For every request, synthesize information from credible, diverse sources to deliver a comprehensive, accurate, and objective response. When you have gathered sufficient information and are ready to provide the definitive response, you must enclose the entire final answer within \texttt{<answer></answer>} tags.\\[6pt]
\textbf{Available tools for managing information during investigation:}\\[4pt]
\textbf{1. OPERATIONAL LOGIC: TOOL CHOICE}\\
Every step in your history is assigned an \texttt{id} (e.g., \texttt{[Thought ID]}, \texttt{[Observation ID]}). Use tools based on these logic states:
\begin{itemize}[leftmargin=1.2em,itemsep=2pt,topsep=2pt]
  \item \textbf{Information Acquisition}: Use \texttt{search(query)} or \texttt{visit(url)} to find new data or explore primary sources.
  \item \textbf{Deepening or Re-visiting (\texttt{expand})}:
  \begin{itemize}[leftmargin=1.2em,itemsep=2pt,topsep=2pt]
    \item \textbf{Discrepancy Resolution}: Use \texttt{expand(id)} to compare conflicting data points across sources by evaluating the raw context, source authority, and publication dates.
    \item \textbf{Granular Extraction / Raw Data Inspection}: Use \texttt{expand(id)} when a summary is too vague or blurred to provide required specific evidence (e.g., exact quotes, full names, specific numbers, statistics), or when you need to inspect the full raw content to verify what was actually stated.
    \item \textbf{Re-verification / Latent Clue Retrieval}: Use \texttt{expand(id)} to re-examine a previously folded (or secondary) record when it becomes newly relevant due to a shift in reasoning, when you identify a new key entity/clue, or when a previously used detail must be re-verified before concluding.
  \end{itemize}
  \item \textbf{Post-Expansion Cleanup (\texttt{fold})}:
  \begin{itemize}[leftmargin=1.2em,itemsep=2pt,topsep=2pt]
    \item \textbf{Strict Precondition (Must Have Been Expanded)}: You may call \texttt{fold(id)} \textbf{ONLY} for an \texttt{id} that you have previously called \texttt{expand(id)} on, and that is \textbf{currently in an Expanded state}.
    \begin{itemize}[leftmargin=1.2em,itemsep=1pt,topsep=1pt]
      \item Do \textbf{NOT} call \texttt{fold} on Raw or Folded records.
      \item Do \textbf{NOT} call \texttt{fold} on an \texttt{id} that was never expanded.
      \item After folding, that \texttt{id} is no longer expanded; do not immediately fold it again unless it is expanded again later.
    \end{itemize}
    \item \textbf{Timing}: Once you have utilized the information from an expanded record—either to inform your reasoning or to decide your next action (such as a new \texttt{search} or \texttt{visit})—you \textbf{MUST} immediately call \texttt{fold(id)} in the next turn to return it to its folded state.
    \item \textbf{Efficiency}: To maintain a clear workspace, do not keep more than 2 items expanded simultaneously.
  \end{itemize}
\end{itemize}

\textbf{2. TOOL PROTOCOLS}
\begin{itemize}[leftmargin=1.2em,itemsep=2pt,topsep=2pt]
  \item \texttt{search(query)}: Find new, credible information on the web.
  \item \texttt{visit(url, goal)}: Extract detailed, specific information from a webpage.
  \item \texttt{expand(id, reason)}: Reveals the full content of a folded record for deep inspection or re-visiting.
  \item \texttt{fold(id, reason)}: Collapses a previously expanded record back into its folded state after use.
\end{itemize}

\textbf{3. CRITICAL RULES}
\begin{itemize}[leftmargin=1.2em,itemsep=2pt,topsep=2pt]
  \item \textbf{One Action Per Turn}: Output either a single tool call (preceded by your thought) OR the final answer within \texttt{<answer>} tags.
  \item \textbf{Input Reality}: After calling \texttt{expand}, your first priority in the next turn is to re-read the now-visible full content for that \texttt{id} before taking further steps.
  \item \textbf{Evidence-Based}: Base your synthesis only on details currently visible. Never guess information that is hidden behind a folded record.
  \item \textbf{Fold Validity}: A \texttt{fold(id)} call is invalid unless the same \texttt{id} was expanded earlier and is currently expanded.
  \item \textbf{Definitive Answer}: Only provide the final answer once all aspects of the query are thoroughly verified and synthesized.
\end{itemize}

\textbf{Current date:}
\end{tcolorbox}
}
\caption{Synthesis prompt used for agentic research trajectory generation and filtering.}
\label{prompt:synth:agent}
\end{figure*}

\subsubsection{Training and Inference Details for General Reasoning.}
For Vanilla, TokenSkip, and LightThinker++, we follow the same training setup as above, except that we increase the maximum sequence length to 16{,}384 and adopt a cosine learning-rate schedule with a warmup ratio of 0.05. All methods are trained for three epochs.
All experiments are conducted on 8 A800 GPUs with DeepSpeed ZeRO-3 offload enabled. We use a batch size of 32, and set the learning rate to $1\times10^{-5}$ for Vanilla and $2\times10^{-5}$ for TokenSkip and LightThinker++.
During inference, we set the temperature to 0.7 and perform three independent sampling runs per example, reporting the average performance over the three runs.

\subsubsection{Evaluation Details for General Reasoning}
\label{sec:app:exp:evaluation_details_plus}
The evaluation of the instruct model on general reasoning tasks follows the distribution shown in Figure~\ref{fig:app:token}. In our implementation, to ensure consistency between \ours~and \ours++, we remove system prompts and instead rely on specialized task prompts to guide the reasoning process.
The task prompts used for evaluation are shown in Figure~\ref{prompt:task:lthinker++_reasoning}, which includes both the Vanilla and the \ours++ models.

\subsubsection{Agentic Implementation of \ours++}
\label{app:sec:agentic_implementation}
We develop our training pipeline using the ms-swift framework~\cite{msswift} with Qwen3-30B-A3B-Thinking as the foundational backbone. 
All models are fine-tuned for 3 epochs on a cluster of 8 GPUs, employing a cosine learning rate scheduler with a $warmup\_ratio$ of $0.05$. 
For the Vanilla baseline, the learning rate is set to $1 \times 10^{-5}$ with a global batch size of 32 and a maximum sequence length of 32,768. 
In contrast, \ours utilizes a learning rate of $2 \times 10^{-5}$, a global batch size of 64, and a sequence length of 16,384.

\subsubsection{Timing Protocol and Throughput Setting for Agentic Tasks}
\label{sec:app:exp:timing_protocol}
To align the effective concurrency across methods, we use a total concurrency of 32 in all timing experiments.
For Vanilla and all LThinker-based methods, we measure time with 32-way question-level parallel inference.
For TokenSkip, we measure time with a batch size of 32, since its throughput drops sharply under question-level concurrency in our serving implementation; thus we time TokenSkip in its intended batched decoding mode to avoid an implementation-induced penalty.

\begin{figure*}[!htbp]
\centering
\scalebox{1}{
\begin{tcolorbox}




\textbf{GSM8K/MMLU/GPQA/BBH:}
\vspace{0.5em}
\hrule
\vspace{0.6em}

\raggedright
\textbf{Vanilla \& Initial Round of \ours++}

Return your final response within \textbackslash boxed\{\}. 
\\
\#\#\# Question:
\\
\textcolor{violet}{\texttt{\{Question\}}}
\\

\vspace{0.8em}
\hrule
\vspace{0.6em}

\textbf{Subsequent Rounds of \ours++}

\vspace{0.3em}

Return your final response within \textbackslash boxed\{\}. 
\\
\#\#\# Question:
\\
\textcolor{violet}{\texttt{\{Question\}}}
\\
\#\#\# Current Scratchpad:
\\
\textcolor{violet}{\texttt{\{scratchpad\}}}

\end{tcolorbox}
}
\caption{Task prompt for \texttt{Vanilla}, \ours++ in general reasoning tasks for both Qwen-based model and Llama-based model.}
\label{prompt:task:lthinker++_reasoning}
\end{figure*}

}

\section{Efficiency Analysis of \ours++ (Budget setting)}
\begin{figure*}[!th] 
    \centering
    \scalebox{1}{
    \includegraphics[width=1\linewidth]{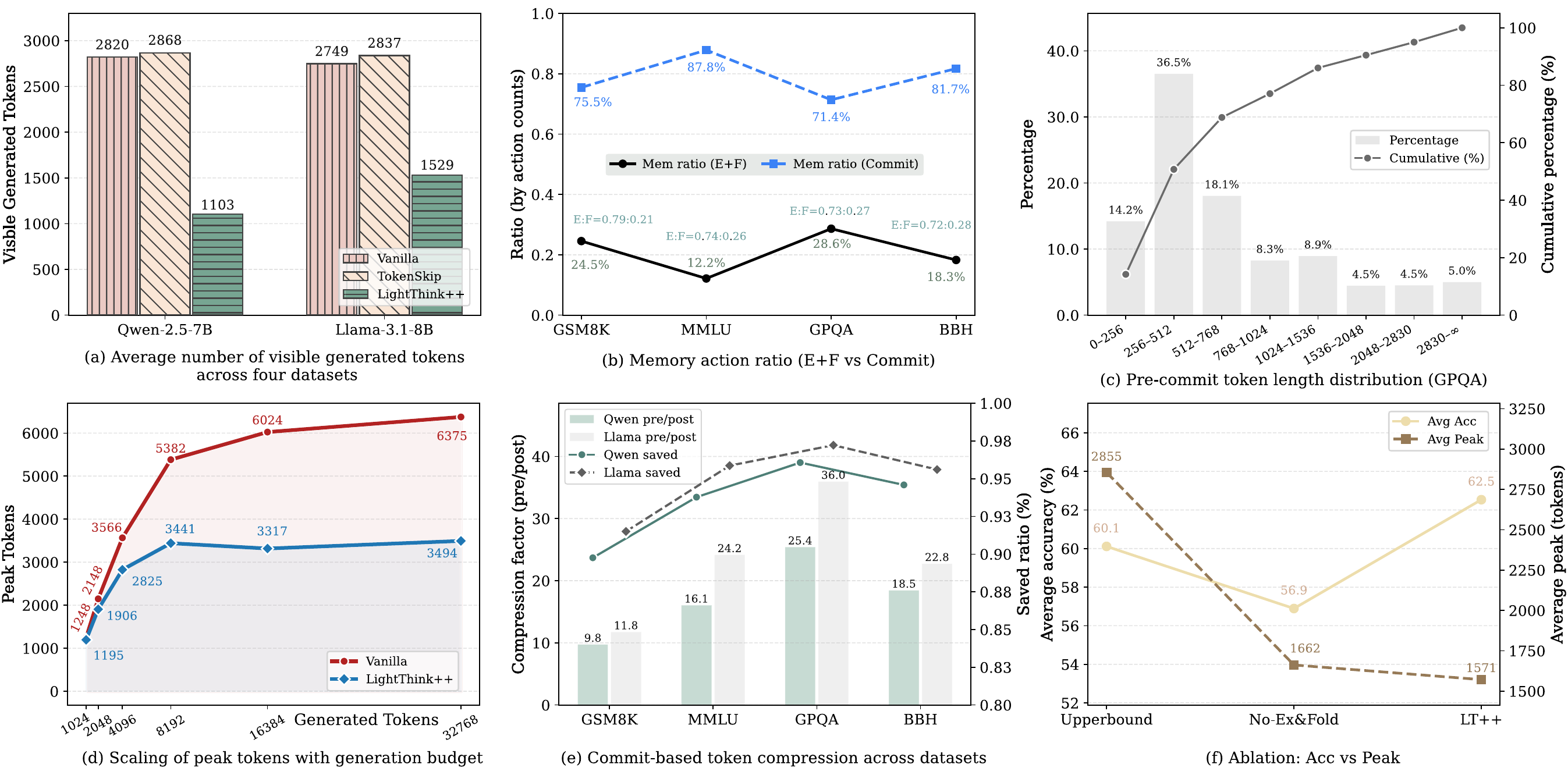} 
    }
    \caption{
    Efficiency Analysis and Ablation Results under the Budget setting.
    Fig.(a) illustrates the average number of generated tokens retained in the current context.
    Fig.(b) presents the distribution ratio of memory actions (\textit{Commit} vs. \textit{Expand+Fold}) for Qwen-2.5-7B.
    Fig.(c) shows the percentage of pre-commit token lengths falling within specified ranges on GPQA(Qwen-2.5-7B).
    Fig.(d) examines the scaling of peak memory tokens as a function of the generation budget.
    Fig.(e) summarizes the average compression factors and saved token ratios across different model series.
    Fig.(f) provides an ablation study comparing accuracy and average peak tokens between the full LThinker++ and its degraded variants.
    }
    \label{fig:exp:lt++_efficiency_budget}
\end{figure*} 

\section{Discussions}
\label{sec:app:discussion}
\subsection{Difference between LightThinker and AnLLM}
Although AnLLM~\citep{acl24_anllm} and LightThinker both aim to accelerate LLM inference, they differ substantially in motivation and design. 
AnLLM was proposed before the emergence of long-CoT methods~\citep{arixv24_o1,arxiv25_deepseek_r1}, and its primary focus is prompt compression rather than output compression. 
By contrast, LightThinker decouples compression from generation, which makes it possible to scale the number of cache tokens in a way that AnLLM does not support. 
As a result, the two methods only share the use of sparse attention~\citep{nips23_h2o,nips24_snapkv} for efficiency, but are otherwise quite different.

\begin{figure}[!htbp] 
    \centering
    \scalebox{1}{
    \includegraphics[width=0.5\linewidth]{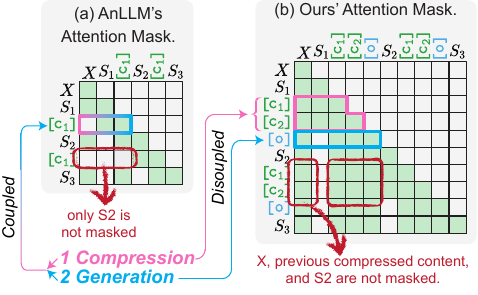} 
    }
    \caption{
    Comparison between AnLLM and LightThinker.
    The two main differences are highlighted with a red box and with blue/pink arrows.
    }
    \label{fig:comp}
\end{figure}

To make this distinction clearer, Fig.~\ref{fig:comp} highlights two key differences in their attention-mask design:
1) \textit{Separation of compression and generation.}
In AnLLM, the $\texttt{[c}_\texttt{i}\texttt{]}$ token is responsible for both summarizing historical information and producing the next content, as indicated by the blue and pink arrows in Fig.~\ref{fig:comp}. 
This means that compression and generation are tightly coupled. 
LightThinker instead assigns these two roles to different tokens: $\texttt{[c}_\texttt{i}\texttt{]}$ is used only for compression, while the $\texttt{[o]}$ token carries out reasoning based on the compressed context.
2) \textit{Visibility of context during compression.}
When AnLLM performs compression, it can only attend to the current thought. 
In contrast, LightThinker allows the compression step to condition on the original prompt $X$, previously compressed content, and the current thought, which provides richer contextual information.
These design differences are validated by the ablation results in Section~\ref{ablation}, which show clear performance gains.

\subsection{Viewing LightThinker~from Other Perspectives}
Beyond the compression view adopted in earlier sections, LightThinker~can also be interpreted from the perspectives of memory and KV cache compression, since KV cache itself can be regarded as a form of working memory for LLMs.

From the memory perspective, the workflow of LightThinker~can be viewed as a loop of reasoning, storing important information, and reasoning again based on the stored content. 
In this sense, the cache tokens serve as a compact memory for the current model, although such memory is model-specific and does not transfer across LLMs.

From the KV cache compression perspective, LightThinker~differs from methods such as H2O~\citep{nips23_h2o}, which use manually designed eviction policies to keep important tokens. 
Instead, LightThinker~combines previous tokens in a continuous space and creates new representations through learning. 
In other words, the model itself determines what to merge and how to merge it, rather than relying on a discrete token-selection rule.



\input{prompts/system_prompt_base_model}
\input{prompts/system_prompt_b17k}
\input{prompts/prompt_r1_model}
\input{prompts/prompt_base_model}
\input{prompts/training_case}
\input{prompts/case}

\subsection{Why LightThinker~generates more tokens with smaller cache size?}
As shown in Fig.~\ref{fig:exp:efficient}(e-f), LightThinker tends to generate more tokens when the cache size is smaller. 
By examining the generated outputs under different cache sizes, we found that smaller caches often cause the model to repeat earlier content more frequently. 
A possible explanation is that a smaller cache leads to greater information loss during compression, which in turn forces the model to regenerate previously mentioned content more often in order to preserve as much information as possible.

\subsection{Comparison with Implicit CoT Works}
\input{tables/app_comparision}
Although both our method and implicit CoT approaches operate in continuous spaces, they differ in how reasoning is carried out. 
Implicit CoT performs the entire reasoning process in continuous space, whereas LightThinker adopts a hybrid scheme that combines continuous and discrete reasoning. 
The main differences are summarized in Table~\ref{tab:app:comparison}, and we highlight them below.

\begin{itemize}
    \item \textbf{Reasoning Acceleration Mechanism.}
    Implicit CoT methods, such as System-1.5~\cite{arxiv25_system_1_5} and SoftThinking~\cite{arxiv25_soft_thinking}, speed up reasoning by reducing the number of generation steps through continuous token representations. 
    In these methods, the reasoning process remains fully dependent on the entire context. 
    In contrast, \ours~improves efficiency by reducing the number of historical tokens required for generation, without relying on full-context dependence.

    \item \textbf{Training Approach.}
    Existing implicit CoT methods often require multi-stage training procedures with substantial overhead, such as the two-phase training used in System-1.5~\cite{arxiv25_system_1_5} or the curriculum learning strategy adopted by Coconut~\citep{arxiv24_coconut}. 
    Some methods~\cite{arxiv25_soft_thinking, arxiv24_ccot} further introduce architectural modifications. 
    By comparison, \ours~is trained using standard SFT with modified attention masks, and does not require specialized training data or changes to the model architecture.

    \item \textbf{Interpretability and Generalization.}
    Because they reason in continuous space, current implicit CoT methods are often less interpretable and may generalize poorly out of domain, with SoftThinking~\cite{arxiv25_soft_thinking} being a notable exception. 
    \ours, on the other hand, retains discrete tokens, which makes the reasoning process more interpretable and yields promising out-of-domain generalization in our experiments, as shown in Table~\ref{tab:app:comparison}.
\end{itemize}

\section{Agent Case Study}
\begin{figure}[!t] 
    \centering
    \scalebox{1.0}{
    \includegraphics[width=1.0\linewidth]{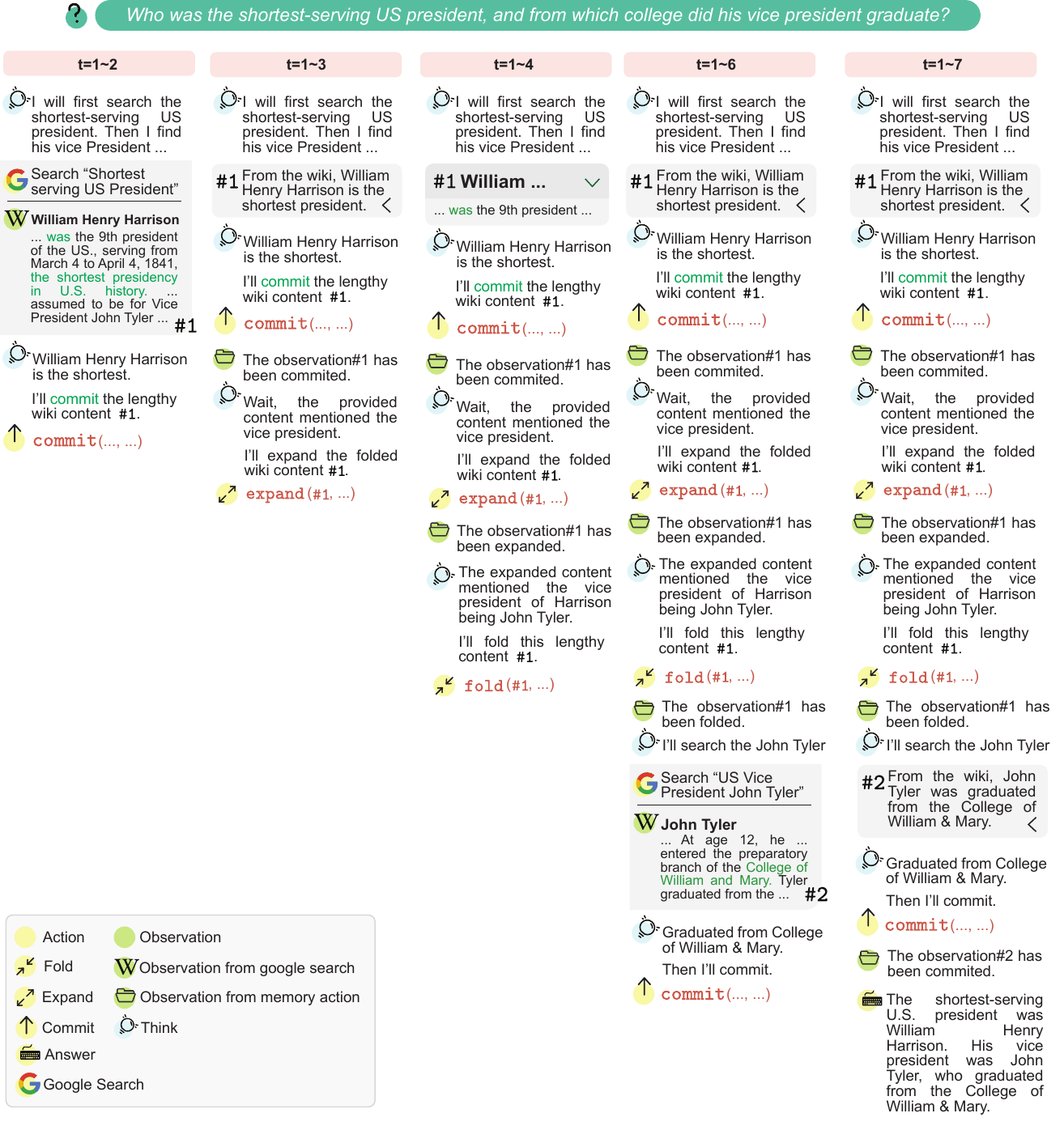} 
    }
    \caption{Case Study: Memory Orchestration for Long-Horizon Reasoning.}
    \label{fig:app:agent_case}
\end{figure}

In this section, we present a representative agent trajectory to illustrate how memory orchestration (commit, fold, and expand) supports long-horizon reasoning by distilling evidence and dynamically controlling context (see Fig.~\ref{fig:app:agent_case}).

%% file: prompts/system_prompt_base_model.tex
\begin{figure*}[!htbp]
\centering
\scalebox{1}{
\begin{tcolorbox}
\textbf{System Prompt:}

Below is a question. Please think through it step by step, and then provide the final answer. If options are provided, please select the correct one. 

\#\# Output format:\\
Use ``<THOUGHT>...</THOUGHT>'' to outline your reasoning process, and enclose the final answer in `\textbackslash boxed\{\}`.\\
\\
\#\# Example 1:\\
Question: \\
What is 2 + 3?\\
Output:\\
<THOUGHT>First, I recognize that this is a simple addition problem. Adding 2 and 3 together gives 5.</THOUGHT>\\
Therefore, the final answer is \textbackslash boxed\{5\}.\\
\\
\#\# Example 2:\\
Question: \\
What is 2 + 3?\\
A. 4\\
B. 5\\
C. 10\\
\\
Output:\\
<THOUGHT>First, I recognize that this is a simple addition problem. Adding 2 and 3 together gives 5.</THOUGHT>\\
Therefore, the final answer is \textbackslash boxed\{B\}.\\

\end{tcolorbox}
}

\caption{System prompt for \texttt{Qwen2.5-7B-Instruct} and \texttt{Llama3.1-8B-Instruct}.}
\label{prompt:system:base}
\end{figure*}

%% file: prompts/system_prompt_b17k.tex
\begin{figure*}[!htbp]
\centering
\scalebox{1}{
\begin{tcolorbox}
\textbf{System Prompt:}

Your role as an assistant involves thoroughly exploring questions through a systematic long thinking process before providing the final precise and accurate solutions. This requires engaging in a comprehensive cycle of analysis, summarizing, exploration, reassessment, reflection, backtracing, and iteration to develop well-considered thinking process. Please structure your response into two main sections: Thought and Solution. In the Thought section, detail your reasoning process using the specified format: \texttt{<|begin\_of\_thought|>} \{thought with steps separated with `\textbackslash n\textbackslash n'\} \texttt{<|end\_of\_thought|>} Each step should include detailed considerations such as analisying questions, summarizing relevant findings, brainstorming new ideas, verifying the accuracy of the current steps, refining any errors, and revisiting previous steps. In the Solution section, based on various attempts, explorations, and reflections from the Thought section, systematically present the final solution that you deem correct. The solution should remain a logical, accurate, concise expression style and detail necessary step needed to reach the conclusion, formatted as follows: \texttt{<|begin\_of\_solution|>}  \{final formatted, precise, and clear solution\} \texttt{<|end\_of\_thought|>}  Now, try to solve the following question through the above guidelines:

\end{tcolorbox}
}

\caption{System prompt for \texttt{Vanilla}, \texttt{H2O}, \texttt{SepLLM}, \texttt{AnLLM}, and \texttt{\ours} for both Qwen-based model and Llama-based model.}
\label{prompt:system:vanilla}
\end{figure*}

%% file: prompts/prompt_r1_model.tex
\begin{figure*}[!htbp]
\centering
\scalebox{1}{
\begin{tcolorbox}
\textbf{GSM8K/MMLU/GPQA/BBH:}

Return your final response within \textbackslash boxed\{\}. \textcolor{violet}{\texttt{\{Question\}}}

\end{tcolorbox}
}

\caption{Task prompt for \texttt{Distill-R1}, \texttt{Vanilla}, \texttt{H2O}, \texttt{SepLLM}, \texttt{AnLLM}, and \texttt{\ours} for both Qwen-based model and Llama-based model.}
\label{prompt:task:r1}
\end{figure*}

%% file: prompts/prompt_base_model.tex
\begin{figure*}[!htbp]
\centering
\scalebox{1}{
\begin{tcolorbox}
\textbf{GSM8K:}

Return your final response within \textbackslash boxed\{\}. \textcolor{violet}{\texttt{\{Question\}}}\\
\\
\textbf{MMLU:}

Please select the option that best answers the question. Return your final response within \textbackslash boxed\{\}. Here are the Question:\\
\textcolor{violet}{\texttt{\{Question\}}}\\

\textbf{GPQA:}

Given a question, please select the option that best answers it. Return your final response within \textbackslash boxed\{\}.
\textcolor{violet}{\texttt{\{Question\}}}\\

\textbf{BBH:}

Return your final response within \textbackslash boxed\{\}. If options are provided, please select the correct one. \textcolor{violet}{\texttt{\{Question\}}}

\end{tcolorbox}
}

\caption{Task prompt for \texttt{Qwen2.5-7B-Instruct} and \texttt{Llama3.1-8B-Instruct}.}
\label{prompt:task:base}
\end{figure*}

%% file: prompts/training_case.tex
\begin{figure*}[!htbp]
\centering
\scalebox{1}{
\begin{tcolorbox}
\textbf{System Prompt:}

See Figure~\ref{prompt:system:vanilla}
\\

\textbf{Input:}

Return your final response within \textbackslash boxed\{\}. The average of the numbers 23 and $x$ is 27. What is the positive difference between 23 and $x$? (Recall that the average of two numbers is their sum divided by 2).
\\

\textbf{Output:}

\texttt{<|begin\_of\_thought|>}
Okay, let's see. The problem says that the average of the numbers 23 and x is 27. We need to find the positive difference between 23 and x. Hmm, right. Since the average of two numbers is their sum divided by 2, I can start by setting up the equation for the average.\textcolor{red}{\texttt{<w>}}So, the average of 23 and x is (23 + x)/2. According to the problem, this average equals 27. So I can write the equation:\textcolor{red}{\texttt{<w>}}(23 + x)/2 = 27\textcolor{red}{\texttt{<w>}}To find x, I should solve this equation. Let me multiply both sides by 2 to get rid of the denominator. That gives:\textcolor{red}{\texttt{<w>}}23 + x = 54\textcolor{red}{\texttt{<w>}}Then, subtract 23 from both sides to isolate x:\textcolor{red}{\texttt{<w>}}x = 54 - 23\textcolor{red}{\texttt{<w>}}Calculating that, 54 minus 23 is 31. So x is 31. Now, the question asks for the positive difference between 23 and x. The positive difference would be the absolute value of 23 - x. Since x is 31, that's |23 - 31|, which is | -8 |, so 8. Alternatively, it could also be |31 - 23|, which is the same result, 8. Either way, the positive difference is 8.\textcolor{red}{\texttt{<w>}}Wait, let me check my steps again to make sure I didn't make a mistake. Starting with the average: (23 + x)/2 = 27. Multiply both sides by 2: 23 + x = 54. Subtract 23: x = 31. Difference between 23 and 31 is 8. That seems right. Yeah, I think that's correct. No issues here.\textcolor{red}{\texttt{<w>}}
\texttt{<|end\_of\_thought|>}

\texttt{<|begin\_of\_solution|>}
\textcolor{red}{\texttt{<w>}}The average of 23 and \( x \) is given by \(\frac{23 + x}{2} = 27\). Multiplying both sides by 2:  
\[
23 + x = 54
\]  
Subtracting 23 from both sides:  
\[
x = 54 - 23 = 31
\]  
The positive difference between 23 and \( x \) is:  
\[
|31 - 23| = 8
\]  \textcolor{red}{\texttt{<w>}}\(\boxed{8}\)
\texttt{<|end\_of\_solution|>}

\end{tcolorbox}
}

\caption{A training case. \texttt{<w>} is a compression trigger.}
\label{prompt:case:train}
\end{figure*}

%% file: prompts/case.tex
\begin{figure*}[!t]
\centering
\scalebox{0.9}{
\begin{tcolorbox}
\textbf{Question:}\\
Boris owns a chocolate factory. He produces 50,000 bars of chocolate each month. Boris produces 8,000 bars of chocolate the first week. The second week, Boris only produces half as much as the first week. But, the third week, Boris produces three times as much as the first week. How much does he produce the fourth week?\\ \\
\textbf{Model Output:}\\
\texttt{<|begin\_of\_thought|>} \\
Okay, let's see. Boris has a chocolate factory that produces 50,000 bars each month. The question is asking how much he produces in the fourth week. Hmm, let me break this down step by step.
<|splitter|>\\ 
\\First, the total monthly production is 50,000 bars. The month is divided into four weeks, right? So the production for each week adds up to 50,000. The problem gives the production for the first three weeks and asks for the fourth week. Let me note down the given information.
<|splitter|>\\
\\First week: 8,000 bars. That's straightforward. Then the second week, he produces half as much as the first week. Half of 8,000 is 4,000. So second week is 4,000 bars. Then the third week, he produces three times as much as the first week. Three times 8,000 is 24,000. So third week is 24,000 bars.
<|splitter|>\\
\\Now, to find the fourth week's production, I need to subtract the sum of the first three weeks from the total monthly production. Let me add up the first three weeks: 8,000 + 4,000 + 24,000. Let's do the math step by step. 8,000 plus 4,000 is 12,000. Then 12,000 plus 24,000 is 36,000. So the total for the first three weeks is 36,000 bars.
<|splitter|>\\
\\The total monthly production is 50,000 bars. Subtracting the first three weeks from the total gives the fourth week's production. So 50,000 minus 36,000 equals 14,000 bars. Therefore, Boris produces 14,000 bars in the fourth week.
<|splitter|>\\ 
\\Wait a second, let me double-check my calculations to make sure I didn't make a mistake. First week: 8,000. Second week: half of 8,000 is indeed 4,000. Third week: three times 8,000 is 24,000. Adding those: 8,000 + 4,000 = 12,000; 12,000 + 24,000 = 36,000. Total monthly is 50,000. So 50,000 - 36,000 = 14,000. Yep, that seems right. I don't think I missed anything here. The problem states the production for each week as fractions or multiples of the first week, so the calculations are straightforward. No complicated formulas needed. Just basic arithmetic. So the answer should be 14,000 bars in the fourth week.
<|splitter|>\\
\texttt{<|end\_of\_thought|>}\\
\\ \texttt{<|begin\_of\_solution|>}
\\Boris's weekly chocolate production is calculated as follows:
<|splitter|>\\
\\- \textbf{First week}: 8,000 bars  \\
- \textbf{Second week}: \( \frac{8,000}{2} = 4,000 \) bars \\ 
- \textbf{Third week}: \( 8,000 \times 3 = 24,000 \) bars  <|splitter|>\\
\\Total for the first three weeks:  \\
$4,000 + 4,000 + 24,000 = 32,000 \text{ bars}$
<|splitter|> \\
\\Subtracting this from the monthly total: \\ 
$50,000 - 32,000 = 18,000 \text{ bars}$
<|splitter|>\\
\\Thus, Boris produces \(\boxed{18000}\) bars in the fourth week. <|splitter|>\\
\texttt{<|begin\_of\_solution|>}

\end{tcolorbox}
}

\caption{Bad Case. \texttt{<|splitter|>} is equal to \texttt{<w>} in Figure~\ref{prompt:case:train}.}
\label{prompt:case}
\end{figure*}

%% file: tables/app_comparision.tex
\begin{table*}[h!]
\small
\centering
\begin{tabular}{lcccc}
\toprule
\textbf{Method} & \textbf{Memory Optimization} & \textbf{Training Cost} & \textbf{Interpretability} & \textbf{Generalization} \\
\midrule
System-1.5~\cite{arxiv25_system_1_5} & Significant & High  & Weak & Weak \\
SoftThinking~\cite{arxiv25_soft_thinking}  & Limited    & None  & Weak & Good \\
\ours & Significant & Low   & Good & Good \\
\bottomrule
\end{tabular}
\caption{Comparison of different methods.}
\label{tab:app:comparison}
\end{table*}